\documentclass{article}

\PassOptionsToPackage{numbers, compress}{natbib}
\usepackage[eandd, preprint]{neurips_2026}

\usepackage[utf8]{inputenc} 
\usepackage[T1]{fontenc}    
\usepackage{hyperref}       
\usepackage{url}            
\usepackage{booktabs}       
\usepackage{amsfonts}       
\usepackage{nicefrac}       
\usepackage{microtype}      
\usepackage[table]{xcolor}  
\usepackage{listings}       
\usepackage{multirow}       
\usepackage{array}
\usepackage{graphicx}
\usepackage{enumitem}
\usepackage{amssymb}
\usepackage{bbding}
\usepackage{marginnote}
\usepackage{tabularx}
\usepackage{makecell}
\usepackage{wrapfig}
\usepackage{caption}
\usepackage[most]{tcolorbox}
\tcbuselibrary{breakable, skins}
\usepackage{threeparttable}
\usepackage{CJKutf8}
\usepackage{tikz}           
\usepackage{etoolbox}       
\usepackage{longtable}
\usepackage{tikz}
\usepackage{tocloft}
\usepackage{xcolor}
\usepackage{titletoc}
\usepackage{changepage}
\usepackage[normalem]{ulem}
\usepackage{placeins}

\newcommand{\posdelta}[1]{\textcolor{blue!35!gray}{#1}}
\newcommand{\negdelta}[1]{\textcolor{red!35!gray}{#1}}

\newcommand\benchmark{KnowVis}
\newcommand\klingimageomnithree{{Kling Image 3.0 Omni}}

\newcommand\cmark{\Checkmark}
\newcommand\xmark{\XSolidBrush}

\definecolor{tagEntity}{HTML}{D6EAF8}
\definecolor{tagEntityDark}{HTML}{2E86C1}

\definecolor{tagVisual}{HTML}{D5F5E3}
\definecolor{tagVisualDark}{HTML}{1E8449}
\definecolor{tagKnowledge}{HTML}{FDEBD0}
\definecolor{tagKnowledgeDark}{HTML}{CA6F1E}

\definecolor{tagRequired}{HTML}{FADBD8}
\definecolor{tagRequiredDark}{HTML}{C0392B}
\definecolor{tagOptional}{HTML}{EAF0FB}
\definecolor{tagOptionalDark}{HTML}{5B6DAE}
\definecolor{tagAdditional}{HTML}{F4ECF7}
\definecolor{tagAdditionalDark}{HTML}{7D3C98}

\definecolor{tagAnnotation}{HTML}{D6EAF8}
\definecolor{tagAnnotationDark}{HTML}{2E86C1}

\definecolor{tagExcl}{HTML}{E8F5E9}
\definecolor{tagExclDark}{HTML}{2E7D32}

\definecolor{tagGood}{HTML}{E3F2FD}
\definecolor{tagGoodDark}{HTML}{1565C0}

\definecolor{tagFair}{HTML}{FFFDE7}
\definecolor{tagFairDark}{HTML}{F9A825}

\definecolor{tagPoor}{HTML}{FFF3E0}
\definecolor{tagPoorDark}{HTML}{EF6C00}

\definecolor{tagBad}{HTML}{FFEBEE}
\definecolor{tagBadDark}{HTML}{C62828}

\newcommand{\badge}[3]{
  \tikz[baseline=-0.6ex]\node[
    fill=#1, draw=#2, text=#2,
    rounded corners=2pt,
    inner xsep=4pt, inner ysep=2pt,
    font=\fontsize{6.5}{7}\selectfont\bfseries\sffamily
  ]{#3};}

\newcommand{\tagbox}[1]{%
  \ifstrequal{#1}{ENTITY}      {\badge{tagEntity}{tagEntityDark}{ENTITY}}{}%
  \ifstrequal{#1}{ENTITY\_TRAIT}{\badge{tagEntity}{tagEntityDark}{ENTITY\_TRAIT}}{}%
  \ifstrequal{#1}{ATTRIBUTE\_BINDING}{\badge{tagEntity}{tagEntityDark}{ATTR\_BIND}}{}%
  \ifstrequal{#1}{ENTITY\_RELATION}  {\badge{tagEntity}{tagEntityDark}{ENTITY\_REL}}{}%
  \ifstrequal{#1}{ANNOTATION}        {\badge{tagAnnotation}{tagAnnotationDark}{ANNOT}}{}%
  \ifstrequal{#1}{ANNOTATION\_BINDING}{\badge{tagAnnotation}{tagAnnotationDark}{ANNOT\_BIND}}{}%
  \ifstrequal{#1}{ANNOTATION\_RELATION}{\badge{tagAnnotation}{tagAnnotationDark}{ANNOT\_REL}}{}%
  \ifstrequal{#1}{VISUAL}    {\badge{tagVisual}{tagVisualDark}{VISUAL}}{}%
  \ifstrequal{#1}{KNOWLEDGE} {\badge{tagKnowledge}{tagKnowledgeDark}{KNOWLEDGE}}{}%
  \ifstrequal{#1}{REQUIRED}  {\badge{tagRequired}{tagRequiredDark}{REQUIRED}}{}%
  \ifstrequal{#1}{OPTIONAL}  {\badge{tagOptional}{tagOptionalDark}{OPTIONAL}}{}%
  \ifstrequal{#1}{ADDITIONAL}{\badge{tagAdditional}{tagAdditionalDark}{ADDITIONAL}}{}%
  \ifstrequal{#1}{5\_EXCELLENT}{\badge{tagExcl}{tagExclDark}{5: EXCELLENT}}{}%
  \ifstrequal{#1}{4\_GOOD}     {\badge{tagGood}{tagGoodDark}{4: GOOD}}{}%
  \ifstrequal{#1}{3\_FAIR}     {\badge{tagFair}{tagFairDark}{3: FAIR}}{}%
  \ifstrequal{#1}{2\_POOR}     {\badge{tagPoor}{tagPoorDark}{2: POOR}}{}%
  \ifstrequal{#1}{1\_BAD}      {\badge{tagBad}{tagBadDark}{1: BAD}}{}%
}

\newcommand{\groundingimg}[1]{%
  \ifx\relax#1\relax
    ---
  \else
    \includegraphics[width=2cm,height=2cm,keepaspectratio]{#1}%
  \fi
}

\definecolor{tocblue}{RGB}{0, 0, 128} 

\hypersetup{
    colorlinks=true,
    linkcolor=tocblue,
    citecolor=tocblue,
    urlcolor=tocblue,
    pdfborder={0 0 0}
}

\titlecontents{section}
  [0em] 
  {\vspace{0.65em}\bfseries\color{tocblue}}
  {\contentslabel{2.0em}}
  {}
  {\hfill\contentspage}

\titlecontents{subsection}
  [2.8em] 
  {\color{tocblue}}
  {\contentslabel{3.0em}}
  {}
  {\titlerule*[0.55em]{.}\contentspage}

\title{{\benchmark}: Dual-View Evaluation and Diagnosis of World-Knowledge Grounding in Text-to-Image Models}

\author{%
  \normalfont
  \textbf{Yonghang Yu}\thanks{These authors contributed equally.}\quad
  \textbf{Ruoyu Wang}\footnotemark[1]\quad
  \textbf{Haoyu Zhang}\\
  \textbf{Lin Qi}\quad
  \textbf{Yawei Zhou}\quad
  \textbf{Changqian Yu}\\[0.5em]
  KlingAI Research
}

\begin{document}

\maketitle


\begin{abstract}
Text-to-image (T2I) models have made substantial progress in visual realism, aesthetic quality, and instruction following. However, real-world prompts often go beyond explicit visual descriptions and require implicit facts, structured knowledge, and domain-specific commonsense. Existing evaluations mainly focus on explicit prompt-to-image semantic alignment or specific knowledge subdomains, lacking unified evaluation and diagnosis of world-knowledge grounding in T2I models. We introduce \textbf{KnowVis} for dual-view evaluation and diagnosis of \textbf{world-knowledge grounding} in T2I generation. Built on a hierarchical knowledge taxonomy with case-level annotations, KnowVis combines two complementary benchmark views with a structured MLLM-based VQA judge protocol. The Skill-Tree Benchmark supports fine-grained atomic knowledge evaluation, while the Generalized Benchmark examines open-ended knowledge composition and expression. The protocol uses Required VQAs to assess atomic knowledge correctness in the Skill-Tree Benchmark, and combines Required, Optional, and dynamically discovered knowledge expressions with human–judge collaborative evaluation in the Generalized Benchmark. We validate the protocol’s reliability on the Skill-Tree validation split, enabling scalable evaluation and diagnostic error attribution. Experiments show that KnowVis reveals notable gaps in world-knowledge grounding across existing T2I models and provides structured diagnostic signals for knowledge-intensive image generation.
\end{abstract}
\section{Introduction}
\label{sec:introduction}

Text-to-image (T2I) generation models have made substantial progress in visual realism, aesthetic quality, and instruction following.
Early systems such as DALL·E 2, Imagen, and Stable Diffusion established the modern T2I paradigm \citep{ramesh2022hierarchical,saharia2022photorealistic,rombach2022high}, while recent multimodal image generation systems and image-capable foundation models have further improved complex prompt following, text rendering, and practical visual content generation
\citep{openai2025gpt4oimage,seedream2025seedream4,qwen2025image}.
Despite these advances, many widely used evaluation protocols still primarily assess whether generated images correctly render the objects, attributes, and relations explicitly specified in the prompt \citep{hu2023tifa,lin2024evaluating,cho2024dsg}.

However, real-world prompts often require more than surface-level prompt rendering.
A prompt may implicitly require factual knowledge, physical commonsense, symbolic conventions, domain rules, cultural knowledge, or reasoning over unstated constraints.
In such cases, the challenge is not only to ``draw what is mentioned,'' but also to ``draw it correctly'': the generated image should satisfy the relevant world-knowledge constraints required by the prompt.
Recent studies have shown that T2I models can still fail on such knowledge-intensive tasks, including physical commonsense, factual concept generation, and beyond-prompt semantic consistency \citep{meng2024phybench,huang2025t2ifactual,huang2024kitten,niu2025wise,zhang2025worldgenbench,zhang2025align}.
These failures suggest that, in knowledge-grounded generation, errors cannot always be fully explained by generic image quality or surface-level text-image alignment alone.
We refer to this capability as \emph{world-knowledge grounding}: the ability to identify, retrieve, reason over, and visually express explicit or implicit knowledge constraints in a visually verifiable manner.
A formal definition and task boundary are provided in Appendix~\ref{app:task_scope_and_definitions}.

Existing work has approached this problem from several directions, including compositional and relation-binding evaluation \citep{huang2023t2i,cho2024dsg}, physical commonsense evaluation \citep{meng2024phybench}, factual and knowledge-intensive generation \citep{huang2025t2ifactual,huang2024kitten}, and beyond-prompt or world-knowledge-aware evaluation \citep{niu2025wise,zhang2025worldgenbench,zhang2025align}.
While these studies provide important insights, they often focus on specific knowledge types, task formats, or evaluation protocols.
There remains a lack of a unified framework that jointly addresses three questions:
\emph{what} constitutes world-knowledge grounding in T2I generation,
\emph{how} to cover both atomic knowledge verification and open-ended knowledge application,
and \emph{how} to evaluate such capabilities with reliable and diagnostic intermediate signals.
A more detailed comparison with prior work is provided in Appendix~\ref{app:related_work}.

\begin{figure*}[t]
    \centering
    \includegraphics[
        width=0.98\textwidth
    ]{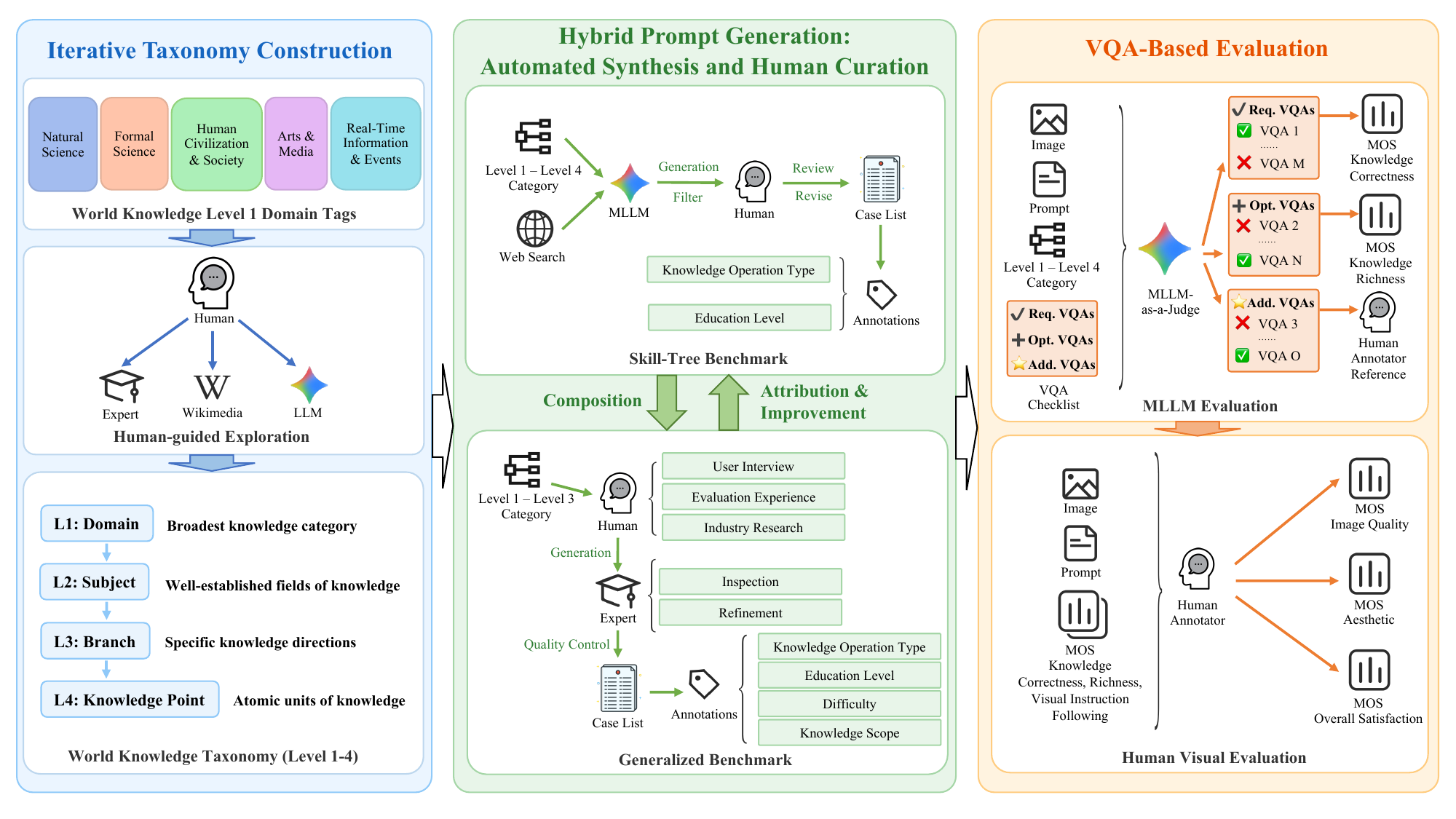}
    \caption{
Overview of the {\benchmark} framework.
It includes iterative taxonomy construction, dual-view benchmark construction, and structured VQA-based evaluation.
The Skill-Tree Benchmark focuses on L4-level atomic knowledge diagnosis, while the Generalized Benchmark targets open-ended knowledge-grounded generation with human--judge collaborative evaluation.
}
    \label{fig:knowvis_framework}
\end{figure*}

To address these issues, we introduce \textbf{\benchmark}, a dual-view diagnostic benchmark for evaluating world-knowledge grounding in T2I models.
As summarized in Figure~\ref{fig:knowvis_framework}, {\benchmark} combines an L1--L4 hierarchical taxonomy, two complementary benchmark views, and a structured VQA-based MLLM-as-a-Judge protocol.
The Skill-Tree Benchmark evaluates controlled L4-level atomic knowledge expression, while the Generalized Benchmark targets open-ended knowledge-grounded generation with human--judge collaborative evaluation.

Building on this design, our main contributions are as follows:
\begin{itemize}[leftmargin=*, itemsep=0.45em, topsep=0.3em, parsep=0pt, partopsep=0pt]
    \item \textbf{A unified task formulation.}
    We formalize world-knowledge grounding in T2I generation and define its boundary against general text-image alignment, image quality evaluation, and surface-level instruction following.

    \item \textbf{A dual-view diagnostic benchmark.}
    We construct \textbf{\benchmark}, consisting of a Skill-Tree Benchmark for atomic knowledge verification and a Generalized Benchmark for open-ended knowledge application, both organized by a shared hierarchical taxonomy.

    \item \textbf{A structured and attributable evaluation protocol.}
    We propose a VQA-based MLLM-as-a-Judge protocol shared across benchmark views, enabling fine-grained assessment of knowledge correctness, knowledge richness, and diagnostic failure patterns.
\end{itemize}

\section{Benchmark Framework}
\label{sec:benchmark_framework}

\subsection{Taxonomy and Dual-View Benchmark Design}
\label{sec:taxonomy_dual_view}

We define \emph{world-knowledge grounding} in T2I generation as the ability of a model to identify, retrieve, reason over, and visually express explicit or implicit knowledge constraints required by a prompt.
Such constraints may involve factual states, physical laws, symbolic conventions, domain rules, structured relations, or culturally grounded concepts that are not fully specified by surface-level visual descriptions.
This differs from general text--image alignment: a prompt such as ``a red sphere on a table'' mainly tests whether explicitly stated objects, attributes, and spatial relations are rendered, whereas a prompt such as ``a convex lens converges parallel light rays'' requires the generated image to satisfy an underlying optical rule.
We provide the formal task definition and boundary in Appendix~\ref{app:task_scope_and_definitions}.

To organize this capability into a diagnosable evaluation space, we construct an L1-L4 world-knowledge taxonomy.
L1/L2 define broad domains and stable subjects, L3 represents scenario-forming knowledge branches, and L4 denotes atomic knowledge points with clear boundaries and stable evaluation criteria.
L4 serves as the basis for controlled sample construction, VQA decomposition, and fine-grained attribution, while more subjective or time-sensitive domains are retained at coarser L2/L3 granularity.
The taxonomy hierarchy and construction criteria are detailed in Appendix~\ref{app:world_knowledge_designing_taxonomy} and Appendix~\ref{subsec:taxonomy_construction_criteria}.

Based on this taxonomy, we construct two complementary benchmark views.
The \textbf{Skill-Tree Benchmark} is centered on L4 atomic knowledge points and evaluates whether a model can correctly express specific knowledge constraints under controlled settings.
The \textbf{Generalized Benchmark} is centered on L3 knowledge branches and evaluates whether a model can apply relevant knowledge in more realistic, open-ended, and multi-knowledge generation scenarios.
Together, the two views support diagnosis from atomic knowledge verification to open-ended knowledge application, while their shared taxonomy enables cross-view analysis and failure attribution.

\subsection{Benchmark Construction Pipeline}
\label{sec:benchmark_construction_pipeline}

The two benchmark views follow a strategy of \emph{separate construction with unified attribution}.
Skill-Tree cases are constructed around L4 atomic knowledge points through LLM-based candidate generation, automatic filtering, and human screening, so that each retained case has a focused target and clear assessment criteria.
Generalized cases are manually curated around L3 branches to preserve realistic, open-ended, and compositional generation scenarios, and are then annotated with potentially involved L4 points for cross-view attribution.
This mapping provides partial failure attribution and helps identify taxonomy coverage gaps without assuming that every open-ended failure can be fully explained by existing L4 nodes.
Full construction details are provided in Appendix~\ref{app:benchmark_construction_details}.

\subsection{Structured Metadata and Benchmark Statistics}
\label{sec:structured_tags_statistics}

To support coverage analysis, capability slicing, and error attribution, we annotate each case with structured metadata.
At the knowledge-structure level, Skill-Tree samples are mapped to a single L1--L4 taxonomy path, with L4 as the primary atomic knowledge point.
Generalized samples are typically annotated at the L3 branch level, with an additional set of potentially involved L4 points; when stable L3/L4 decomposition is not available, a case may be annotated at the L2 level.

\begin{figure}[t]
    \centering
    
    \begin{minipage}[t]{0.46\linewidth}
        \centering
        \captionof{table}{Overview of the two benchmark view.}
        \label{tab:benchmark_overview}
        
        \scriptsize
        \setlength{\tabcolsep}{4.5pt}
        \renewcommand{\arraystretch}{1}
        \setlength{\extrarowheight}{0.8pt}
        
        \begin{tabular}{
            >{\raggedright\arraybackslash}p{0.26\linewidth}
            >{\centering\arraybackslash}p{0.27\linewidth}
            >{\centering\arraybackslash}p{0.36\linewidth}
        }
            \toprule
            \textbf{Statistic} & \textbf{Skill-Tree} & \textbf{Generalized} \\
            \midrule
            
            \#Cases 
            & 1,622 
            & 550 \\
            \midrule
            
            Main unit 
            & \begin{tabular}[c]{@{}l@{}}L4 atomic\\point\end{tabular}
            & L3 branch \\
            \midrule
            
            \begin{tabular}[c]{@{}l@{}}Taxonomy\\coverage\end{tabular}
            & \begin{tabular}[c]{@{}c@{}}3 L1 domains\\240 L4 points\end{tabular}
            & \begin{tabular}[c]{@{}c@{}}5 L1 domains\\20 L2 subjects\\70 L3 branches\end{tabular} \\
            \midrule
            
            Density 
            & \begin{tabular}[c]{@{}c@{}}6.8 cases\\per L4\end{tabular}
            & -- \\
            \midrule
            
            \begin{tabular}[c]{@{}l@{}}Knowledge\\operation\end{tabular}
            & 3 types
            & 4 types \\
            \midrule
            
            Extra metadata 
            & \begin{tabular}[c]{@{}c@{}}Education\\level\end{tabular}
            & \begin{tabular}[c]{@{}c@{}}Education level\\Difficulty\\Knowledge quantity\end{tabular} \\
            
            \bottomrule
        \end{tabular}
    
    \end{minipage}
    \hfill
    \begin{minipage}[t]{0.5\linewidth}
        \centering
        \vspace{0.35em}
        \includegraphics[width=0.75\linewidth]{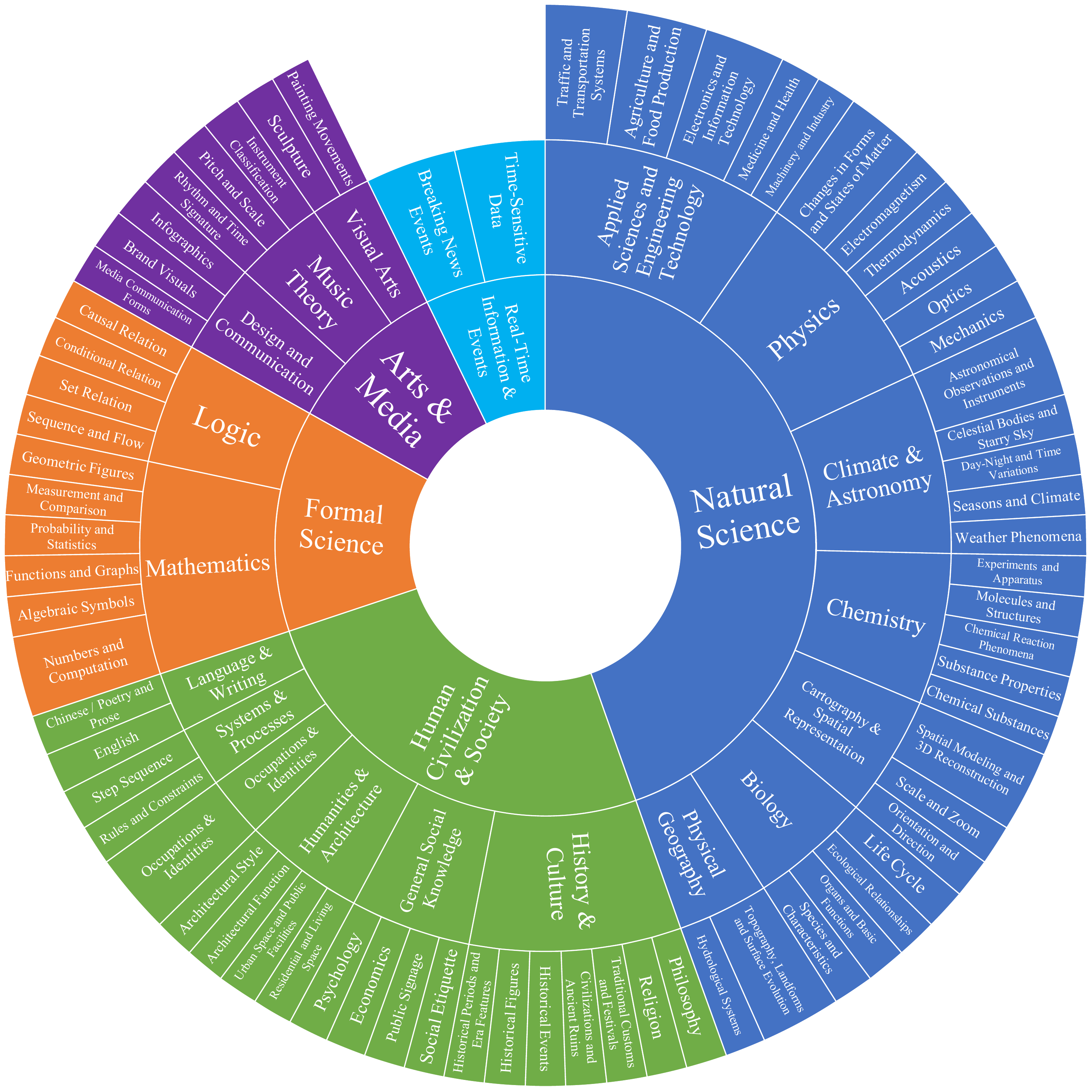}
        \captionof{figure}{Taxonomy coverage, shown up to the L3 branch level for legibility.}
        \label{fig:taxonomy_sun_burst_l3}
    \end{minipage}

\end{figure}

At the task-attribute level, we annotate education level, knowledge operation type, and, for Generalized cases, difficulty.
Education level indicates the typical learning stage or knowledge source associated with a case.
Knowledge operation type describes the primary way in which knowledge is used in the generation task, including Explicit Knowledge Rendering, Explicit Knowledge Recall, Implicit Knowledge Recall, and Knowledge Reasoning.
When multiple operation types are involved, we annotate the primary and highest-level knowledge use.
The Skill-Tree Benchmark uses three operation types and does not annotate difficulty, while the Generalized Benchmark uses the full four-way operation schema and additionally annotates difficulty to capture the combined complexity of semantic understanding and visual presentation.The full metadata schema is defined in Appendix ~\ref{subsec:case_level_metadata_schema}.
\paragraph{Benchmark statistics and scope.}
Table~\ref{tab:benchmark_overview} summarizes the scale and construction units of the two benchmark views.
Skill-Tree contains 1,622 cases covering 240 L4 points, while Generalized contains 550 cases covering 20 L2 subjects and 70 L3 branches.
The reported distributions are coverage-driven and evaluability-driven rather than an exhaustive distribution of world knowledge.
Full statistics are reported in Appendix~\ref{subsec:benchmark_statistics}.

\section{Evaluation Protocol and Diagnostic Framework}
\label{sec:evaluation_protocol}

\subsection{Unified VQA-Based MLLM-as-a-Judge Protocol}
\label{sec:unified_vqa_judge_protocol}

Both benchmark views share the same structured judge target: determining whether a generated image satisfies the explicit or implicit world-knowledge constraints required by the prompt.
Rather than asking for a holistic judgment of whether an image is ``knowledge-correct,'' we decompose each case into visually verifiable VQA items.
Each item checks a localized piece of evidence, such as the presence of an entity, the correctness of an attribute, the validity of a relation, the binding between annotations and visual elements, or the satisfaction of a knowledge-dependent structure.
This decomposition converts open-ended knowledge judgment into structured intermediate signals that can be aggregated and analyzed.

We use three judge types.
\textbf{Required VQAs} define the minimal evidence needed for the core knowledge constraint to hold.
\textbf{Optional VQAs} define predefined valid supplementary expressions and are used to measure knowledge richness.
\textbf{ADDITIONAL VQAs} record valid supplementary expressions that are not covered by the current OPTIONAL set. Since outputs from different systems can contain diverse, open-ended valid knowledge expressions, we dynamically extract ADDITIONAL VQAs during evaluation to capture reasonable supplementary content beyond the predefined OPTIONAL pool; these items are used for analysis and for iteratively refining the OPTIONAL set, rather than being directly included in the main richness score.

Each VQA item is further associated with structured fields for stable judgment and attribution.
The \textit{component} field specifies the visual or annotation element being inspected; \textit{knowledge\_hint} provides background knowledge and judgment criteria; and \textit{visual\_grounding} points the judge to relevant visual evidence or reference cues when needed.
Detailed VQA schema definitions and auxiliary annotations for knowledge hints and visual grounding are provided in Appendix~\ref{app:structured_judge_protocol}. In our implementation, we use \verb+Gemini 3.1 Pro Preview+ as the MLLM judge backend.
The judge receives the original prompt, the generated image, and the structured VQA items, and returns a ternary decision for each item, i.e., \texttt{PASS}, \texttt{FAIL}, or \texttt{NA}, together with a brief rationale. Here, \texttt{NA} is used when an optional or additional expression is not applicable or not present in the image, rather than being counted as an incorrect required knowledge constraint.

The protocol design is model-agnostic in the sense that VQA decomposition, schema fields, scoring rules, and aggregation are not tied to a specific MLLM backend. This portability does not imply comparable reliability across judge models with different knowledge, visual perception, and reasoning capabilities. We use Gemini 3.1 Pro Preview as the default judge for the primary benchmark results and assess robustness through a same-setting rerun and alternative strong MLLM judges in Section~\ref{subsec:judge_reliability_validation}. A protocol-level comparison with prior automatic evaluation methods is provided in Appendix~\ref{app:related_work}.

\subsection{Skill-Tree and Generalized Evaluation}
\label{sec:skill_tree_generalized_evaluation}

The Skill-Tree Benchmark is designed for scalable diagnosis over L4 atomic knowledge points.
Because each case targets a bounded knowledge point with a stable expected visual expression, we use automatic VQA-based evaluation and focus on Required VQAs.
A Skill-Tree case is considered correct only when all Required VQAs are satisfied, reflecting whether the generated image correctly expresses the minimal knowledge frame of the target L4 point.
For a model $m$ and a Skill-Tree case $c$, the case-level correctness score is:
\begin{equation}
S_{\mathrm{st}}(m,c)=
\prod_{i=1}^{N_{\mathrm{req}}(c)}
r_i^{m,c},
\qquad
r_i^{m,c} \in \{0,1\},
\label{eq:skill_tree_correctness}
\end{equation}
where $S_{\mathrm{st}}$ denotes the Skill-Tree case-level correctness score, $N_{\mathrm{req}}(c)$ is the number of Required VQAs and $r_i^{m,c}$ indicates whether the $i$-th Required VQA is satisfied.

The Generalized Benchmark targets open-ended scenarios where a prompt may involve multiple knowledge constraints, complex contexts, and multiple plausible expression paths.
We therefore combine structured judge scores with human MOS ratings instead of using a strict pass/fail criterion alone.
Required VQAs are averaged into a knowledge correctness score:
\begin{equation}
S_{\mathrm{kc}}(m,c)
=
\frac{1}{N_{\mathrm{req}}(c)}
\sum_{i=1}^{N_{\mathrm{req}}(c)}
r_i^{m,c},
\qquad
r_i^{m,c}\in\{0,1\}.
\label{eq:generalized_knowledge_correctness}
\end{equation}
Optional VQAs are used to measure knowledge richness. For each optional item, we set
$o_i^{m,c}=1$ if the corresponding supplementary knowledge expression is present and correct.
We set $o_i^{m,c}=0$ if the expression is present but incorrect, or if it is expected but missing.
The knowledge richness score is:
\begin{equation}
S_{\mathrm{kr}}(m,c)
=
\frac{1}{N_{\mathrm{opt}}(c)}
\sum_{i=1}^{N_{\mathrm{opt}}(c)} o_i^{m,c},
\qquad
o_i^{m,c} \in \{0,1\}.
\label{eq:generalized_knowledge_richness}
\end{equation}
Here, $S_{\mathrm{kc}}$ measures the satisfaction of core knowledge requirements, while
$S_{\mathrm{kr}}$ measures the coverage of predefined supplementary knowledge expressions.
For Optional VQAs, both missing expressions and incorrectly expressed knowledge are counted as 0,
because they indicate insufficient knowledge richness.

In addition, the Generalized Benchmark evaluates \textbf{non-knowledge visual instruction following}, which measures explicitly stated visual requirements independent of world knowledge, such as subject count, viewpoint, color, layout, and composition. We compute this dimension using an internal visual-instruction judge that decomposes prompts into weighted atomic visual units; details are provided in Appendix~\ref{app:non_knowledge_if}, Eq.~\ref{eq:non_knowledge_if}.

\begin{wrapfigure}[17]{r}{0.6\linewidth}
    \vspace{-24pt}
    \centering
    \includegraphics[
        width=0.7\linewidth
    ]{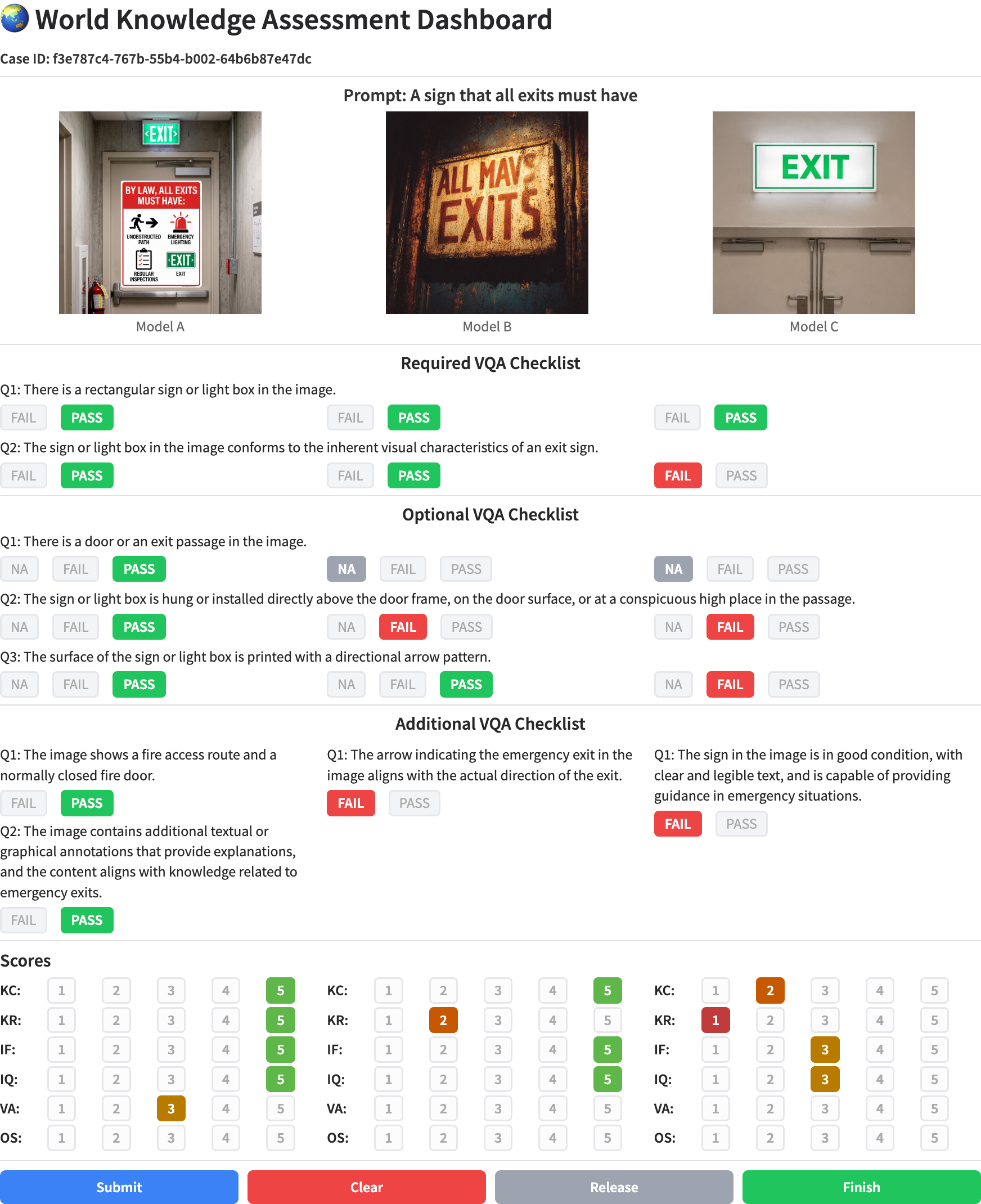}
    \captionof{figure}{Annotation interface.}
    \label{fig:human_judge_interface}
    \vspace{-4pt}
\end{wrapfigure}

Figure~\ref{fig:human_judge_interface} shows the annotation interface
used in the Generalized Benchmark. For cross-dimensional analysis,
judge-derived scores are linearly mapped to the same numerical scale
as human MOS, while image quality, aesthetics, and overall satisfaction
are rated directly by human annotators.

Knowledge correctness and image quality are evaluated separately.
A factual error does not, by itself, lower the image-quality score;
a lower score requires an independently observable visual defect,
such as blurring, structural distortions, or geometric interpenetration.
Overall satisfaction is rated separately as a holistic judgment
that may reflect both factual correctness and visual quality,
rather than being computed as a fixed weighted combination of
the individual dimensions.

Human ratings are collected independently across annotators,
followed by discrepancy review when needed. Detailed scoring
examples are provided in
Table~\ref{tab:example_overall_satisfaction_scoring}, and the
conversion of judge-derived scores to the MOS scale is specified
in Appendix~\ref{subsubsec:judge_derived_mos_conversion}.
Evaluation rubrics, pre-adjudication inter-annotator agreement,
and score aggregation rules are detailed in
Appendix~\ref{app:human_evaluation_protocol_for_the_generalized_benchmark}.

\subsection{Attribution and Diagnostic Analysis}
\label{sec:attribution_diagnostic_analysis}

We combine structured evaluation signals with case-level metadata for attribution. The signals include
Required-VQA pass/fail results, Optional-VQA coverage, non-knowledge instruction-following scores, and human
MOS ratings, while the metadata include taxonomy paths, education level, knowledge operation type, difficulty,
and knowledge quantity. This supports failure localization by both evaluation dimension and knowledge-task
type. Details are provided in Appendix~\ref{subsec:case_level_metadata_schema} and
Appendix~\ref{app:attribution_framework}.

For Skill-Tree, we localize failures at the L4 level by aggregating Required-VQA failures across component types
and L1--L4 taxonomy levels, with separate tracking of annotation-related failures. For Generalized, we analyze
knowledge correctness, richness, instruction following, and human MOS across L2/L3 branches and task metadata
to identify open-ended failure patterns. 

\section{Experiments}
\label{sec:experiments}

We evaluate {\benchmark} along three axes: judge reliability and robustness,
agreement gains from structured auxiliary fields, and whether the dual-view
benchmark reveals world-knowledge grounding failures beyond general T2I capability.
We report human--judge agreement, cross-judge and human-correction analyses,
component ablations, automatic-metric comparisons, and model performance on
the Skill-Tree and Generalized benchmarks.

\subsection{Experimental Setup}
\label{subsec:experimental_setup}

We use three evaluation sets with complementary roles. For Skill-Tree, we sample 480 cases from the full
1,622-case benchmark, covering all 240 L4 atomic knowledge points. We split them into 240 visible development
cases for judge-prompt iteration and 240 unseen validation cases for model evaluation and ranking comparison
across judges. Human--judge agreement and cross-judge agreement use the full 480-case subset.

For the Generalized Benchmark, we evaluate a 100-case subset of the full 550-case benchmark, selected to cover all L2 subjects and L3 branches. The evaluation follows a human–judge collaborative protocol: the judge assesses knowledge correctness, knowledge richness, and non-knowledge visual instruction following, while human annotators rate image quality, aesthetics, and overall satisfaction. This subset supports analysis of broad model-level trends and representative failure patterns; its limited size does not support precise branch-level rankings or strong conclusions about small differences between models.
For General-Domain T2I, we use a 100-case subset from a 600-case
pool as a control for general generation capability.

We evaluate overlapping groups of general-purpose, aesthetics-oriented, and text-/knowledge-oriented T2I models.

Each case includes bilingual prompts. For each model, we use the language setting that best matches its documented or observed prompting behavior, while keeping the underlying case, semantic intent, metadata, and evaluation protocol fixed across models. This design reduces language-specific confounds, but we do not interpret the results as a controlled comparison of Chinese-versus-English prompting. Further details are in Appendix~\ref{subsec:experimental_setup_details} and
Appendix~\ref{app:resources_and_reproducibility}.

\subsection{Judge Reliability and Robustness}
\label{subsec:judge_reliability_validation}

We first validate Gemini 3.1 Pro Preview, our default judge, against human
annotations on the Skill-Tree and Generalized benchmarks. We then examine
human--judge agreement and model-ranking stability under repeated evaluation
and alternative judge models.
Precision and recall throughout the judge-reliability and ablation analyses treat \texttt{FAIL} as the positive class: precision measures the correctness of flagged errors, and recall measures the fraction of human-labeled errors detected.

\paragraph{Default-judge reliability on the Skill-Tree Benchmark.}
We evaluate the default judge on 480 generated images corresponding to 480 distinct Skill-Tree prompts, with approximately 80 images from each of six T2I models, at two levels: VQA-level agreement measures whether individual checklist judgments
match human labels, while case-level agreement measures whether the aggregated
pass/fail decision matches the human reference. The default judge achieves
94.51\% VQA-level accuracy and 88.12\% case-level accuracy, with Cohen's
$\kappa$ of 0.8236 and 0.7562, respectively. VQA-level precision and recall
are 92.07\% and 80.22\%. These results support the reliability of the
structured judge on the evaluated set, while indicating that some judgment
errors remain.

To assess the reliability of the human reference labels used for judge
validation, five annotators independently labeled 288 Required VQAs from
50 sampled Skill-Tree cases. Inter-annotator agreement reaches a nominal
Krippendorff's $\alpha$ of 0.766, with unanimous agreement on 82.64\% of
VQAs. These results support the consistency of the human references against
which judge performance is assessed. Detailed annotation procedures and
agreement distributions are provided in
Appendix~\ref{app:vqa_gt_annotation_interface}.

\paragraph{Validation in open-ended scenarios.}
We additionally assess the default judge on 50 sampled cases from the
Generalized Benchmark, evaluating Required and Optional VQAs separately.
VQA-level accuracy against human annotations reaches 98.41\% for Required
VQAs and 98.53\% for Optional VQAs. These results show high human--judge
agreement for both knowledge correctness and knowledge richness on the
sampled open-ended cases. Detailed human--judge agreement results for the
Generalized Benchmark are reported in
Appendix~\ref{app:generalized_judge_agreement}.

\paragraph{Contribution of auxiliary fields.}
We further ablate taxonomy context (\texttt{category}), knowledge hints
(\texttt{knowledge\_hint}), and visual grounding cues
(\texttt{visual\_grounding}) while keeping models, samples, VQAs, scoring
rules, and aggregation fixed. Removing any auxiliary field reduces accuracy
against human annotations. Removing all three fields lowers case-level and
VQA-level accuracy by 2.08 and 2.02 percentage points, respectively. The full
ablation results are reported in
Appendix~\ref{app:judge_reliability_ablation}.

\paragraph{Cross-judge validation and ranking robustness.}
To examine dependence on the default judge, we compare the original
evaluation with a same-setting Gemini 3.1 Pro Preview rerun and
evaluations using Gemini 3.5 Flash and GPT 5.5.
Table~\ref{tab:cross_judge_reliability} summarizes VQA-level agreement
with human annotations and ranking consistency across the ten
evaluated T2I models.

Across all tested settings, VQA-level accuracy remains above 91\%.
Although the rerun and alternative judges yield lower agreement
metrics than the original evaluation, they retain substantial
agreement with human annotations, with Cohen's $\kappa$ ranging
from 0.7140 to 0.7475 and accuracy ranging from 91.33\% to 92.21\%.
These results indicate that agreement with human judgments remains
at a comparatively high level despite variation across evaluations.

Model rankings also remain highly consistent with the original
ranking: Kendall's $\tau_b$ is 0.8222, and Spearman's $\rho$ ranges
from 0.9273 to 0.9394. The top two models remain unchanged across
all settings; Gemini 3.5 Flash and GPT 5.5 also preserve the
original top-three ordering. No model moves by more than two
positions. Together, the maintained human--judge agreement and
stable ranking trends support the robustness of the protocol
for comparative model evaluation under the tested strong MLLM
judge settings. Complete case-level agreement results and
per-model rankings are reported in
Appendices~\ref{app:cross_judge_agreement}
and~\ref{app:cross_judge_ranking}, respectively.

We further assess the impact of judge errors through a
human-correction audit of a total of 50 prompt--image pairs sampled
from the Skill-Tree outputs of five T2I models.
Correcting the identified errors preserves the
top-ranked model and only swaps the second and third positions,
yielding Kendall's $\tau$ of 0.8000 and Spearman's $\rho$ of
0.9000 between the original and corrected rankings.
These results provide additional evidence of ranking stability
within the audited subset. Detailed human-correction results
are reported in Appendix~\ref{app:judge_error_ranking_impact};
judge failure modes are analyzed in
Appendix~\ref{app:empirical_analysis_of_automatic_evaluation_failure}.

\begin{table}[t]
\centering
\small
\caption{Human--judge agreement and ranking consistency across judge settings.
Accuracy, precision, and recall are reported as percentages; $\kappa$ denotes
Cohen's $\kappa$. Ranking correlations are computed relative to the model
ranking obtained with the original Gemini 3.1 Pro Preview evaluation.
Higher values indicate stronger agreement or ranking consistency.}
\label{tab:cross_judge_reliability}
\begingroup
\renewcommand{\arraystretch}{1.12}
\setlength{\tabcolsep}{4pt}
\resizebox{\linewidth}{!}{%
\begin{tabular}{lrrrrrr}
\toprule
& \multicolumn{4}{c}{\textbf{VQA-level human--judge agreement}}
& \multicolumn{2}{c}{\textbf{Ranking consistency}} \\
\cmidrule(lr){2-5}\cmidrule(lr){6-7}
\textbf{Judge setting}
& \textbf{Acc.} $\uparrow$
& \textbf{Prec.} $\uparrow$
& \textbf{Rec.} $\uparrow$
& $\boldsymbol{\kappa}\uparrow$
& \textbf{Kendall's} $\tau_b\uparrow$
& \textbf{Spearman's} $\rho\uparrow$ \\
\midrule
Gemini 3.1 Pro Preview (original)
& 94.51 & 92.07 & 80.22 & 0.8236 & -- & -- \\
Gemini 3.1 Pro Preview (rerun)
& 91.74 & 83.63 & 74.44 & 0.7366 & 0.8222 & 0.9273 \\
Gemini 3.5 Flash
& 92.21 & 86.74 & 73.41 & 0.7475 & 0.8222 & 0.9394 \\
GPT 5.5
& 91.33 & 86.03 & 69.11 & 0.7140 & 0.8222 & 0.9394 \\
\bottomrule
\end{tabular}%
}
\endgroup
\end{table}

\subsection{Comparison with Automatic Metrics}
\label{subsec:metric_comparison}

We compare our judge-derived VQA pass rate with CLIPScore and VQAScore, using the human-annotated VQA pass rate as the reference. CLIPScore represents embedding-based text--image alignment, while VQAScore represents VQA-based alignment. We report Kendall's $\tau$ and Spearman's $\rho$ across L1 domains, and visualize L2-level consistency with human labels. Other preference- or benchmark-specific metrics are discussed qualitatively in Appendix~\ref{app:related_work}, as their task definitions, inputs, and scoring protocols differ substantially.

\begin{figure}[t]
\centering
\small
\renewcommand{\arraystretch}{1.90}
\setlength{\tabcolsep}{3.5pt}

\begin{minipage}[t]{0.58\textwidth}
\centering
\textbf{(a) L1-domain correlation with human GT VQA pass rate.}

\vspace{4pt}

\resizebox{\linewidth}{!}{
\begin{tabular}{
>{\raggedright\arraybackslash}p{0.15\textwidth}
cc
cc
cc
}
\toprule
\textbf{Method}
& \multicolumn{2}{c}{\textbf{Natural Sci.}}
& \multicolumn{2}{c}{\textbf{Formal Sci.}}
& \multicolumn{2}{c}{\textbf{Human Civ.}} \\
\midrule
&
$\tau(\uparrow)$ & $\rho(\uparrow)$
& $\tau(\uparrow)$ & $\rho(\uparrow)$
& $\tau(\uparrow)$ & $\rho(\uparrow)$ \\
\midrule
CLIPScore
& 0.07 & 0.11
& -0.09 & -0.12
& 0.16 & 0.22 \\
\midrule
VQAScore
& 0.16 & 0.25
& 0.03 & 0.04
& 0.14 & 0.19 \\
\midrule
\textbf{Ours}
& \textbf{0.74} & \textbf{0.88}
& \textbf{0.76} & \textbf{0.88}
& \textbf{0.85} & \textbf{0.95} \\
\bottomrule
\end{tabular}
}
\end{minipage}
\hfill
\begin{minipage}[t]{0.38\textwidth}
\centering

\vspace{2pt}

\includegraphics[
    width=1\linewidth]{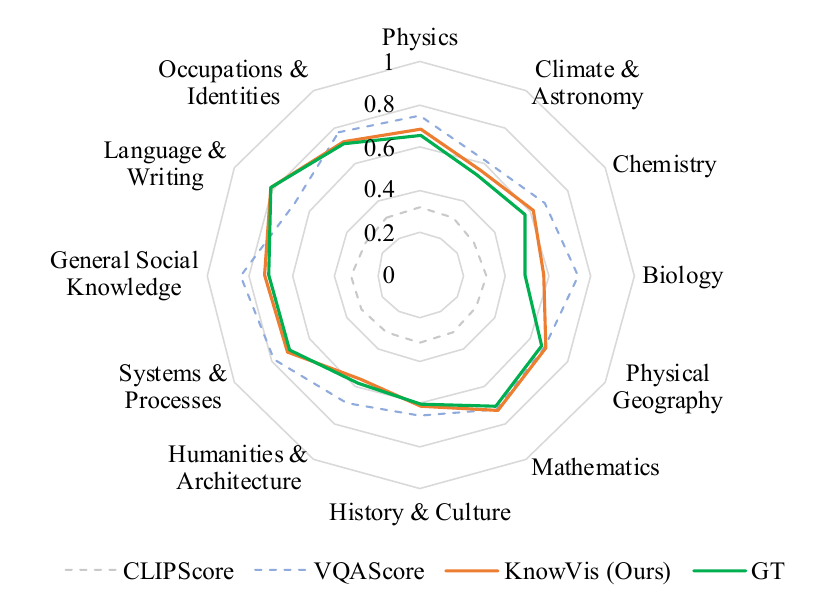}
    \textbf{(b) L2-level consistency pattern.}
\end{minipage}

\caption{
Quantitative comparison with existing automatic metrics. 
(a) Correlation between automatic metrics and human GT VQA pass rate across three L1 knowledge domains. 
(b) L2-level consistency pattern across 12 subjects, where {\benchmark} more closely follows the human GT profile than CLIPScore and VQAScore.
}
\label{fig:metric_comparison_clip_score_vqa_score}
\end{figure}

As shown in Figure~\ref{fig:metric_comparison_clip_score_vqa_score}, our protocol achieves the strongest correlation with human knowledge labels across all three L1 domains. The gap is most evident in Formal Science, where CLIPScore is negatively correlated with human labels. The L2-level radar plot further shows that structured VQA judgments better preserve human-observed domain-wise patterns, whereas general alignment metrics are less stable across knowledge branches. These results suggest that general text--image alignment metrics are insufficient for world-knowledge correctness evaluation, and motivate our use of knowledge-specific VQA judgments for scalable Skill-Tree evaluation.

\subsection{Benchmark Results and Diagnostic Analysis}
\label{subsec:benchmark_results_analysis}

We report model performance on the Skill-Tree and Generalized benchmarks.
The goal is not only to compare overall scores, but also to examine whether the benchmark exposes domain- and task-level failure patterns.
\paragraph{Skill-Tree Benchmark.}
\begin{figure}[t]
\centering

\begin{minipage}[t]{0.49\linewidth}
\centering
\includegraphics[
    height=0.20\textheight,
    keepaspectratio,
    trim=15 0 0 0,
    clip
]{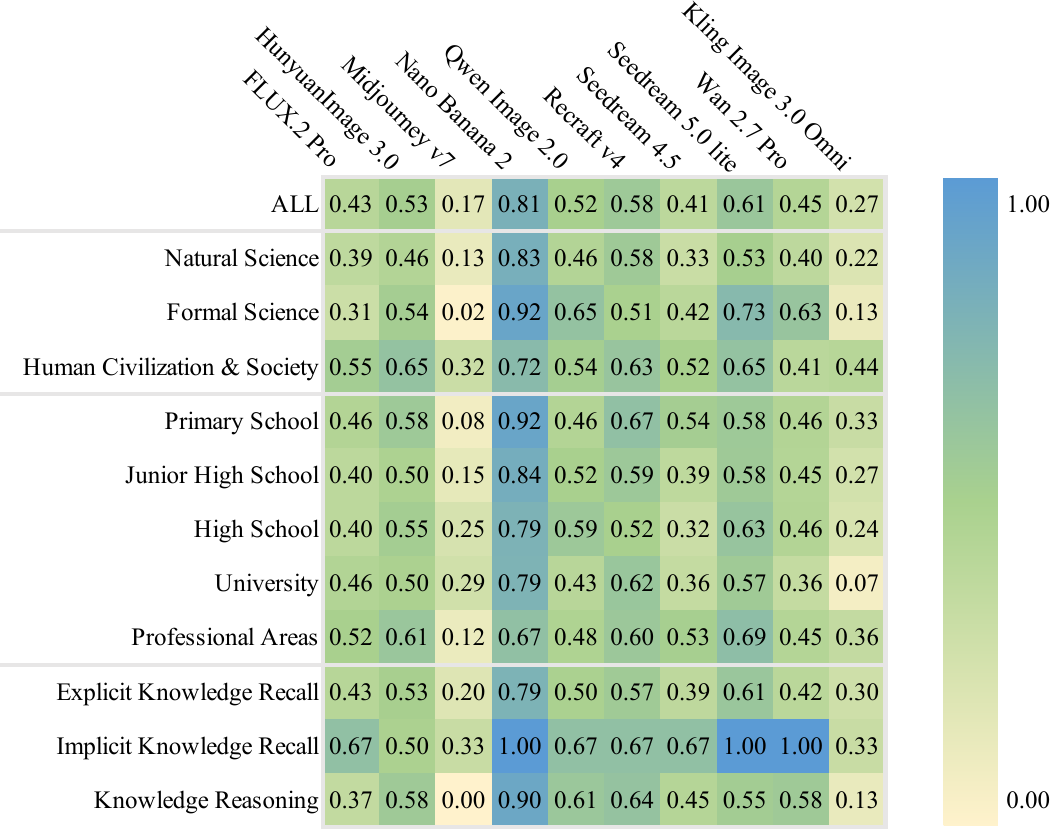}

\vspace{0.3em}
\textbf{(a) Case-level correctness.}
\end{minipage}
\hspace{0.005\linewidth}
\begin{minipage}[t]{0.49\linewidth}
\centering
\includegraphics[
    height=0.20\textheight,
    keepaspectratio,
    trim=15 0 0 0,
    clip
]{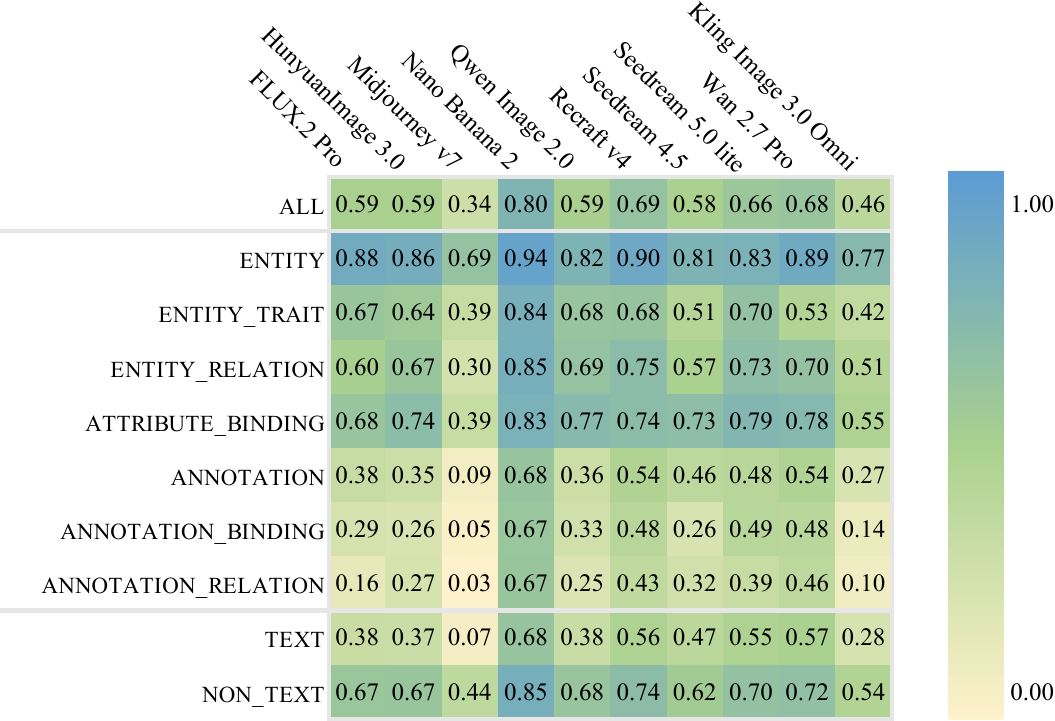}

\vspace{0.3em}
\textbf{(b) VQA component pass rates.}
\end{minipage}

\vspace{0.5em}
\caption{
Skill-Tree diagnostic heatmaps.
(a) Case-level correctness aggregated by taxonomy and task metadata.
(b) VQA-level pass rates across component types and models.
Together, the two views show both where knowledge failures concentrate and which structured VQA components are more challenging.
}
\label{fig:skill_tree_diagnostic_heatmaps}
\end{figure}

Figure~\ref{fig:skill_tree_diagnostic_heatmaps} summarizes Skill-Tree results from two complementary views:
case-level correctness and VQA-level component pass rates. The case-level results show clear model tiers:
Nano Banana 2 performs best, several models fall into a middle range, and Midjourney V7 and {\klingimageomnithree}
are relatively weaker on Skill-Tree knowledge correctness. The L1 slices further reveal domain-specific variation
across Natural Science, Formal Science, and Human Civilization \& Society, indicating that world-knowledge
grounding cannot be assessed by an overall score alone.

Education level and knowledge operation type provide complementary diagnostic views. Education level should
be interpreted as the typical learning stage rather than a direct proxy for generation difficulty, while Knowledge
Reasoning more clearly separates weaker models than recall-oriented categories. The VQA-level results further
show that entity-related checks are generally easier than relation- and annotation-related checks, suggesting
that models often generate relevant objects but fail to express structured knowledge through relations, labels,
formulas, arrows, or text--image bindings. A more complete heatmap including L2-level subjects is provided in Figure~\ref{fig:full_skill_tree_and_generalized_heatmaps}(a);
detailed VQA-level model failure analysis is provided in Appendix~\ref{app:vqa_component_failure_analysis}.

\paragraph{Generalized Benchmark.}
The Generalized Benchmark evaluates open-ended world-knowledge generation through both judge-derived knowledge dimensions and human MOS dimensions.
\begin{figure}[t]
    \centering
    \includegraphics[width=\linewidth]{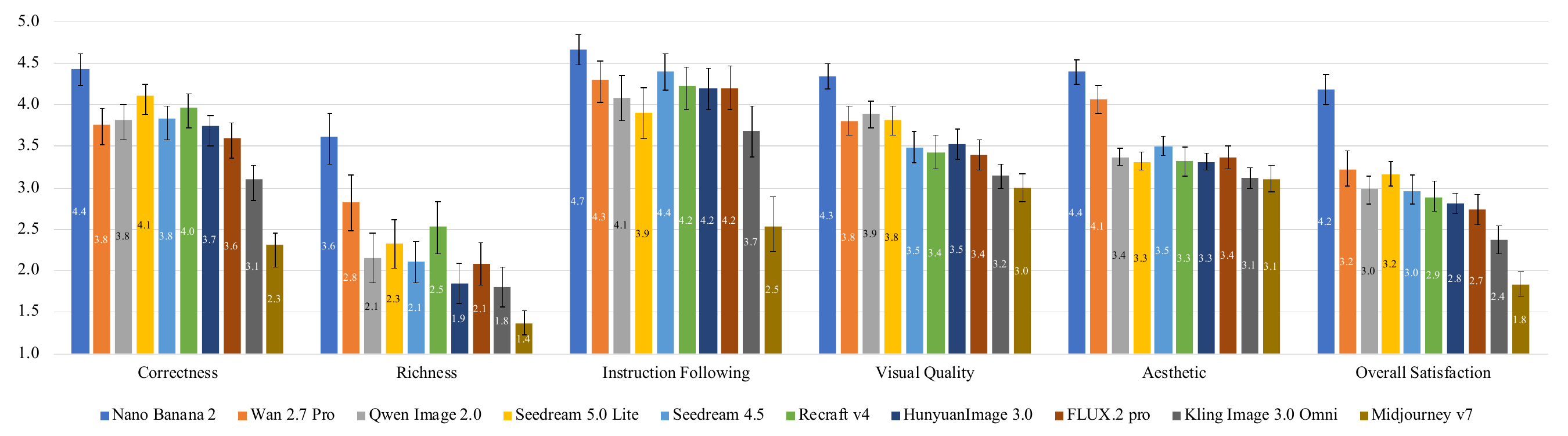}
    \caption{Mean Opinion Score (MOS) of different models across evaluation dimensions.}
    \label{fig:generalized_2d_bar}
\end{figure}
As shown in Figure~\ref{fig:generalized_2d_bar}, Nano Banana 2 leads on multiple dimensions, including
knowledge correctness, knowledge richness, instruction following, and overall satisfaction. This suggests
that its advantage is not merely driven by perceptual quality, but by stronger performance across both
knowledge-centered and general visual dimensions. In contrast, Midjourney V7, despite strong aesthetics,
obtains lower knowledge correctness and richness, showing that aesthetic quality does not necessarily imply
world-knowledge grounding capability.

We also observe consistent domain-level patterns between the two benchmark views. As shown in Figure~\ref{fig:full_skill_tree_and_generalized_heatmaps}(b), models with
weak Skill-Tree performance in Formal Science also tend to obtain lower Generalized MOS in corresponding
formal-science cases. These observations suggest that L4-level weaknesses can surface in open-ended generation,
although the relationship should not be interpreted as a strict causal mapping.

Complementary General-Domain T2I results are reported in Appendix~\ref{app:general_domain_t2i}.
They show that general instruction following, visual quality, and aesthetics are not fully aligned with world-knowledge grounding, supporting the need for dedicated knowledge-grounded evaluation.

\section{Discussion}
\label{sec:discussion}

To examine whether explicit factual guidance improves generation, we conduct a prompt rewriting experiment on 240 unseen validation cases, converting the original prompts into direct visual instructions that supply the relevant factual details. Across the selected cases, the pass rate increases from 0.271 to 0.408, while the Implicit Knowledge Recall subset improves from 0.333 to 0.667 Table~\ref{tab:prompt_rewriting_results}.

These gains suggest that some failures involve knowledge invocation or prompt formulation, rather than rendering alone. However, rewriting also changes instruction-following demands, preventing a clean separation of these factors. Remaining errors may reflect knowledge interpretation, instruction following, or visual rendering. We therefore treat this experiment as a preliminary diagnostic analysis.

Future work will develop controlled prompt pairs targeting the same factual content and evaluation criteria, with relevant facts either unstated or explicitly supplied. Other prompt characteristics and task demands should be matched as closely as possible, and correctness reported separately for each condition. The performance gap would characterize the benefit of factual guidance, but should not be interpreted as a pure measure of knowledge invocation.

\begin{table}[t]
\centering
\small
\caption{
Prompt rewriting results across knowledge operation types.
Rewriting implicit or underspecified knowledge constraints into direct visual instructions improves the overall case pass rate from 0.271 to 0.408.
}
\label{tab:prompt_rewriting_results}

\renewcommand{\arraystretch}{1.18}
\setlength{\tabcolsep}{6pt}

\begin{tabular}{
>{\raggedright\arraybackslash}p{0.30\linewidth}
>{\raggedright\arraybackslash}p{0.32\linewidth}
ccc
}
\toprule
\textbf{Slice} & \textbf{Category} & \textbf{Rewritten} & \textbf{Default} & \textbf{$\Delta$} \\
\midrule

Overall & ALL & \textbf{0.408} & 0.271 & \textbf{+0.137} \\
\midrule

\multirow{3}{*}{Knowledge operation}
& Explicit Knowledge Recall & 0.432 & 0.296 & +0.136 \\
& Implicit Knowledge Recall & \textbf{0.667} & 0.333 & \textbf{+0.334} \\
& Knowledge Reasoning & 0.263 & 0.132 & +0.131 \\
\bottomrule

\end{tabular}
\end{table}

\section{Conclusion}
\label{sec:conclusion}

We present {\benchmark}, a dual-view benchmark for diagnosing world-knowledge grounding in T2I models.
It combines an L1--L4 hierarchical taxonomy, a Skill-Tree Benchmark for L4-level atomic knowledge diagnosis, and a Generalized Benchmark for open-ended knowledge-grounded generation.
Together with a structured VQA-based MLLM-as-a-Judge protocol, {\benchmark} evaluates whether explicit or implicit knowledge constraints are correctly transformed into visual expressions.

Experiments show that the protocol aligns well with human judgments, better reflects world-knowledge correctness than general automatic metrics, and reveals failures that are not captured by general T2I capability alone.
These results suggest that world-knowledge grounding requires dedicated diagnostic evaluation beyond text--image alignment, image quality, or aesthetic preference.

\begin{ack}
We thank the anonymous reviewers for their constructive feedback.
This work was supported by KlingAI Research.
The authors declare no additional competing interests.
\end{ack}


\bibliographystyle{unsrtnat}
\bibliography{references}


\newpage

\appendix
\clearpage
\appendix

\renewcommand{\thesection}{\Alph{section}}
\renewcommand{\thesubsection}{\Alph{section}.\arabic{subsection}}

\startcontents[appendix]

\section*{Appendix}
\label{app:appendix}

\vspace{1em}
\noindent{\Large\bfseries Table of Contents}
\vspace{0.35em}
\hrule
\vspace{0.7em}

\begin{adjustwidth}{1.2cm}{0cm}
\printcontents[appendix]{}{1}{\setcounter{tocdepth}{2}}
\end{adjustwidth}

\vspace{0.6em}
\hrule
\clearpage
\section{Task Scope and Taxonomy Definition}
\label{app:task_scope_and_definitions}

\paragraph{Definition of World-Knowledge Grounding in T2I Generation.}
We define world-knowledge grounding in text-to-image generation as the ability of a model to identify, retrieve, reason over, and visually express the explicit or implicit knowledge constraints required by a prompt. These constraints may involve entities, attributes, relations, processes, rules, factual states, symbolic systems, physical commonsense, or culturally grounded concepts that are not fully specified by surface-level visual descriptions. A generated image is considered world-knowledge-grounded when it satisfies the relevant knowledge constraints and expresses them in a visually verifiable manner.

This notion differs from general text-image alignment, which mainly evaluates whether explicitly mentioned objects, attributes, and relations are rendered in the image. It also differs from generic image quality evaluation, which focuses on realism, aesthetics, perceptual quality, and artifacts rather than the correctness of the underlying knowledge.

\paragraph{Task Boundary.}
Our benchmark focuses on cases where correct image generation requires more than surface-level prompt rendering, including factual knowledge, structured relations, domain rules, symbolic conventions, physical commonsense, cultural knowledge, or reasoning over implicit constraints. We exclude prompts that primarily test generic image quality, aesthetics, stylistic preference, or low-level visual alignment without requiring world knowledge, such as ``a beautiful portrait,'' ``cinematic lighting,'' or ``a red cube on a blue table.'' Such prompts are included only when they impose additional knowledge constraints beyond the explicitly stated visual elements.

We further distinguish world-knowledge grounding from general instruction following: a model may follow the surface prompt while failing the implicit knowledge requirement, and an image may be visually appealing while still being knowledge-incorrect.

\subsection{Designing the {\benchmark} Taxonomy}
\label{app:world_knowledge_designing_taxonomy}

\paragraph{Hierarchy definition.}
Table~\ref{tab:definition_taxonomy_hierarchy} summarizes the functional role of each level in our L1-L4 hierarchical world-knowledge taxonomy.

\begin{table}[ht!]
    \centering
    \scriptsize
    \setlength{\tabcolsep}{2.0pt}
    \caption{Functional definitions of taxonomy levels L1–L4. The taxonomy progressively constrains and refines agent behavior from L1 to L4, with each level imposing stricter and more granular specifications than the previous.}
    \label{tab:definition_taxonomy_hierarchy}

    \begin{tabular}{llm{6cm}m{4cm}}
        \toprule
        
        \textbf{Level} & \textbf{Organizational Unit} & \textbf{Definition} & \textbf{Primary Role} \\

        \midrule
        
        \textbf{L1}
        & Domain
        & \textbf{The broadest knowledge category, defining the top-level thematic scope of the taxonomy.} Each L1 domain groups a set of related subjects that share a common epistemic nature (e.g., natural sciences, human civilization).
        & Supports overall coverage analysis and cross-domain capability comparison across the benchmark. \\
        
        \midrule
    
        \textbf{L2}
        & Subject
        & \textbf{A stable disciplinary subdivision within an L1 domain.} Each L2 subject corresponds to a well-established field of knowledge (e.g., architecture, biology), providing the structural backbone for organizing lower-level branches.
        & Ensuring a smooth transition from abstract domains to concrete branches. \\
        
        \midrule
        
        \textbf{L3}
        & Branch
        & \textbf{A specific knowledge direction within an L2 subject.} Concrete enough to define a visual task scenario (e.g., architectural styles, animal anatomy).
        & Serves as the basis for constructing the Generalized Benchmark. \\
        
        \midrule
        
        \textbf{L4}
        & Knowledge Point
        & \textbf{The minimal, atomic unit of knowledge.} Each L4 knowledge point has a clearly defined scope, a single verifiable claim, and a stable correctness criterion.
        & Directly used as the basic unit for Skill-Tree Benchmark construction, MLLM-as-a-judge evaluation, and attribution analysis. \\
        
        \bottomrule
    \end{tabular}
\end{table}

L4 is the minimal diagnosable knowledge capability unit in the taxonomy. It denotes an atomic knowledge point with clear boundaries, a limited invocation scope, and stable evaluation criteria, so that it can be individually assessed with a clear binary judgment. L4 points serve as the foundation for prompt construction, VQA checklist decomposition, and attribution analysis in the Skill-Tree Benchmark. We therefore exclude open-ended or ambiguous entries when defining L4 points.

\subsection{Taxonomy Construction Criteria}
\label{subsec:taxonomy_construction_criteria}

We construct the world-knowledge taxonomy from encyclopedia-style resources, library knowledge classification
systems, commonsense categories, LLM-assisted candidate expansion, and internal human review. L1/L2 levels are
organized from broad disciplinary and commonsense categories, while L3/L4 entries are expanded and refined
through LLM suggestions and manual review.

To adapt the taxonomy to text-to-image generation, we retain only knowledge points that are visually expressible
and judgeable through visual inspection or structured VQA-based criteria. We then iteratively check branch
coverage, remove ambiguous or weakly visualizable entries, and fill missing categories.

\section{Related Work}
\label{app:related_work}

The most relevant prior work falls into three categories: evaluation of image generation models, world-knowledge-oriented text-to-image generation tasks, and structured automatic evaluation.

\paragraph{General-Purpose T2I Evaluation.}

Early T2I evaluation work primarily centered on two axes: image quality and instruction following. On the preference-comparison side, PickScore \citep{kirstain2023pick} and ImageReward \citep{xu2023imagereward} learn pairwise human preferences, and HPSv3 \citep{ma2025hpsv3} further broadens this with a wide-spectrum comparison set and a VLM-based preference model; These methods scale well but produce holistic preference scores with limited interpretability. On the automatic-scoring side, CLIP and CLIPScore \citep{radford2021learning, hessel2021clipscore} measure text–image alignment via embedding similarity with high computational efficiency; GenEval \citep{ghosh2023geneval} pairs structured prompts with automatic metrics to assess color fidelity and spatial relations; T2I-CompBench and its extended version \citep{huang2023t2i, huang2025t2i} refine the evaluation granularity for attribute binding, spatial relationships, and object counting; GenAI-Bench \citep{li2024genai} introduces large-scale human annotations at the attribute–relation–logic level and demonstrates that VQAScore \citep{lin2024evaluating} substantially outperforms CLIPScore on fine-grained alignment. These methods effectively measure image quality, aesthetic performance, and overall instruction following. However, they share a common assumption that all relevant semantic information is explicitly encoded in the prompt. When an evaluation task requires the model to retrieve world knowledge beyond what is described in the prompt, or to perform deep semantic alignment or implicit reasoning, neither preference models nor embedding-similarity methods can reliably distinguish whether generated outputs satisfy factual constraints or commonsense regularities, limiting evaluation confidence.

\paragraph{World-Knowledge-Oriented T2I Benchmarks.}
 As interest in models' knowledge capabilities has grown, a number of benchmarks targeting specific knowledge types have been proposed. In the direction of factuality and knowledge-intensive evaluation, T2I-FactualBench \citep{huang2025t2ifactual} focuses on models' ability to memorize and discriminate knowledge concepts; MMMG \citep{luo2025mmmg} extends evaluation coverage to ten disciplines—including biology, chemistry, and mathematics—and six educational levels ranging from preschool to PhD, and represents each target image with a knowledge graph of entities and relations; it does not organize knowledge points within each discipline and offers insufficient granularity for assessing specific knowledge points. PhyBench \citep{meng2024phybench} systematically covers physical commonsense across mechanics, optics, thermodynamics, and material properties; T2I-ReasonBench \citep{sun2025t2ireasonbench} decomposes reasoning ability into four dimensions: idiom interpretation, textual image design, entity reasoning, and scientific reasoning. In the direction of implicit semantics and knowledge beyond the prompt, WISE \citep{niu2025wise} addresses world knowledge along three axes: cultural common sense, spatio-temporal reasoning, and natural science; ABP \citep{zhang2025align} explicitly introduces the concept of beyond-prompt knowledge, emphasizing attention to deep semantic consistency; LongT2IBench \citep{yang2026longt2ibench} focuses on semantic relationship generation under long-text conditions. In the direction of broad multi-domain coverage, WorldGenBench \citep{zhang2025worldgenbench} organizes human cultural and natural knowledge down to country, region, and natural science subfields; AEGIS \citep{lin2026aegis} further subdivides three macro-areas—STEM, humanities, and everyday life—into 21 sub-domains and introduces multi-dimensional labels to support attribution analysis. Overall, existing benchmarks have individually touched on key aspects such as factuality, common sense, reasoning, and implicit knowledge. However, most focus on a single knowledge type and remain limited in the systematic coverage of knowledge distributions, explicit stratification of difficulty levels, and atomic design of knowledge points. A unified and hierarchically organized model of the "world knowledge" capability space has yet to be established.

\paragraph{ Automated Evaluation Methods.}
 At the level of evaluation protocol, automated evaluation has evolved from embedding matching to structured decomposition to large-model judging. Early CLIP-based methods are computationally efficient but have limited discriminative power in complex compositional constraints and multi-step reasoning scenarios. TIFA \citep{hu2023tifa} and DSG \citep{cho2024dsg} substantially improve evaluation granularity and interpretability by decomposing prompts into atomic question-answer units, establishing the foundational paradigm for structured decomposition evaluation. As large model capabilities have advanced, the LLM/MLLM-as-a-Judge paradigm—exemplified by G-Eval, LLaVA-Bench, and MM-Vet \citep{liu2023g, liu2023visual, yu2024mm} has demonstrated strong potential for evaluating open-ended outputs. Further, WorldGenBench \citep{zhang2025worldgenbench} and AEGIS \citep{lin2026aegis} incorporate checklist mechanisms to improve evaluation stability; GenEval 2 \citep{kamath2025geneval} mitigates the tendency of purely atomic evaluation to diverge from holistic human judgment by combining atomic and open-ended evaluation via Soft-TIFA; GIR-Bench \citep{li2025gir} reveals the gap between cross-task knowledge transfer and faithful generation through understanding–generation consistency with explicit logical constraints; UniGenBench++ \citep{wang2025unigenbench++} provides a unified semantic assessment over hierarchical criteria, with point-wise scores that make individual test points interpretable. These advances demonstrate the practical feasibility of automated evaluation for T2I tasks. Nevertheless, world-knowledge-grounded generation imposes higher demands on automated evaluation than general instruction-following tasks: assessing correctness requires accurate verification against factual constraints, commonsense regularities, or implicit background knowledge, and the evaluation criteria themselves must be grounded in verifiable knowledge sources. Without structured evaluation design and reliability validation mechanisms, automated judgment results are difficult to trust and cannot support fine-grained attribution analysis.

 Table~\ref{tab:comparison_related_work} provides a systematic comparison of representative T2I world-knowledge benchmarks. As shown, existing works exhibit consistent limitations in both knowledge coverage systematicity and automatic evaluation reliability.

\begin{table}[ht!]
    \centering
    \small
    \setlength{\tabcolsep}{1pt}
    \resizebox{\columnwidth}{!}{
        \begin{threeparttable}
            \caption{Systematic comparison of existing T2I world-knowledge benchmarks across coverage distribution, benchmark hierarchy, evaluation protocol, and attribution labeling, revealing shared limitations in knowledge coverage systematicity and automatic evaluation reliability.}
            \label{tab:comparison_related_work}
            \begin{tabular}{l|ccccc|cc|ccc|cc}
                \toprule
        
                \multirow{2}{*}{\textbf{Benchmarks}}
                & \multicolumn{5}{c|}{\textbf{Taxonomy \tnote{$\star$}}}
                & \multicolumn{2}{c|}{\textbf{Dataset Structure}}
                & \multicolumn{3}{c|}{\textbf{Evaluation Protocol \tnote{$\dagger$}}}
                & \multicolumn{2}{c}{\textbf{Attribution \tnote{$\ddagger$}}} \\
                
                \cline{2-13}
                
                & N
                & F
                & H
                & A
                & R
                & Atomic 
                & Generalized
                & Form. 
                & MLLM.
                & Human Val.
                & Diff. \& Edu.
                & Text/Anno. \\
                
                \hline
                
                WISE~\citep{niu2025wise}                        & \cmark & \xmark & \cmark & \xmark & \xmark & \xmark & \cmark & Score-based     & \cmark & \cmark & \xmark & \xmark \\
                ABP~\citep{zhang2025align}                      & \cmark & \xmark & \cmark & \xmark & \xmark & \xmark & \cmark & Score-based     & \cmark & \cmark & \xmark & \xmark \\
                T2I-ReasonBench~\citep{sun2025t2ireasonbench}   & \cmark & \xmark & \cmark & \cmark & \xmark & \xmark & \cmark & Score-based     & \cmark & \cmark & \xmark & \xmark \\
                GIR-Bench (T2I)~\citep{li2025gir}               & \xmark & \cmark & \xmark & \xmark & \xmark & \xmark & \cmark & Score-based     & \xmark & \cmark & \xmark & \xmark \\
                UniGenBench++~\citep{wang2025unigenbench++}     & \xmark & \xmark & \cmark & \cmark & \xmark & \xmark & \cmark & Score-based     & \cmark & \xmark & \xmark & \xmark \\
                T2I-FactualBench~\citep{huang2025t2ifactual}    & \cmark & \xmark & \cmark & \xmark & \cmark & \cmark & \cmark & Score-based     & \cmark & \cmark & \xmark & \xmark \\
                PhyBench~\citep{meng2024phybench}               & \cmark & \xmark & \xmark & \xmark & \xmark & \cmark & \xmark & Score-based     & \cmark & \cmark & \xmark & \xmark \\
                MMMG~\citep{luo2025mmmg}                        & \cmark & \cmark & \cmark & \xmark & \xmark & \cmark & \xmark & Graph-based     & \cmark & \cmark & \xmark & \xmark \\
                WorldGenBench~\citep{zhang2025worldgenbench}    & \cmark & \xmark & \cmark & \xmark & \xmark & \xmark & \cmark & Checklist-based & \cmark & \xmark & \xmark & \xmark \\
                AEGIS (T2I)~\citep{lin2026aegis}                & \cmark & \cmark & \cmark & \cmark & \xmark & \xmark & \cmark & Checklist-based & \cmark & \cmark & \cmark & \xmark \\
                
                \hline
        
                \textbf{{\benchmark} (Ours)}                    & \cmark & \cmark & \cmark & \cmark & \cmark & \cmark & \cmark & Checklist-based & \cmark & \cmark & \cmark & \cmark \\
                
                \bottomrule
            \end{tabular}
    
            \begin{tablenotes}
                \scriptsize
                \item[$\star$] N: Natural Science; F: Formal Science; H: Human Civilization \& Society; H: Arts \& Media; R: Real-Time Information \& Events.
                \item[$\dagger$] Form. denotes evaluation formulation; MLLM. denotes the benchmark uses MLLM-as-a-Judge; Human Val. indicates whether the evaluation protocol is validated against human annotations or human judgments.
                \item[$\ddagger$] Diff. \& Edu. denotes drill down analysis by difficulty \& education level; Text/Anno. indicates whether the benchmark has conducted a drill-down attribution analysis on whether text rendering capability is required in text-to-image tasks.
            \end{tablenotes}
        \end{threeparttable}
    }
\end{table}

In summary, existing work has made substantive progress along two separate directions—breadth of knowledge coverage and automated evaluation protocols—yet a systematic integration of the two remains absent. Knowledge-oriented benchmarks largely lack unified modeling of the knowledge capability space and hierarchically structured benchmark design, while existing automated evaluation protocols have yet to simultaneously achieve reliability, interpretability, and fine-grained attribution in world-knowledge tasks. Our work sits at the intersection of these two lines of research, aiming to construct a unified world-knowledge benchmark framework paired with an attributable, structured automated evaluation protocol—one that jointly supports atomic capability diagnosis, hierarchical knowledge organization, and generalization analysis in open, complex scenarios.
 
\section{Benchmark Construction Details}
\label{app:benchmark_construction_details}

We construct two independent yet logically aligned data pipelines to support \textbf{the Skill-Tree Benchmark} and \textbf{the Generalized Benchmark}, respectively. Both benchmarks share the same L1–L4 hierarchical taxonomy, but their evaluation objectives, sample organization, knowledge granularity, and intended use cases differ fundamentally, making a unified construction pipeline inappropriate.

The Skill-Tree Benchmark is designed for \textbf{systematic diagnosis of model capabilities}. Its central questions are: Does the model possess a given type of atomic world knowledge? At which knowledge points does it fail consistently? How do different models compare on the same knowledge point? Accordingly, this benchmark requires broad coverage of L4 atomic knowledge points, with each point anchored by samples that are decidable, attributable, and amenable to large-scale statistical analysis. Its construction therefore follows a bottom-up, L4-focused batch generation approach: LLM-driven prompt generation centered on atomic knowledge points, followed by automated filtering and human review.

The Generalized Benchmark is designed to \textbf{assess model performance on realistic, complex tasks}. Its central question is: When a user issues a complex, natural, open-ended generation request, can the model draw on multiple knowledge points simultaneously and produce correct, rich, and coherent knowledge expressions in the generated image? Real user prompts rarely isolate a single knowledge point; they typically involve entities, attributes, relations, cultural context, physical constraints, and contextual reasoning in combination. As such, the Generalized Benchmark is not suited to direct bottom-up assembly from L4 points. Instead, its samples are crafted manually around L3 knowledge branches, with L4 annotations added after sample completion, ensuring that even complex tasks remain traceable within the unified knowledge attribution framework.

We therefore adopt a strategy of \textbf{separate construction with unified attribution}: both data streams originate from the same taxonomy but follow distinct construction pipelines according to their evaluation objectives, and both ultimately converge on a shared evaluation and attribution framework via L4 labels and VQA checklists. This design ensures that the Skill-Tree Benchmark achieves high coverage, scale, and diagnostic precision, while the Generalized Benchmark preserves the complexity, compositional diversity, and ecological validity required for realistic evaluation scenarios.

\subsection{Skill-Tree Benchmark Synthesis}
\label{app:skill_tree_benchmark_synthesis}

The Skill-Tree Benchmark is constructed via a pipeline of \textbf{LLM-based automatic generation, LLM-based review and filtering, and human screening and rewriting}. The overall workflow proceeds as follows:

\begin{enumerate}
    \item An LLM generates candidate prompts in batch according to predefined knowledge points.
    \item A separate model performs automated review and filtering on the candidates:
    \begin{itemize}
        \item Samples that pass review enter the result pool.
        \item Samples that fail, along with their identified failure reasons, are fed back into the next generation iteration.
    \end{itemize}
    \item Generation and filtering form a closed-loop feedback process, which is iteratively repeated until the result pool reaches the target size.
\end{enumerate}

\paragraph{Automated Filtering Criteria.}
The automated filtering stage is designed to remove candidate prompts that are difficult to evaluate reliably or insufficiently aligned with the target L4 knowledge point. A prompt is accepted only if it clearly targets the current L4 knowledge point, can be visually executed, and does not violate predefined format or content constraints. Conversely, a prompt is filtered or marked for revision if it is off-topic, visually unverifiable, overly complex, conflates multiple knowledge points, or violates the required prompt format. We describe the main filtering criteria below and provide representative examples of candidates rejected by the reviewer model.

\begin{enumerate}
    \item \textbf{Knowledge alignment: must be anchored to the current L4.}
    The core subject of a prompt must correspond to the L4 leaf node in the current taxonomy. It should not drift toward other knowledge points, other disciplinary concepts, or content on the exclusion list. If the prompt introduces knowledge unrelated to the current L4, or touches content prohibited under ``prompt types to exclude,'' it should be filtered or marked for revision.
    \begin{tcolorbox}[
      enhanced,
      colback=gray!3,
      colframe=black!60,
      title={Example: filtered by the reviewer model},
      fonttitle=\bfseries,
      coltitle=black,
      colbacktitle=gray!15,
      boxrule=0.5pt,
      arc=1mm,
      left=1.5mm,
      right=1.5mm,
      top=1.2mm,
      bottom=1.2mm
    ]
    \small
    \textbf{Target taxonomy:}
    Natural Science $\rightarrow$ Physics $\rightarrow$ Mechanics $\rightarrow$ Concurrent Force Equilibrium.

    \vspace{0.45em}
    \textbf{Candidate prompt:}
    Draw a force diagram of a wooden block resting on an inclined plane held by a string, and label the equilibrium of concurrent forces.

    \vspace{0.45em}
    \textbf{Filtering reason:}
    Although the prompt is related to concurrent force equilibrium, it relies on an inclined-plane force diagram and may introduce multiple confounding mechanics concepts, including gravity, friction, support force, tension, and force decomposition. This makes the evaluation target less focused and increases visual complexity and attribution ambiguity for a single L4 knowledge point.

    \vspace{0.45em}
    \textbf{Revision suggestion:}
    Rewrite the prompt into a more focused static-equilibrium scene, such as an educational illustration of a decorative lamp suspended by three strings connected at the same point and remaining stationary, with a label indicating concurrent force equilibrium.
    \end{tcolorbox}
    \item \textbf{Visual executability: must be drawable.}
    A prompt must be translatable into a concrete image, scene, experiment, diagram, or visualization. Prompts that are purely theoretical, purely abstract, purely derivational, philosophical, or lack a visual subject should be filtered or marked for revision.
    \begin{tcolorbox}[
      enhanced,
      colback=gray!3,
      colframe=black!60,
      title={Example: filtered by the reviewer model},
      fonttitle=\bfseries,
      coltitle=black,
      colbacktitle=gray!15,
      boxrule=0.5pt,
      arc=1mm,
      left=1.5mm,
      right=1.5mm,
      top=1.2mm,
      bottom=1.2mm
    ]
    \small
    \textbf{Target taxonomy:}
    Human Civilization and Society $\rightarrow$ Humanities and Architecture $\rightarrow$ Architectural Styles $\rightarrow$ Renaissance Architecture.

    \vspace{0.45em}
    \textbf{Candidate prompt:}
    Show the front elevation of the Villa Rotonda, using geometric auxiliary lines to outline its symmetry.

    \vspace{0.45em}
    \textbf{Filtering reason:}
    The prompt asks the model to visualize an abstract design principle rather than a concrete architectural component. Although symmetry is relevant to Renaissance architecture, ``outlining symmetry'' is an open-ended analytical requirement and lacks a clearly decidable visual target.

    \vspace{0.45em}
    \textbf{Revision suggestion:}
    Shift the focus from an abstract property to a concrete physical component, such as: draw the front elevation of the Villa Rotonda and label the triangular pediment at the top with its name.
    \end{tcolorbox}
    \item \textbf{Atomicity: only one primary knowledge point should be tested.}
    A prompt should involve a single, well-defined knowledge call. It should not test multiple coordinate concepts within the same image, require a complete depiction of a complex system, or rely on extensive specialist annotations. Prompts that simultaneously test multiple knowledge points or make failure attribution difficult should be filtered or marked for revision.
    \begin{tcolorbox}[
      enhanced,
      colback=gray!3,
      colframe=black!60,
      title={Example: filtered by the reviewer model},
      fonttitle=\bfseries,
      coltitle=black,
      colbacktitle=gray!15,
      boxrule=0.5pt,
      arc=1mm,
      left=1.5mm,
      right=1.5mm,
      top=1.2mm,
      bottom=1.2mm
    ]
    \small
    \textbf{Target taxonomy:}
    Natural Science $\rightarrow$ Biology $\rightarrow$ Organs and Basic Functions $\rightarrow$ Animal Organs and Their Basic Functions.

    \vspace{0.45em}
    \textbf{Candidate prompt:}
    Draw an anatomical diagram of a fish and label the name of the core organ responsible for gas exchange.

    \vspace{0.45em}
    \textbf{Filtering reason:}
    The prompt requires a fish anatomical diagram, which may involve multiple organs and internal structures beyond the target knowledge point. This makes the evaluation target less atomic and increases the difficulty of attributing errors to a single organ-function relation.

    \vspace{0.45em}
    \textbf{Revision suggestion:}
    Simplify the visual carrier to a single organ or local region, such as: draw a structural diagram of fish gills and label their function in gas exchange, or draw a complete fish and circle the location of the gills.
    \end{tcolorbox}

    \item \textbf{Clarity and universality: specific expression, universally recognizable carrier.}
    A prompt must clearly specify what object, scene, or diagram should be generated. It should avoid vague expressions such as ``some,'' ``various,'' ``roughly,'' or ``depict a certain situation.'' The visual carrier should rely on broadly recognizable objects or diagrams, rather than niche cultural items or highly domain-specific carriers that may introduce knowledge conflation.
    \begin{tcolorbox}[
      enhanced,
      colback=gray!3,
      colframe=black!60,
      title={Example: filtered by the reviewer model},
      fonttitle=\bfseries,
      coltitle=black,
      colbacktitle=gray!15,
      boxrule=0.5pt,
      arc=1mm,
      left=1.5mm,
      right=1.5mm,
      top=1.2mm,
      bottom=1.2mm
    ]
    \small
    \textbf{Target taxonomy:}
    Natural Science $\rightarrow$ Physics $\rightarrow$ Thermodynamics $\rightarrow$ Calorific Value and Heat Engine.

    \vspace{0.45em}
    \textbf{Candidate prompt:}
    Draw a Sankey diagram of heat engine energy flow, accurately labeling useful work and dissipated heat.

    \vspace{0.45em}
    \textbf{Filtering reason:}
    The prompt requires an accurate Sankey diagram but does not specify the underlying energy distribution. Without concrete values for useful work, exhaust losses, and dissipated heat, the visual solution is underdetermined and difficult to evaluate consistently.

    \vspace{0.45em}
    \textbf{Revision suggestion:}
    Provide specific energy distribution values, such as total heat input 100\%, useful work 30\%, exhaust heat loss 40\%, and thermal dissipation 30\%, and specify where the labels should appear.
    \end{tcolorbox}

    \item \textbf{Structural regularity: short, single-sentence, no answer spoilers.}
    Prompts should be concise, structurally simple, and within the preset length limit. Compound prompt structures using conjunctions such as ``and,'' ``also,'' or ``moreover'' should be avoided when they introduce additional requirements. A prompt should also avoid directly revealing the target answer or final knowledge outcome, unless the task is explicitly designed as knowledge visualization rather than knowledge recall.

    \begin{tcolorbox}[
      enhanced,
      colback=gray!3,
      colframe=black!60,
      title={Example: filtered by the reviewer model},
      fonttitle=\bfseries,
      coltitle=black,
      colbacktitle=gray!15,
      boxrule=0.5pt,
      arc=1mm,
      left=1.5mm,
      right=1.5mm,
      top=1.2mm,
      bottom=1.2mm
    ]
    \small
    \textbf{Target taxonomy:}
    Formal Science $\rightarrow$ Mathematics $\rightarrow$ Numbers and Calculation $\rightarrow$ Derivatives and Differentials.

    \vspace{0.45em}
    \textbf{Candidate prompt:}
    Draw a freely falling ball, and write the formula $a = dv/dt$ beside it.

    \vspace{0.45em}
    \textbf{Filtering reason:}
    The prompt combines multiple requirements in one sentence and directly provides the key formula in the prompt. This weakens the intended knowledge-recall or knowledge-grounding objective and makes the evaluation less diagnostic.

    \vspace{0.45em}
    \textbf{Revision suggestion:}
    Rewrite the prompt into a more focused task that does not directly reveal the target answer, or explicitly treat it as a knowledge-visualization task if the formula is intended to be provided.
    \end{tcolorbox}

\end{enumerate}

\paragraph{Human Screening and Rewriting Criteria.}
Human screening and rewriting is conducted as a subsequent refinement stage after automated filtering. While the automated reviewer model removes candidates with obvious violations, such as off-topic content, non-visualizability, excessive complexity, or format errors, human screening further focuses on borderline cases that require semantic judgment, evaluation stability, and benchmark-level consistency. In other words, automated filtering serves as a first-pass quality gate, whereas human screening and rewriting acts as a second-pass refinement step to improve evaluability, attributability, and stability of the Skill-Tree Benchmark.

The human screening and rewriting process follows the principles below.
\begin{enumerate}
    \item \textbf{Single-variable principle.}
    Each L4 knowledge point should focus on exactly one key variable, avoiding the inclusion of multiple core examination targets within the same sample. This reduces judgment interference and improves capability isolation.

    \item \textbf{Visualization constraint.}
    Each selected knowledge point must be clearly mappable to image content {\textemdash} i.e., it must have a well-defined visual expression basis. Since the core objective of the Skill-Tree Benchmark is to assess the model's knowledge-to-image mapping ability, knowledge content that is difficult to visualize or that primarily requires textual reasoning for evaluation is not suitable for inclusion.

    \item \textbf{Decidability principle.}
    Each sample must have a clear and stable ground truth, capable of reliably distinguishing correct knowledge expression from incorrect knowledge expression, while avoiding cases where evaluation boundaries are ambiguous or subjective interpretation dominates.

    \item \textbf{Disambiguation principle.}
    Sample construction should minimize ambiguous expressions and multiple interpretable readings, ensuring that the knowledge boundary targeted by the prompt is clear and semantically unambiguous, thereby reducing evaluation noise introduced by linguistic ambiguity.

    \item \textbf{High-coverage principle.}
    Knowledge point selection should prioritize the following:
    \begin{itemize}
        \item \textbf{High-frequency knowledge:} knowledge content that appears more commonly in real-world use scenarios and is more salient to users;
        \item \textbf{General commonsense:} foundational knowledge with broad applicability;
        \item \textbf{Cross-culturally stable knowledge:} knowledge whose content and evaluation criteria remain relatively consistent across different cultural contexts.
    \end{itemize}
\end{enumerate}

\paragraph{Examples of Human Rejection after Automated Acceptance.}
The following examples illustrate candidates that passed automated filtering but were removed during human screening. These cases are not obviously invalid at the surface level: they are generally aligned with the target L4 knowledge point and have a drawable visual carrier. However, human screening further removes them because they violate higher-level requirements such as variable isolation, decidability, or semantic disambiguation.

\begin{enumerate}
    \item \textbf{Violation of the single-variable principle.}
    A sample should focus on a single core variable associated with the target L4 knowledge point. If a prompt mixes the target knowledge with unrelated generation abilities such as slide design, layout design, formula rendering, or excessive visual composition, it may reduce attribution clarity and should be removed or rewritten.

    \begin{tcolorbox}[
      enhanced,
      colback=gray!3,
      colframe=black!60,
      title={Example: accepted by automated filtering but rejected by human screening},
      fonttitle=\bfseries,
      coltitle=black,
      colbacktitle=gray!15,
      boxrule=0.5pt,
      arc=1mm,
      left=1.5mm,
      right=1.5mm,
      top=1.2mm,
      bottom=1.2mm
    ]
    \small
    \textbf{Target taxonomy:}
    Natural Science $\rightarrow$ Physics $\rightarrow$ Mechanics $\rightarrow$ Pressure.

    \vspace{0.45em}
    \textbf{Candidate prompt:}
    Design a physics course slide with the formula $p=\rho gh$ centered on the page and draw a schematic diagram of water depth.

    \vspace{0.45em}
    \textbf{Reason for automated acceptance:}
    The prompt is clearly related to the target knowledge point of pressure. The formula $p=\rho gh$ and the schematic diagram of water depth both directly correspond to the relationship between liquid pressure and depth. The visual target is explicit, and the prompt does not obviously drift toward unrelated mechanics concepts.

    \vspace{0.45em}
    \textbf{Reason for human rejection:}
    Although the prompt is semantically aligned with pressure, the actual generation target mixes multiple abilities, including formula rendering, course-slide layout, diagram composition, and physical relation visualization. For an atomic Skill-Tree sample, a failure may come from text rendering or slide formatting rather than from misunderstanding the pressure knowledge point itself. Therefore, the sample is removed under the single-variable principle.
    \end{tcolorbox}

    \item \textbf{Violation of the decidability principle.}
    A sample should have a clear and stable expected outcome. If multiple visual outputs can be considered reasonable under the same prompt, or if the correctness boundary cannot be reliably determined, the sample should be removed or rewritten.

    \begin{tcolorbox}[
      enhanced,
      colback=gray!3,
      colframe=black!60,
      title={Example: accepted by automated filtering but rejected by human screening},
      fonttitle=\bfseries,
      coltitle=black,
      colbacktitle=gray!15,
      boxrule=0.5pt,
      arc=1mm,
      left=1.5mm,
      right=1.5mm,
      top=1.2mm,
      bottom=1.2mm
    ]
    \small
    \textbf{Target taxonomy:}
    Natural Science $\rightarrow$ Physics $\rightarrow$ Mechanics $\rightarrow$ Buoyancy.

    \vspace{0.45em}
    \textbf{Candidate prompt:}
    Show the most reasonable posture of a person moving freely in the Dead Sea.

    \vspace{0.45em}
    \textbf{Reason for automated acceptance:}
    The prompt is related to the knowledge that the high salinity of the Dead Sea makes human bodies easier to float. It has a clear visual subject, namely a person in the Dead Sea, and can be converted into an image-generation task. It also does not directly reveal the expected result of floating, which makes it superficially suitable for testing knowledge-driven generation.

    \vspace{0.45em}
    \textbf{Reason for human rejection:}
    The expression ``moving freely'' can correspond to many visually plausible states, such as floating, standing, swimming, walking, sitting, or playing in water. The phrase ``most reasonable posture'' also lacks a stable and unique evaluation criterion. As a result, it is difficult to reliably determine whether the generated image correctly expresses the target buoyancy-related knowledge. The sample is therefore removed because it does not satisfy the decidability principle.
    \end{tcolorbox}

    \item \textbf{Violation of the disambiguation principle.}
    A sample should avoid ambiguous wording and underspecified conditions. If the prompt admits multiple interpretations that correspond to different expected answers, the sample should be removed or rewritten to make the intended knowledge target explicit.

    \begin{tcolorbox}[
      enhanced,
      colback=gray!3,
      colframe=black!60,
      title={Example: accepted by automated filtering but rejected by human screening},
      fonttitle=\bfseries,
      coltitle=black,
      colbacktitle=gray!15,
      boxrule=0.5pt,
      arc=1mm,
      left=1.5mm,
      right=1.5mm,
      top=1.2mm,
      bottom=1.2mm
    ]
    \small
    \textbf{Target taxonomy:}
    Natural Science $\rightarrow$ Climate and Astronomy $\rightarrow$ Day-Night and Time Change $\rightarrow$ Time Zone and Local Time Calculation.

    \vspace{0.45em}
    \textbf{Candidate prompt:}
    Show clock faces for Tokyo and Los Angeles in the morning.

    \vspace{0.45em}
    \textbf{Reason for automated acceptance:}
    The prompt uses clock faces as a clear visual carrier and appears to test the representation of local time or time-zone difference between Tokyo and Los Angeles. The knowledge target is relatively focused, and the visual output can in principle be judged through the displayed clock times and their city associations.

    \vspace{0.45em}
    \textbf{Reason for human rejection:}
    The prompt does not specify whether ``morning'' refers to morning in Tokyo, morning in Los Angeles, or the morning time in both cities separately. Without a reference time such as ``when it is 8:00 a.m. in Tokyo,'' the task may be interpreted as drawing two independent morning clocks rather than performing a time-zone conversion. This ambiguity makes the expected answer underdetermined, so the sample is removed under the disambiguation principle.
    \end{tcolorbox}

\end{enumerate}

\subsection{Generalized Benchmark Curation}
\label{subsec:generalized_benchmark_curation}

Unlike the Skill-Tree Benchmark, the Generalized Benchmark targets more open-ended and realistic generation scenarios. Therefore, its cases are not constructed around a single L4 atomic knowledge point as the only center. Instead, they are manually designed and curated primarily around L3 knowledge branches. Each generalized case is typically associated with a target L3 branch. After the case is finalized, we further annotate the set of potentially involved L4 atomic knowledge points, which supports subsequent cross-view analysis and error attribution.

During manual construction, we follow the principles below.

\begin{enumerate}
    \item \textbf{Alignment with the Target L3 Branch.}
    Each generalized case should first be clearly aligned with a target L3 branch. We use L3 as the main organizational unit for open-ended scenario construction because it is more suitable than L4 for supporting complex prompts, while being more specific than L1/L2 and thus maintaining clear task boundaries.

    For example, under \textit{Natural Science $\to$ Physics $\to$ Mechanics}, we can construct open-ended scenarios involving mechanical phenomena such as force, motion, equilibrium, pressure, and buoyancy. Under \textit{Formal Science $\to$ Mathematics $\to$ Geometric Figures}, we can construct open-ended scenarios involving geometric objects, angular relations, plane figures, solid geometry, volume, or geometric formulas. Such cases are not required to correspond to only one L4 atomic knowledge point. Instead, they may naturally involve multiple related L4 points within the same L3 branch.

    \item \textbf{Branch Representativeness.}
    Cases should cover knowledge scenarios that are representative, common, and diagnostically valuable within the target L3 branch, rather than overly marginal or artificially designed cases.

    For example, within \textit{Physics $\to$ Mechanics}, representative cases may jointly examine L4 atomic knowledge points such as gravity, friction, pressure, Newton's laws, torque equilibrium, and force composition and decomposition. These L4 points are common and visually expressible within Mechanics, making them suitable as potential knowledge anchors in generalized cases.

    \item \textbf{Task Diversity.}
    The Generalized Benchmark should not be limited to a single prompt format, such as formula diagrams or instructional slides. During manual construction, we cover diverse real-world generation needs, including instructional diagrams, science posters, realistic scene illustrations, structural diagrams, annotated explanatory figures, comparison diagrams, and integrated explanatory visuals.

    For example, cases under \textit{Physics $\to$ Mechanics} may include both instructional diagrams for mechanics and real-world scenes illustrating equilibrium, buoyancy, or pressure. Cases under \textit{Mathematics $\to$ Geometric Figures} may include geometric structure diagrams as well as learning-oriented illustrations with formulas, angles, side lengths, or volume annotations. This diversity prevents the benchmark from evaluating only one type of visual capability and enables a broader assessment of world-knowledge grounding in open-ended generation tasks.

    \item \textbf{Supplementary L4 Annotation.}
    Although the Generalized Benchmark is primarily constructed around L3 branches, we supplement each finalized case with annotations of the potentially involved L4 atomic knowledge points. This supplementary L4 annotation is not intended to forcibly decompose a generalized case into single-point tests. Instead, it establishes a mapping between the Generalized Benchmark and the Skill-Tree Benchmark, allowing us to analyze whether failures in open-ended scenarios are associated with weaknesses in specific foundational L4 atomic knowledge points.

    \item \textbf{Cases Annotated Only at the L2 Level.}
    For some domains, the boundaries of L3/L4 may be unstable, or the knowledge may be difficult to decompose into objective and visually judgeable atomic points. In such cases, the Generalized Benchmark allows a small number of samples to be annotated only at the L2 level.

    Typical examples include \textit{Arts and Media $\to$ Visual Arts} and \textit{Arts and Media $\to$ Design and Communication}. Knowledge in these domains often involves style, aesthetics, cultural context, and open-ended expression. Compared with domains such as physics and mathematics, they may not be stably decomposable into clearly bounded L4 atomic knowledge points that can be judged through a small set of VQAs. Forcibly decomposing them may introduce substantial subjectivity.
\end{enumerate}

\paragraph{Examples of Generalized Benchmark Curation.}
The following examples illustrate how generalized cases are manually curated around target L3 branches and then annotated with potentially involved L4 knowledge points. Unlike the Skill-Tree Benchmark, which focuses on a single L4 atomic knowledge point, the Generalized Benchmark allows a prompt to involve multiple related L4 points under the same L3 branch, as long as the task remains representative, visually assessable, and close to realistic open-ended generation scenarios.

\begin{tcolorbox}[
  enhanced,
  colback=gray!3,
  colframe=black!60,
  title={Example 1: Mechanics-oriented generalized case},
  fonttitle=\bfseries,
  coltitle=black,
  colbacktitle=gray!15,
  boxrule=0.5pt,
  arc=1mm,
  left=1.5mm,
  right=1.5mm,
  top=1.2mm,
  bottom=1.2mm
]
\small
\textbf{Target L3 branch:}
Natural Science $\rightarrow$ Physics $\rightarrow$ Mechanics.

\vspace{0.45em}
\textbf{Prompt:}
Draw an adult and a child on both sides of a wooden door pushing it in opposite directions to keep it stationary, with the adult close to the door hinge and the child far away, using arrows to mark the direction of their pushing forces.

\vspace{0.45em}
\textbf{Potential L4 annotations:}
Torque Equilibrium; Force Composition and Decomposition; Concurrent Force Equilibrium.

\vspace{0.45em}
\textbf{Curation rationale:}
This case is aligned with the Mechanics branch and represents a realistic physical scenario rather than an isolated textbook formula. The setting of an adult pushing near the hinge and a child pushing farther away implicitly tests whether the model can express the effect of force arm length on rotational equilibrium. The requirement to mark pushing directions with arrows further introduces structured visual annotation. Therefore, this sample is suitable for the Generalized Benchmark because it combines scenario realism, L3 alignment, multiple related L4 knowledge points, and moderate visual reasoning complexity.
\end{tcolorbox}

\subsection{Cross-view Iterative Refinement.}
The two views are complementary through a cross-view iterative refinement loop. The Skill-Tree Benchmark provides an attributable and localized profile of atomic knowledge capabilities, which helps explain the sources of failure in complex open-ended cases from the Generalized Benchmark. Conversely, the Generalized Benchmark exposes new knowledge combinations, implicit constraints, and expression patterns that emerge in realistic open-ended scenarios. These observations help identify coverage gaps in the Skill-Tree taxonomy and further drive the supplementation and refinement of L3/L4 categories.

\section{Annotation Schema and Benchmark Statistics}
\label{app:annotation_schema_and_benchmark_statistics}

\subsection{Case-level Metadata Schema}
\label{subsec:case_level_metadata_schema}

To support cross-view analysis, capability slicing, and error attribution, we define a unified case-level metadata schema for both the Skill-Tree Benchmark and the Generalized Benchmark. The schema describes each case along three task attributes: knowledge operation type, difficulty, and education level.

Skill-Tree only uses three categories under knowledge operation type and does not use difficulty. Generalized uses four categories under knowledge operation type and additionally annotates difficulty. Education level is used in both benchmark views. This shared schema keeps the two views aligned under a common annotation language, while allowing each view to preserve its own evaluation focus.

\paragraph{Knowledge Operation Type.}
Knowledge operation type describes the main way in which a model is expected to use world knowledge when completing a case. If multiple knowledge operation types are involved, we annotate the primary and highest-level knowledge use.

\begin{itemize}[leftmargin=*]
    \item \textbf{Explicit Knowledge Rendering.}
    The prompt directly provides the expected answer or knowledge content, and the model is mainly required to visualize it. This includes directly expressing scientific principles, entity names, quantitative relations, or logical processes through non-natural indicative elements such as text, numbers, symbols, arrows, scales, or callouts. It also includes cases where the prompt fully specifies the inherent properties of a knowledge-intensive entity and only requires visual rendering.

    \item \textbf{Explicit Knowledge Recall.}
    The prompt directly specifies the name of a knowledge-intensive entity, concept, or phenomenon. The model can generate the expected visual form by recalling the corresponding keyword and its associated visual patterns from training data, without necessarily understanding the underlying mechanism, related knowledge, or real-world simulation process.

    \item \textbf{Implicit Knowledge Recall.}
    The prompt does not directly mention the target knowledge point. Instead, it describes scene context, initial states, actions, typical appearances, or interaction conditions, requiring the model to indirectly recall the relevant physical state, interaction outcome, scientific phenomenon, or knowledge concept from world knowledge and visualize it.

    \item \textbf{Knowledge Reasoning.}
    Beyond explicit or implicit recall, the model needs to reason, decompose, judge, or calculate in order to apply knowledge to a specific problem. When a case involves both recall and reasoning, we annotate it as Knowledge Reasoning because it represents a higher-level form of knowledge use.
\end{itemize}

\paragraph{Difficulty.}
Difficulty describes the overall answering difficulty of a generalized case and is used only in the Generalized Benchmark. It is annotated by jointly considering semantic complexity and presentation complexity. Semantic complexity captures noise, long-range dependencies, implicit conditions, abstract descriptions, or metaphors in the prompt. Presentation complexity captures how difficult it is to express the knowledge in visual form, such as transforming a simple scene into a complex cutaway diagram, arranging multiple entities, binding annotations to visual elements, or organizing formulas, arrows, and explanatory structures.

We use three levels: high, medium, and low. High-difficulty cases usually involve complex semantics and complex visual presentation. Medium-difficulty cases have relatively clear knowledge targets but still require non-trivial scene understanding, knowledge composition, or visual organization. Low-difficulty cases have stable knowledge targets and simple visual forms.

Difficulty is not the same as education level. Difficulty describes the complexity of a specific generation task, while education level describes the typical learning stage or knowledge source of the involved knowledge.

\paragraph{Education Level.}
Education level indicates the typical learning stage or knowledge source associated with a case. It is used to analyze model performance across different levels of knowledge, rather than to directly represent task difficulty.

We use five labels: primary school, middle school, high school, university, and professional domain. The first four correspond to knowledge typically learned at the respective educational stages. \textit{Professional domain} refers to knowledge that is not part of general education, but is usually acquired through professional training, occupational practice, industry experience, or social contexts. Examples include national standards, industry procedures, professional roles, specialized skills, social conventions, traditional customs, public-space signage, and architectural functional zones.

\subsection{Benchmark Statistics}
\label{subsec:benchmark_statistics}

Based on the case-level metadata schema defined above, we summarize the benchmark statistics from two perspectives: taxonomy coverage and task-attribute distribution. These statistics are intended to verify the coverage of the two benchmark views, their cross-view alignment, and the diagnostic utility of the proposed annotation schema.

\paragraph{L1 Domain Distribution.}
Table~\ref{tab:l1_distribution} reports the L1 domain distribution of both views. The two views share a similar major-domain trend: Natural Science, Human Civilization and Society, and Formal Science are the primary domains in both benchmarks. This indicates that the two views are aligned under the same world-knowledge taxonomy. At the same time, the Generalized Benchmark further includes Arts and Media and Real-time Dynamic Events, which are more open-ended and less suitable for stable L4-level atomic decomposition. 

\begin{table}[t]
\centering
\small
\caption{L1 domain distribution of the Skill-Tree and Generalized benchmarks.}
\label{tab:l1_distribution}
\begin{tabular}{lrrrr}
\toprule
\multirow{2}{*}{L1 Domain} & \multicolumn{2}{c}{Skill-Tree} & \multicolumn{2}{c}{Generalized} \\
\cmidrule(lr){2-3} \cmidrule(lr){4-5}
 & Count & Percent & Count & Percent \\
\midrule
Natural Science & 841 & 51.8\% & 224 & 40.7\% \\
Human Civilization and Society & 489 & 30.1\% & 180 & 32.7\% \\
Formal Science & 292 & 18.0\% & 71 & 12.9\% \\
Arts and Media & - & - & 50 & 9.1\% \\
Real-time Dynamic Events & - & - & 25 & 4.5\% \\
\bottomrule
\end{tabular}
\end{table}

\paragraph{Skill-Tree Task Attributes.}
Table~\ref{tab:skill_tree_attributes} reports the task-attribute distribution of the Skill-Tree Benchmark. Since Skill-Tree is centered on L4 atomic knowledge points, it does not annotate difficulty. It uses three knowledge operation types: Explicit Knowledge Recall, Implicit Knowledge Recall, and Knowledge Reasoning. The distribution is dominated by Explicit Knowledge Recall, which is consistent with the goal of testing whether models can reliably recall and visualize canonical L4 knowledge concepts. The benchmark also includes Knowledge Reasoning and a small number of Implicit Knowledge Recall cases, enabling analysis of whether models can infer or apply atomic knowledge beyond direct concept-name recall.

\begin{table}[t]
\centering
\small
\caption{Task-attribute distribution of the Skill-Tree Benchmark.}
\label{tab:skill_tree_attributes}
\begin{tabular}{llrr}
\toprule
Attribute & Sub-label & Count & Percent \\
\midrule
\multirow{5}{*}{Education level}
& Primary school & 166 & 10.2\% \\
& Middle school & 715 & 44.1\% \\
& High school & 435 & 26.8\% \\
& University & 99 & 6.1\% \\
& Professional domain & 207 & 12.8\% \\
\midrule
\multirow{3}{*}{Knowledge operation type}
& Explicit Knowledge Recall & 1,322 & 81.5\% \\
& Implicit Knowledge Recall & 33 & 2.0\% \\
& Knowledge Reasoning & 267 & 16.5\% \\
\bottomrule
\end{tabular}
\end{table}

\paragraph{Generalized Task Attributes.}
Table~\ref{tab:generalized_attributes} reports the task-attribute distribution of the Generalized Benchmark. Compared with Skill-Tree, Generalized uses the full four-way knowledge operation schema and additionally annotates difficulty and knowledge quantity. The benchmark contains 126 Knowledge Reasoning cases and 235 compositional-knowledge cases, showing that it evaluates not only direct knowledge recall but also higher-level knowledge use and multi-knowledge composition. In addition, professional-domain cases account for a large proportion, reflecting that open-ended T2I generation often involves knowledge from professional, social, product, and public-space contexts rather than only school-based knowledge.

\begin{table}[t]
\centering
\small
\caption{Task-attribute distribution of the Generalized Benchmark.}
\label{tab:generalized_attributes}
\begin{tabular}{llrr}
\toprule
Attribute & Sub-label & Count & Percent \\
\midrule
\multirow{3}{*}{Difficulty}
& Low & 293 & 53.3\% \\
& Medium & 176 & 32.0\% \\
& High & 81 & 14.7\% \\
\midrule
\multirow{5}{*}{Education level}
& Primary school & 27 & 4.9\% \\
& Middle school & 170 & 30.9\% \\
& High school & 94 & 17.1\% \\
& University & 10 & 1.8\% \\
& Professional domain & 249 & 45.3\% \\
\midrule
\multirow{4}{*}{Knowledge operation type}
& Explicit Knowledge Rendering & 30 & 5.5\% \\
& Explicit Knowledge Recall & 372 & 67.6\% \\
& Implicit Knowledge Recall & 22 & 4.0\% \\
& Knowledge Reasoning & 126 & 22.9\% \\
\midrule
\multirow{2}{*}{Knowledge quantity}
& Single & 315 & 57.3\% \\
& Compositional & 235 & 42.7\% \\
\bottomrule
\end{tabular}
\end{table}

\paragraph{Cross-tabulation Analysis.} The cross-tabulation~\ref{tab:education_difficulty} between education level and difficulty  shows that difficulty is not equivalent to education level. Professional-domain cases span low, medium, and high difficulty levels, and school-level knowledge can also appear in medium or high-difficulty generation tasks. This supports the design choice in our metadata schema: education level describes the typical learning stage or knowledge source, whereas difficulty captures the overall complexity of a specific generation task.

\begin{table}[t]
\centering
\small
\caption{Education level $\times$ difficulty in the Generalized Benchmark.}
\label{tab:education_difficulty}
\begin{tabular}{lrrrr}
\toprule
Education Level & Low & Medium & High & Total \\
\midrule
Primary school & 18 & 9 & 0 & 27 \\
Middle school & 120 & 40 & 10 & 170 \\
High school & 52 & 24 & 18 & 94 \\
University & 5 & 1 & 4 & 10 \\
Professional domain & 98 & 102 & 49 & 249 \\
\midrule
Total & 293 & 176 & 81 & 550 \\
\bottomrule
\end{tabular}
\end{table}

\paragraph{Current Limitations and Future Refinement.}
The current benchmark statistics also reveal several directions for refinement. First, the domain distribution should not be interpreted as an unbiased distribution of world knowledge. Our sampling is coverage-driven and evaluability-driven: domains such as Natural Science and Human Civilization and Society are emphasized because they more often support objective visual judgment and structured attribution. Future versions can further balance domain coverage while preserving evaluability.

Second, Real-time Dynamic Events currently contain a relatively small number of cases. This is because time-sensitive knowledge requires explicit timestamping, source verification, and update mechanisms, which are more difficult to stabilize than static knowledge. Future extensions should increase this subset with clearer temporal references and reproducible evaluation criteria.

Third, the Generalized Benchmark contains relatively fewer primary-school and university-level cases. Primary-school knowledge is often visually simple and may overlap with common everyday scenes, making it less diagnostic in open-ended generation unless the prompt introduces richer visual or contextual constraints. University-level knowledge, by contrast, is often abstract, specialized, or difficult to express visually in a stable manner. Future versions can expand both categories by constructing more representative open-ended scenarios while maintaining visual judgeability and clear evaluation criteria.

\section{Structured Judge Protocol and Automatic Evaluation}
\label{app:structured_judge_protocol}

\subsection{VQA Schema Definition}
\label{app:vqa_schema_definition}

In the benchmark defined in this paper, each prompt is decomposed into a variable number of VQA items. The field definitions are as follows:

\paragraph{Question:}
The Question is the core element of each VQA item, describing the visual or semantic aspect being assessed. Questions are always formulated as affirmative declarative statements. When the generated image conforms to the statement, the item is judged as \textbf{\texttt{PASS}}; otherwise, it is judged as \textbf{\texttt{FAIL}}. For questions of type \texttt{OPTIONAL}, when the generated image does not contain the potentially expected content at all, the item is judged as \textbf{\texttt{NA}}.

\paragraph{Judge Type:}
Judge Type specifies the necessity of verification for a given VQA item, and directly governs how VQA-level judgments are aggregated into case-level MOS scores for knowledge correctness and knowledge richness. Table~\ref{tab:definition_judge_type} shows detailed definition of Judge Type

\begin{table}[ht!]
    \centering
    \scriptsize
    \renewcommand{\arraystretch}{1.18}
    \setlength{\tabcolsep}{2pt}
        \caption{Detailed definition of VQA Judge Type. Shows how different Judge Type affect metrics.}
        \label{tab:definition_judge_type}

        \begin{tabular}{lm{9cm}l}
            \toprule
            
            \textbf{Judge Type} & \textbf{Definition} & \textbf{Affected Metrics} \\
    
            \midrule
    
            \textbf{\texttt{REQUIRED}}
            & Mandatory items. Elements explicitly required by the original prompt, or conditions that are physically absolute and implicitly necessary to realize the described scene or knowledge point in the physical world. \textit{Example: a prompt requesting "a submarine underwater" must include an underwater environment; a prompt requesting "an apple falling" must depict the apple airborne with a downward trajectory.}
            & \textbf{Knowledge Correctness}
            \\
    
            \midrule
            
            \textbf{\texttt{OPTIONAL}}
            & Optional items. Elements not explicitly required by the original prompt, but highly likely to be spontaneously generated by T2I models when processing knowledge-intensive tasks—most typically: additional textual annotations, arrows, auxiliary diagrams, or explanatory labels. The judgment logic for this type must follow a "penalize only if attempted incorrectly" principle: items are silently marked \textbf{\texttt{NA}} if absent, and marked FAIL only if generated incorrectly. \textit{Common failure cases: a model asked to plot a function curve that adds axis-point annotations with incorrect coordinates; a model depicting a historical scene whose surrounding environment is anachronistic.}
            & \textbf{Knowledge Richness}
            \\
            \midrule
            \textbf{\texttt{ADDITIONAL}}
            & Supplementary items. Like \textbf{\texttt{OPTIONAL}}, but derived from multimodal understanding of the actually generated image: additional elements spontaneously produced by the T2I model are identified, converted into verification checkpoints that define their correct appearance, and appended as supplementary VQA items.
            & \textit{None}
            \\
    
            \bottomrule
        \end{tabular}
\end{table}

\paragraph{Component:}
Component specifies the linguistic constituent of the prompt that the VQA item corresponds to. We organize components into two kinds and seven component types. VQA items are first divided by whether they assess entities or virtual textual/graphical annotations that are detached from entities, yielding the \textbf{Entity Centric} and \textbf{Annotation Centric} kinds. Each kind is further subdivided as follows in Table~\ref{tab:definition_component}:

\begin{table}[ht!]
    \centering
    \scriptsize
    \renewcommand{\arraystretch}{1.18}
    \setlength{\tabcolsep}{2pt}
    \caption{Detailed definition of VQA Component. Based on semantics within prompt, vqa are first divided into two major kinds, and then further subdivided into 7 component types}
    \label{tab:definition_component}

    \begin{tabular}{llm{8cm}}
        \toprule
        
        \textbf{Kind} & \textbf{Component Type} & \textbf{Definition} \\

        \midrule

        \multirow{4}{*}[-8ex]{Entity Centric}
        & \textbf{\texttt{ENTITY}}
        & Entity existence. Whether an entity composed of general, common-level concepts is present in the image {\textemdash} including specific persons, objects, phenomena, scenes, and backgrounds. Scene text and graphics attached to entities, as well as the overall style and genre of the image, are also included in this type.
        \\

        \cline{2-3}

        & \textbf{\texttt{ENTITY\_TRAIT}}
        & Entity traits. Objective, intrinsic properties of knowledge-intensive entities that remain invariant across viewpoints, external environments, and variant subtypes. These represent the characteristics that a T2I model must correctly and fully render {\textemdash} beyond merely generating a generic entity {\textemdash} in order for the depiction to be recognizable as that specific knowledge-intensive entity. Accordingly, when an entity is classified as knowledge-intensive, an \textbf{\texttt{ENTITY\_TRAIT}} VQA item is added alongside the corresponding generic ENTITY existence item.
        \\
        
        \cline{2-3}
        
        & \textbf{\texttt{ATTRIBUTE\_BINDING}}
        & Attribute binding. An entity's color, shape, size, material, state, and similar properties.
        \\
        
        \cline{2-3}

        & \textbf{\texttt{ENTITY\_RELATION}}
        & Inter-entity relations. Spatial positioning, physical interaction, and temporal ordering among entities.
        \\
        
        \midrule
        
        \multirow{3}{*}[-4ex]{Annotation Centric}
        & \textbf{\texttt{ANNOTATION}}
        & Annotation content and form. Virtual textual or graphical annotations detached from entities, including text of all types, symbols, formulas, charts, figures, and diagrams.
        \\

        \cline{2-3}
        
        & \textbf{\texttt{ANNOTATION\_BINDING}}
        & Entity–annotation binding. The directional or positional relationship between an annotation and its associated entity.
        \\

        \cline{2-3}
        
        & \textbf{\texttt{ANNOTATION\_RELATION}}
        & Multi-annotation relations. Layout and logical relationships among multiple annotations.
        \\

        \bottomrule
    \end{tabular}
\end{table}

\paragraph{Eval Type:}

Eval Type specifies whether a VQA item assesses purely visual rendering quality or world-knowledge grounding.

\begin{itemize}
    \item \textbf{VISUAL:} Pure visual rendering verification. The prompt already provides all information sufficient for visual rendering—including entity types, corresponding attribute states, and inter-entity interaction relationships—and the entities described are generic persons, objects, or scenes without associated world knowledge. Only basic rendering, composition, and layout capabilities are required; no world knowledge is needed.
    \item \textbf{KNOWLEDGE:} World-knowledge grounding verification. Tasks that can only be completed correctly by drawing on the T2I model's world knowledge—either by explicitly requiring entities whose depiction is directly tied to domain-specific world knowledge, or implicitly requiring operations such as knowledge-associated retrieval, state simulation, knowledge annotation, chart rendering, or scientific illustration.
\end{itemize}

\paragraph{Has Text Rendering:}
Has Text Rendering indicates whether a VQA item involves the assessment of text rendering and typesetting capabilities.

This field is set to \textbf{true} if: 

\begin{enumerate}
\renewcommand{\labelenumi}{(\alph{enumi})}
\item Within the Entity Centric kind, there exists a verification of scene text attached to an entity;
\item Within the Annotation Centric kind, there exists a verification of virtual textual annotations detached from entities.
\end{enumerate}

Otherwise, if the VQA item does not involve text rendering or typesetting in any form, the field is set to \textbf{false}. Notably, graphical elements in the scene and virtual graphical annotations are treated as visual rendering—they assess fundamentally different model capabilities from the highly sensitive text rendering and typesetting ability—and are therefore not classified as text rendering. In such cases, Has Text Rendering is likewise set to \textbf{false}.

\subsection{Evaluability-Oriented Annotations}
\label{app:evaluability_oriented_annotations}

We augment each VQA item with evaluation-oriented auxiliary annotations, including:

\paragraph{Knowledge Hint:}
A credibility-oriented guidance field for world-knowledge assessment. It contains: the knowledge operation type assessed by the VQA item, background knowledge provided to the judge model, and detailed judgment criteria—all consolidated into a single field. When the Judge Type is \textbf{\texttt{OPTIONAL}}, the judgment criteria additionally enumerate all plausible outcome scenarios.

\paragraph{Visual Grounding:}
An optional visual anchoring field for knowledge-intensive entities. High-quality reference images are sourced from Wikimedia Commons, retrieved either automatically via agent-invoked tool calls or through manual curation.

Knowledge Hint and Visual Grounding serve complementary roles. The specific retrieval procedure is as follows:

\begin{enumerate}
    \item When the VQA decomposition agent encounters information postdating its knowledge cutoff, uncertain domain-specific knowledge, complex physical or chemical mechanisms, or obscure historical details, it \textbf{MUST} invoke an internet search tool to consult encyclopedic or authoritative sources. After thoroughly understanding the relevant background, the agent extracts the entity's visually discriminative features, core judgment logic, and equivalent variants, and supplements the corresponding VQA item's \verb+knowledge_hint+ field with this information, ensuring the explanation is highly detailed and accurate.
    \item When the content being assessed involves knowledge-intensive entities—such as specific paleontological fossils, particular spacecraft models, rare laboratory instruments, or region-specific traditional costumes—or contains information from 2024 onward, the agent is required to invoke the custom image search tool \verb+wikimedia_image_search+ for visual grounding.
    \item Up to three images are retrieved from the Wikimedia Commons API and passed into the context of the agent model (Gemini 3.1 Pro Preview), which is prompted to assess image quality—verifying that each image is sharp, complete, highly recognizable, and representative, and that it captures the typical intrinsic visual characteristics of the knowledge-intensive entity.
    \item Upon confirmation, the image URL is inserted into the \verb+visual_grounding+ field of the corresponding VQA item, and the \verb+knowledge_hint+ field is enriched based on the image content. If the retrieved images do not meet the requirements, the search query is revised and retried; a maximum of one retry is allowed.
    \item Both \verb+knowledge_hint+ and \verb+visual_grounding+ are then subject to human review, with corrections applied as needed.
\end{enumerate}

The positive contribution of these VQA-level knowledge auxiliary fields to judge performance has been validated through ablation studies, as detailed in Section~\ref{subsec:judge_reliability_validation}

\subsection{Examples}

Table~\ref{tab:example_vqa_checklist_and_visual_grounding} presents a representative instance from the VQA judge protocol, shown with accompanying model generated image and visual grounding.

\begin{tcolorbox}[
    colbacktitle=gray!20,
    coltitle=black,
    colback=gray!5,
    colframe=gray!40,
    title={\textbf{Case ID} \quad \texttt{943aca39-bd1d-5b93-accd-10943434b4bf}},
    fonttitle=\small,
    left=1pt,
    right=1pt,
    breakable
]
    \noindent
    \hspace{2pt}
    \begin{minipage}{5cm}
        \includegraphics[width=5cm]{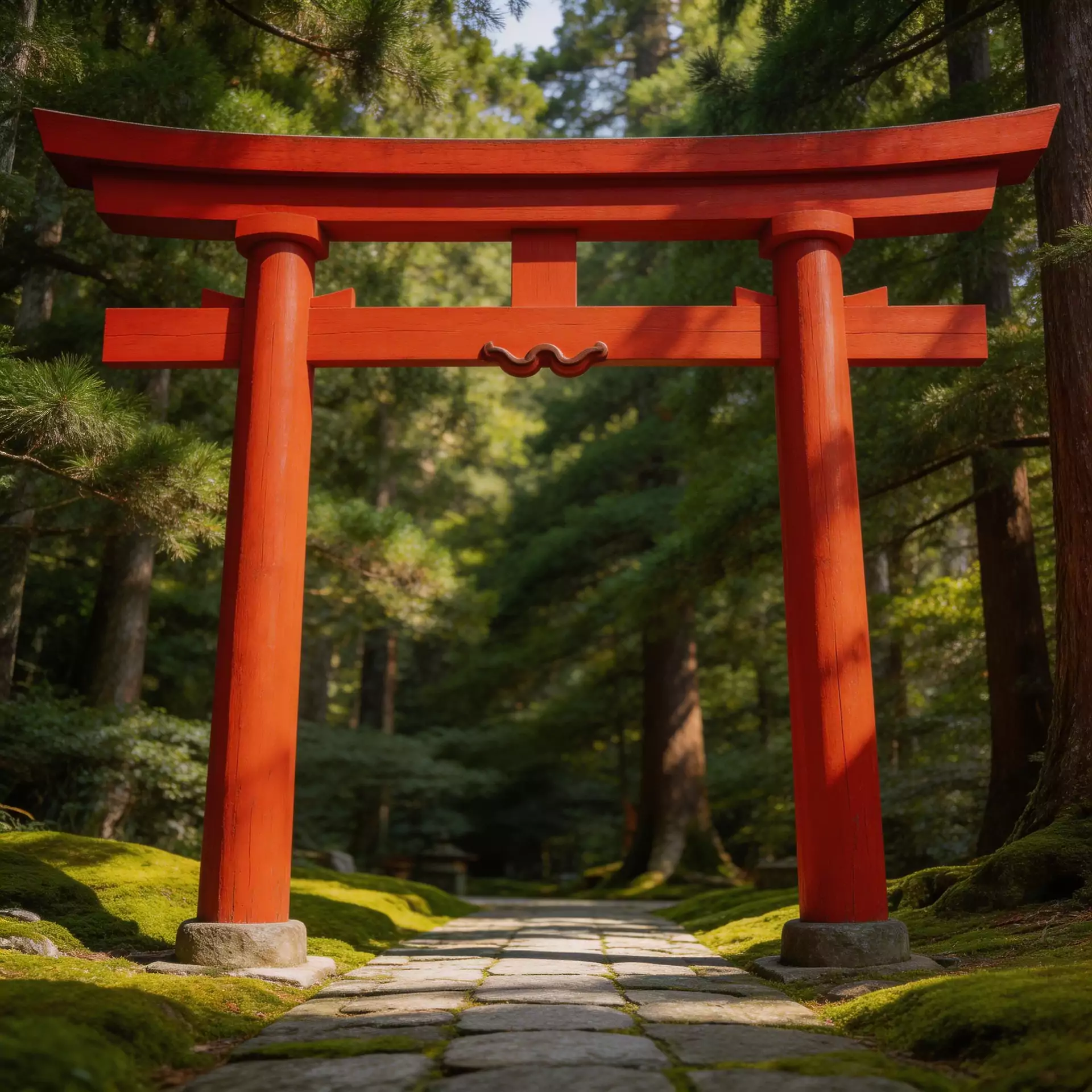}
    \end{minipage}
    \hspace{4pt}
    \begin{minipage}{8cm}
        \textbf{Prompt:} \textit{Draw a torii gate at the entrance of a traditional Japanese shrine.}\\
        \textbf{Category:}
        \begin{enumerate}[leftmargin=*]
            \item \texttt{Human Civilization \& Society}
            \item \texttt{Humanities \& Architecture}
            \item \texttt{Architectural Styles}
            \item \texttt{Japanese Architecture and Key Components}
        \end{enumerate}
    \end{minipage}

    \renewcommand{\arraystretch}{1.18}
    \setlength{\tabcolsep}{4pt}

    \begin{longtable}{
      >{\centering\arraybackslash}p{0.2cm}
      p{4cm}
      p{8.4cm}
    }
        \caption{An example showcasing the VQA judge protocol, shown with accompanying model generated image and visual grounding. Demonstrates the composition of VQA Checklist, its internal logical relationships, and machine-scoring auxiliary information.}
        \\
        \label{tab:example_vqa_checklist_and_visual_grounding}
        \\
        
        \textbf{\#} & \textbf{Question, Labels \& VG.} & \textbf{Knowledge Hint} \\
        \midrule
        
        1
        &
        The image contains a gate-like architectural structure consisting of two upright pillars and a top crossbar.
        \newline
        \tagbox{ENTITY} \tagbox{VISUAL} \tagbox{REQUIRED}
        &
        \textbf{Pure visual assessment.} Irrelevant to the current world knowledge question, it confirms whether the basic entity of an architectural structure similar to an archway or a door frame exists in the image. \\
        \midrule
        2
        &
        The gate-like architectural structure in the image conforms to the visual characteristics of a traditional Japanese torii gate.
        \newline
        \tagbox{ENTITY\_TRAIT} \tagbox{KNOWLEDGE} \tagbox{REQUIRED}
        \newline
        \newline
        \includegraphics[width=4cm]{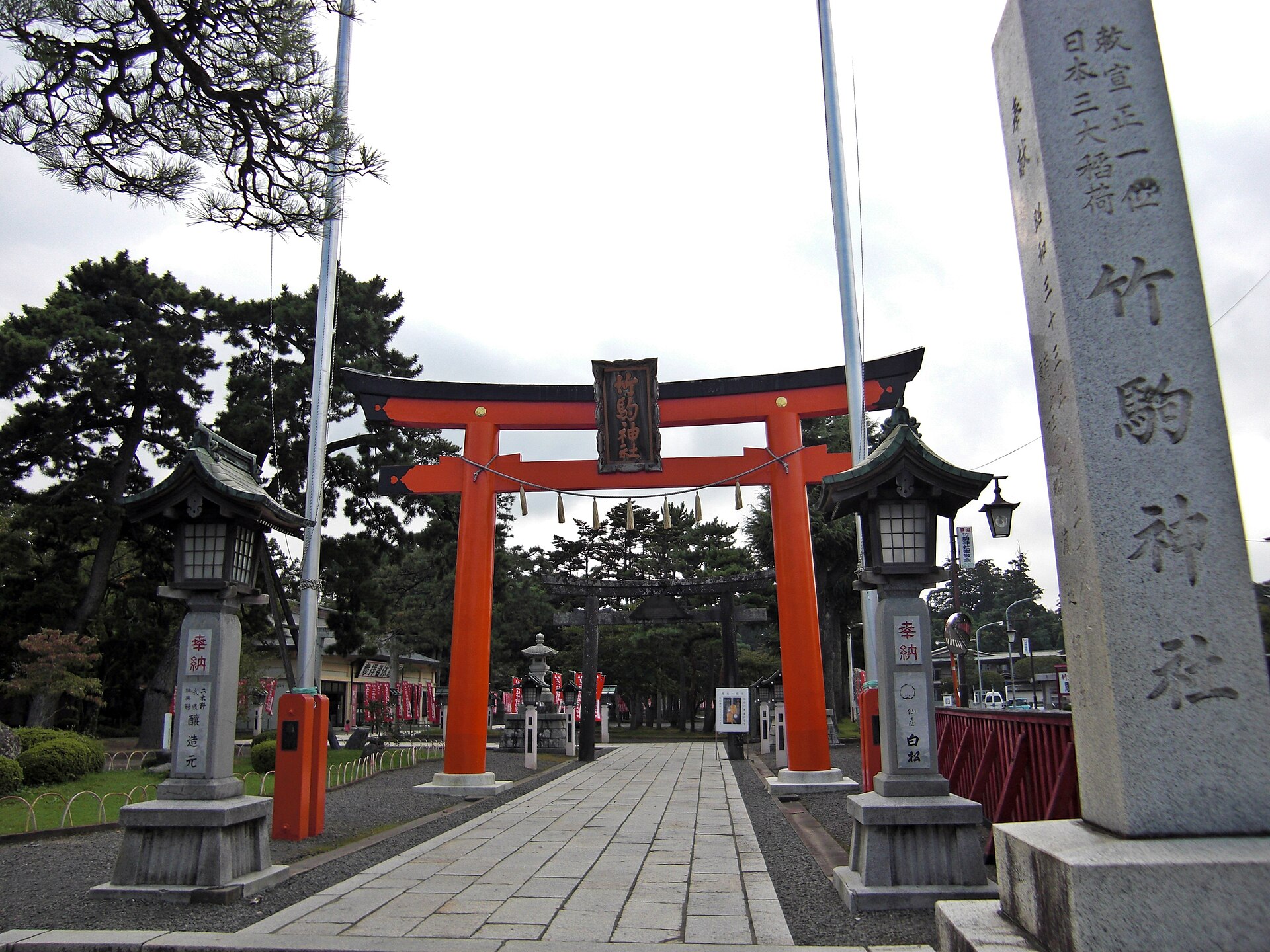}
        &
        \textbf{Explicit knowledge recall verification.} The typical structure of a torii gate includes two upright pillars and a top crossbar (the topmost being the "kasagi", often with a "nuki" passing through the pillars below it), and some have a short vertical wooden piece or plaque ("gakuzuka") between the two crossbars. The style could be a Myojin torii with the ends of the top crossbar curving upwards, or a Shinmei torii with a straight top. The most common color is vermilion (often with black caps or bases), or the natural colors of stone or wood. There may also be a shimenawa (sacred rope) or shide (paper streamers) hung on the crossbar. \textbf{PASS} if the architecture in the image possesses the above characteristics; \textbf{FAIL} if it is drawn as a Chinese-style archway (paifang, including multiple complex flying eaves and brackets, or multi-pillar structures like three doors and four pillars), features Chinese-style coiled dragon sculptures, a Western stone arch, or a simple modern door frame. \\

        &
        \multicolumn{2}{p{12.6cm}}{
            VG. Credits: \texttt{Torii gate of Takekoma-jinja shrine.JPG} by Bachstelze, CC BY-SA 3.0 \url{https://creativecommons.org/licenses/by-sa/3.0>}, via Wikimedia Commons. 
            \newline
            Source: \url{https://commons.wikimedia.org/wiki/File:Torii_gate_of_Takekoma-jinja_shrine.JPG}
        }\\
        
        \midrule

        3
        &
        The torii gate is situated in an entrance environment leading to a sacred precinct, with the background containing elements such as a path and a natural forest or traditional wooden architecture.
        \newline
        \tagbox{ENTITY_RELATION} \tagbox{KNOWLEDGE} \tagbox{REQUIRED}
        &
        \textbf{Implicit environmental knowledge reasoning.} The prompt requires the torii gate to be at the \"entrance of a traditional Japanese shrine\", marking the boundary between the secular world and the sacred precinct. \textbf{PASS} if the image shows the torii gate standing in front of a path (sando), and the surrounding environment has natural forests, stone lanterns, komainu (guardian lion-dogs), or faintly visible traditional Japanese building complexes deep in the background, which reflect the entrance characteristics of a sacred precinct; \textbf{FAIL} if the torii gate is placed right in the middle of an unrelated modern commercial street without any shrine background, placed indoors, or surrounded entirely by European/Chinese-style architecture that does not conform to the setting of a traditional Japanese shrine entrance. \\
        \midrule

        4
        &
        The image contains additional text annotations explaining the name of the torii gate structure or the background of the shrine.
        \newline
        \tagbox{ANNOTATION} \tagbox{KNOWLEDGE} \tagbox{OPTIONAL}
        &
        \textbf{Optional knowledge annotation task.} If there are no additional text annotations in the image (excluding the characters on the plaque inherent to the torii gate itself), evaluate as \textbf{NA} silently. If the model proactively generates introductory text, \textbf{PASS} if the text content is related to "Torii", "Shrine", etc., and is spelled correctly; \textbf{FAIL} if there are obvious spelling errors or if the annotations are inaccurate texts such as "Archway". \\
        \midrule

        5
        &
        The additional text annotations in the image correctly correspond to concepts related to the torii gate or shrine.
        \newline
        \tagbox{ANNOTATION\_BINDING} \tagbox{KNOWLEDGE} \tagbox{OPTIONAL}
        &
        \textbf{Optional knowledge annotation task.} If there are no additional text annotations in the image, evaluate as \textbf{NA} silently. If the model proactively generates text, \textbf{PASS} if the text content correctly describes the corresponding structure or overall concept of the torii gate, and logically matches the image elements; \textbf{FAIL} if the text cannot correspond to the architectural structure, resulting in a logical contradiction. \\

        6
        &
        The central connection between the two horizontal beams of the torii gate is depicted as a short vertical wooden post, a rectangular plaque, or left completely empty.
        \newline
        \tagbox{ENTITY\_TRAIT} \tagbox{ADDITIONAL}
        &
        \textbf{Potential checkpoint for explicit knowledge recall.} In a traditional Japanese torii gate, the exact center between the two horizontal beams (the upper kasagi/shimaki and the lower nuki) typically features a short vertical wooden post (gakuzuka), and sometimes a rectangular plaque (shingaku) bearing the shrine's name is hung on it; in some minimalist styles (such as the shinmei torii), this position lacks any connecting components. \textbf{PASS} if the image depicts this area as a vertical wooden post, a rectangular plaque, or a completely empty space without any structure. \textbf{FAIL} if the model spontaneously generates bizarre wavy or twisted components, or decorations such as European wrought iron patterns or Chinese auspicious cloud/ruyi motifs that completely contradict the common knowledge of traditional Japanese architecture. \\

    \end{longtable}
\end{tcolorbox}
\subsection{Human Ground-Truth Annotation for Judge Validation}
\label{app:vqa_gt_annotation_interface}

\begin{figure}[h]
    \centering
    \includegraphics[
        width=0.88\linewidth,
        trim=10 10 10 10,
        clip
    ]{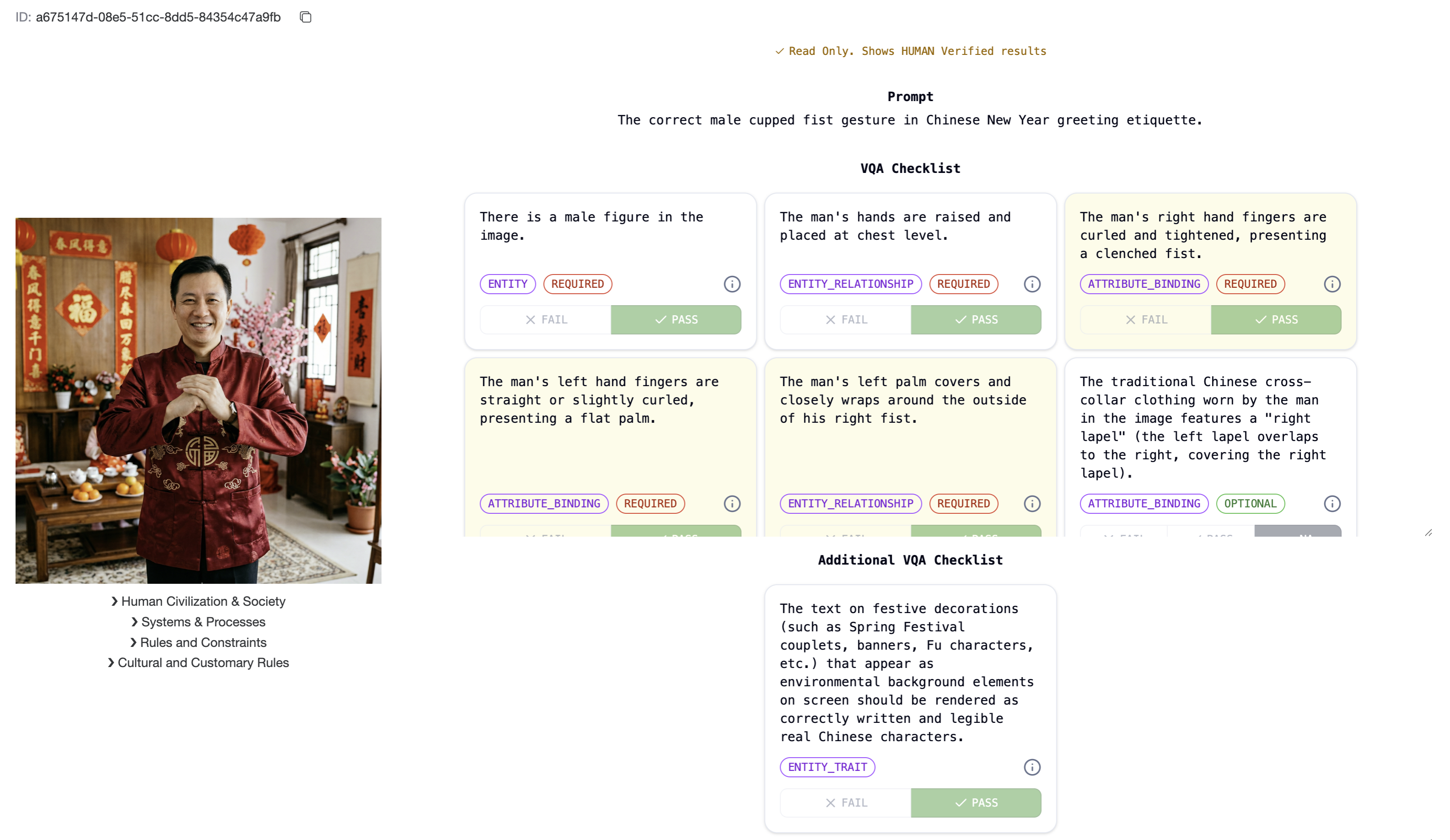}
    \caption{
    Human ground-truth annotation interface for judge reliability validation.
    Annotators label structured VQA items, which are used to compute VQA-level and case-level human--judge agreement.
    }
    \label{fig:vqa_gt_annotation_interface}
\end{figure}

Figure~\ref{fig:vqa_gt_annotation_interface} shows the human ground-truth annotation interface used for judge reliability validation on the Skill-Tree Benchmark.
Annotators review the generated image, the original prompt, and the structured VQA checklist, then assign ground-truth labels to each VQA item.
These annotations are used to compute VQA-level agreement and aggregated case-level agreement between human annotators and the MLLM judge.

\paragraph{Reliability of the human reference labels.}
To assess the consistency of the human references used for judge validation,
five annotators independently labeled the Required VQAs in 50 sampled
Skill-Tree cases, yielding 288 VQA units. We measure agreement among the five
annotators using nominal Krippendorff's $\alpha$, obtaining 0.766.
Table~\ref{tab:human_reference_agreement} reports the distribution of vote
patterns and the corresponding majority labels.

\begin{table}[htbp]
\centering
\small
\caption{Inter-annotator agreement on 288 Required VQA units from 50 sampled
Skill-Tree cases. A vote pattern of 5--0, 4--1, or 3--2 denotes the number of
annotators supporting the majority and minority labels, respectively; it does
not specify whether the majority label is PASS or FAIL.}
\label{tab:human_reference_agreement}
\begin{tabular}{lrrrr}
\toprule
Vote pattern & Count & Percentage & Majority PASS & Majority FAIL \\
\midrule
5--0 & 238 & 82.64\% & 190 & 48 \\
4--1 & 29 & 10.07\% & 23 & 6 \\
3--2 & 21 & 7.29\% & 12 & 9 \\
\midrule
Total & 288 & 100.00\% & 225 & 63 \\
\bottomrule
\end{tabular}
\end{table}

Unanimous agreement is observed on 82.64\% of the VQAs, and 92.71\% receive
agreement from at least four of the five annotators. Only 7.29\% have a
3--2 split. These results support the consistency of the human references
on the sampled cases. Inter-annotator agreement describes consistency among
human labels; it is distinct from the human--judge agreement reported in
Section~\ref{subsec:judge_reliability_validation} and does not by itself
establish the factual correctness of every reference label.

\section{Human Evaluation Protocol for the Generalized Benchmark}
\label{app:human_evaluation_protocol_for_the_generalized_benchmark}

\subsection{Human-Machine Collaborative Evaluation Details}

As described in Section~\ref{sec:skill_tree_generalized_evaluation}, the Generalized Benchmark adopts a human-judge collaborative protocol because open-ended world-knowledge-grounded generation requires both structured knowledge assessment and subjective human perception. This appendix provides the detailed human evaluation dimensions, scoring rubrics, quality-control procedure, and representative scoring examples.

We introduce a comprehensive evaluation encompassing six dimensions (Table~\ref{tab:definition_dimensions}) and design a human-judge collaborative evaluation framework to leverage the respective strengths of both Machine Evaluation and Human Evaluation. 

\begin{table}[ht!]
    \centering
    \scriptsize
    \renewcommand{\arraystretch}{1.18}
    \setlength{\tabcolsep}{2.0pt}
    \caption{Definitions and boundaries of the evaluation dimensions. This taxonomy explicitly disentangles world-knowledge grounding from general semantic and visual metrics. Notably, overall satisfaction is not a linear aggregation of sub-dimensions, but a holistic, task-dependent judgment that captures complex trade-offs like factual vetoes and aesthetic compensation.}
    \label{tab:definition_dimensions}

    \begin{tabular}{m{3cm}m{2.5cm}m{8cm}}
        \toprule
        \textbf{Dimension} & \textbf{Evaluation Dimension}& \textbf{Definition} \\

        \midrule
        \multirow{2}{*}[-3ex]{{World-Knowledge Grounding}}
            & \texttt{Knowledge Correctness}    & The degree to which core knowledge expressions are rendered correctly, reflecting the model's understanding of the relevant world knowledge. \\
            & \texttt{Knowledge Richness}       & The extent to which the model supplements core knowledge points with additional topically relevant knowledge, reflecting the depth and breadth of its world-knowledge understanding. \\
        \midrule
        Semantic
            & \texttt{Non-Knowledge Visual Instruction Following} & The degree to which visual requirements specified in the prompt—beyond knowledge correctness and richness—are fulfilled; i.e., the semantic alignment between the image and the instruction. \\
        \midrule
        \multirow{2}{*}[-4.5ex]{Visual}
            & \texttt{Aesthetics}       & An assessment of the aesthetic quality of the generated image, including lighting, color, composition, and other aesthetic factors, reflecting the model's capacity for visually appealing generation. \\
            & \texttt{Image Quality}    & An assessment of visual quality, including sharpness, structural integrity, detail fidelity, and realism. Factual errors do not, by themselves, lower this score; a lower score requires an independently observable visual defect, such as blurring, structural distortions, or geometric interpenetration. \\
        \midrule
        Holistic
            & Overall Satisfaction & A holistic judgment representing the ultimate user acceptance of the generated image. Rather than a linear weighted average of other dimensions, it reflects a comprehensive, task-context-dependent utility assessment that accounts for complex cross-dimensional interactions \\
        \bottomrule
    \end{tabular}
\end{table}

Overall Satisfaction (OS) is treated as an independent holistic MOS rather than a fixed weighted sum of sub-dimensions. Its interpretation depends on the task context. In strongly knowledge-constrained tasks, such as scientific diagrams or historical/cultural depictions, errors in core knowledge can dominate the final judgment even when the image is visually appealing. In more open-ended or weakly knowledge-constrained tasks, strong image quality and aesthetics may partially compensate for minor non-core factual imperfections. Therefore, OS is scored as a task-dependent utility judgment that reflects the interaction among knowledge correctness, knowledge richness, visual instruction following, image quality, and aesthetics.

\subsubsection{Machine-scored Non-knowledge Visual Instruction Following}
\label{app:non_knowledge_if}

For the Generalized Benchmark, non-knowledge visual instruction following is evaluated by an internal
visual-instruction judge. This dimension is separated from world-knowledge grounding because many prompts
contain explicit visual requirements that do not require external knowledge, such as the number of subjects,
viewpoint, layout, composition, color, style, and other directly specified visual constraints. These requirements
affect the perceived success of a generation, but should not be conflated with knowledge correctness or
knowledge richness.

Given a case $c$, the judge decomposes the non-knowledge visual requirements into $N_{\mathrm{if}}(c)$
atomic visual units. Each unit is assigned a weight $w_j^c$ according to its importance in the prompt, and
the judge predicts whether the unit is correctly satisfied by model $m$. The raw instruction-following score
is computed as a weighted response ratio:
\begin{equation}
S_{\mathrm{if}}(m,c)
=
\frac{
\sum_{j=1}^{N_{\mathrm{if}}(c)} w_j^c v_j^{m,c}
}{
\sum_{j=1}^{N_{\mathrm{if}}(c)} w_j^c
},
\qquad
v_j^{m,c} \in \{0,1\}.
\label{eq:non_knowledge_if}
\end{equation}

Here, $S_{\mathrm{if}}(m,c)$ denotes the raw non-knowledge visual instruction-following score,
$N_{\mathrm{if}}(c)$ is the number of decomposed atomic visual units, $w_j^c$ is the weight of the
$j$-th unit, and $v_j^{m,c}$ indicates whether this unit is correctly satisfied by model $m$.
The score is bounded in $[0,1]$, where a larger value indicates better adherence to explicitly specified
visual requirements.

To make this machine-scored dimension comparable with human-rated MOS dimensions, we calibrate the raw
score to the same MOS scale:
\begin{equation}
\mathrm{MOS}_{\mathrm{if}}(m,c)
=
1 + 4 \cdot S_{\mathrm{if}}(m,c).
\label{eq:non_knowledge_if_mos}
\end{equation}

This linear calibration maps $S_{\mathrm{if}}=0$ to MOS 1 and $S_{\mathrm{if}}=1$ to MOS 5. We use this
calibrated score only as a comparable evaluation dimension in the human--judge collaborative protocol; it is
not used as a fixed component in the overall satisfaction score. Overall satisfaction remains an independent
holistic human MOS that reflects task-dependent trade-offs among knowledge correctness, knowledge richness,
instruction following, aesthetics, image quality, and other perceptual factors.

\subsubsection{Judge-derived MOS Conversion for Knowledge Scores}
\label{subsubsec:judge_derived_mos_conversion}

In the Generalized Benchmark, structured judge scores and human-rated MOS dimensions are reported on different numerical scales. The MLLM judge produces normalized scores in $[0,1]$ for knowledge correctness and knowledge richness, while human annotators provide MOS ratings on a discrete 1--5 scale. To make these dimensions comparable in joint analysis and visualization, we linearly convert the judge-derived knowledge scores to the MOS scale.

For a model $m$ and a generalized case $c$, the judge-derived knowledge correctness score $S_{\mathrm{kc}}(m,c)$ is computed from Required VQAs detailed in \eqref{eq:generalized_knowledge_correctness}, while the knowledge richness score is computed from Optional VQAs detailed in \eqref{eq:generalized_knowledge_richness}.

We then convert each normalized judge score $S_x(m,c) \in [0,1]$ to a judge-derived MOS value using a linear mapping:
\begin{equation}
\mathrm{MOS}^{\mathrm{judge}}_{x}(m,c)
=
1 + 4 \cdot S_x(m,c),
\qquad
x \in \{\mathrm{kc}, \mathrm{kr}\}.
\label{eq:judge_mos_conversion}
\end{equation}
Thus, a score of 0 corresponds to MOS 1, and a score of 1 corresponds to MOS 5. This conversion does not change the ranking induced by the original judge scores; it only places judge-derived knowledge dimensions and human-rated dimensions on a common 1-5 scale.

For model-level reporting, we average the converted MOS values over all generalized cases:
\begin{equation}
\overline{\mathrm{MOS}}_{x}^{\mathrm{judge}}(m)
=
\frac{1}{|\mathcal{C}_{\mathrm{gen}}|}
\sum_{c \in \mathcal{C}_{\mathrm{gen}}}
\mathrm{MOS}_{x}^{\mathrm{judge}}(m,c),
\qquad
x \in \{\mathrm{kc}, \mathrm{kr}\}.
\label{eq:model_level_judge_mos}
\end{equation}
Human-rated dimensions, including image quality, aesthetics, and overall satisfaction, are collected directly on the 1--5 MOS scale and averaged separately. We emphasize that the converted judge-derived MOS values are used only for scale alignment and joint analysis, not as substitutes for human MOS. In particular, overall satisfaction is kept as an independent human judgment rather than computed as a fixed weighted sum of knowledge correctness, knowledge richness, image quality, and aesthetics.

\subsection{Human Evaluation Scoring Rubrics}
\label{app:human_evaluation_scoring_rubrics}

Annotators assign an integer rating from 1 to 5 separately for
image quality, aesthetics, and overall satisfaction, using the
anchors in Table~\ref{tab:human_mos_rubrics}.

\paragraph{Applying the dimension definitions.}
Ratings follow the dimension definitions in
Table~\ref{tab:definition_dimensions}.
For image quality, factual errors alone do not lower the rating;
a lower rating requires an independently observable visual defect,
such as blurring, structural distortion, or geometric
interpenetration.
Aesthetics assesses visual appeal and coherence within the intended
visual form, while compliance with explicitly requested visual
attributes is assessed under non-knowledge visual instruction
following.
Overall satisfaction remains a holistic, task-dependent judgment
rather than a fixed weighted combination of dimension scores.

\begin{table}[t]
\centering
\small
\caption{Five-point rubrics for human evaluation.
The descriptions provide rating anchors rather than a checklist
whose conditions must all be satisfied.
Overall satisfaction is rated independently and is not computed
as a fixed weighted sum of the other dimensions.}
\label{tab:human_mos_rubrics}
\renewcommand{\tabularxcolumn}[1]{m{#1}}
\begin{tabularx}{\linewidth}{
    >{\centering\arraybackslash}m{0.08\linewidth}
    >{\raggedright\arraybackslash}X
    >{\raggedright\arraybackslash}X
    >{\raggedright\arraybackslash}X
}
\toprule
Score
& Image Quality
& Aesthetics
& Overall Satisfaction \\
\midrule

\shortstack[c]{\textbf{1}\\Bad}
& Severe blur, distortion, fragmentation, or structural collapse
makes the main subjects or essential visual information
unrecognizable. The image cannot support basic comprehension.
& Severely disorganized composition, conflicting colors, or
incoherent visual treatment. The image lacks a clear hierarchy
and produces a strongly unfavorable visual impression.
& The task is fundamentally unfulfilled. Severe core-knowledge
errors, missing essential content, or overwhelming visual
defects make the result unusable. \\

\midrule

\shortstack[c]{\textbf{2}\\Poor}
& Subjects remain recognizable, but substantial deformation,
misalignment, interpenetration, or poorly rendered details
seriously impair visual clarity and usability.
& Weak composition or an unclear focal point, with distracting
clutter, discordant colors, or inconsistent styling.
These issues substantially impair visual appeal.
& The result falls substantially short of the task.
Major knowledge errors, missing supporting information,
instruction violations, or poor visual execution substantially
limit its usefulness. \\

\midrule

\shortstack[c]{\textbf{3}\\Fair}
& Main subjects and structures are discernible and broadly intact.
Noticeable local artifacts, blur, repetition, or weak details
remain, but the image is visually usable.
& Acceptable composition and generally compatible colors and
styling, but limited hierarchy, balance, or refinement.
The overall appearance is adequate but unremarkable.
& The result is usable but only partially fulfills the task.
Knowledge coverage, instruction following, or visual presentation
has noticeable limitations that reduce its overall usefulness. \\

\midrule

\shortstack[c]{\textbf{4}\\Good}
& Clear subjects and most details, with coherent structures and
spatial relations. Only minor local artifacts or simplifications
remain, without materially impairing comprehension.
& Well-organized composition, a clear focal point, and harmonious
visual treatment. Minor weaknesses in spacing, balance, or
refinement do not undermine the favorable overall impression.
& The task is largely fulfilled, with substantially correct core
knowledge, sufficient supporting content, and good visual
execution. Minor omissions or imperfections remain. \\

\midrule

\shortstack[c]{\textbf{5}\\Excellent}
& Crisp, complete rendering with stable details and coherent
geometry. Textures and lighting are appropriate to the visual
form, with no evident distracting artifacts.
& Refined composition, clear visual hierarchy, and cohesive
colors and styling. The presentation is highly effective and
visually appealing for the intended task.
& The task is fulfilled exceptionally well. Core knowledge is
accurate, relevant supporting content is comprehensive, and
instructions are faithfully followed, with strong visual
execution and high overall usefulness. \\

\bottomrule
\end{tabularx}
\end{table}

\paragraph{Task-dependent overall satisfaction.}
In strongly knowledge-constrained tasks, errors in core knowledge
can dominate overall satisfaction even when the image is visually
appealing. In more open-ended tasks, strong visual presentation
may partially compensate for minor, non-core factual imperfections.
Annotators therefore apply the overall-satisfaction anchors in
the context of the task rather than mechanically averaging
dimension scores.

\subsection{Quality Control for Human MOS Evaluation}
\label{app:annotation_quality_control}

To ensure the stability and reproducibility of human MOS evaluation, we adopt a standardized annotation quality-control procedure. Before formal evaluation, annotators receive unified training on the task definition, evaluation dimensions, scoring rubrics, representative examples, and common boundary cases. A pilot annotation round is then conducted on a diverse subset of cases, followed by calibration discussions to resolve disagreements and refine boundary rules.

Annotators are admitted to the formal evaluation only after passing a qualification test against reference annotations.
\paragraph{Independent ratings.}
During formal evaluation, each image is initially rated by three
annotators on image quality, aesthetics, and overall satisfaction,
using the five-point rubrics in Table~\ref{tab:human_mos_rubrics}.
Annotators complete their initial ratings independently and cannot
see one another's scores. The interface provides the structured
judge evidence described in Section~3.2; independence therefore
refers to ratings across annotators rather than blinding to
judge outputs.

\paragraph{Pre-adjudication agreement.}
We assess inter-annotator agreement using the initial ratings,
before discrepancy review or adjudication. The agreement analysis
yields Krippendorff's $\alpha$ of 0.7172. We additionally
characterize disagreements using the range between the highest
and lowest ratings assigned by the three annotators to the same
image on the same dimension.
Table~\ref{tab:mos_pre_adjudication_agreement} summarizes the
distribution of initial rating patterns.

\begin{table}[htbp]
\centering
\small
\caption{Distribution of initial human rating patterns before
discrepancy review or adjudication. Score range denotes the
difference between the highest and lowest scores assigned by
the three annotators to the same image on the same dimension.}
\label{tab:mos_pre_adjudication_agreement}
\begingroup
\renewcommand{\arraystretch}{1.15}
\setlength{\tabcolsep}{8pt}
\begin{tabular}{lcr}
\toprule
\textbf{Initial rating pattern}
& \textbf{Score range}
& \textbf{Percentage} \\
\midrule
Unanimous agreement
& 0
& 57.02\% \\
Adjacent-level disagreement
& 1
& 30.17\% \\
Larger disagreement
& $\geq 2$
& 12.81\% \\
\bottomrule
\end{tabular}
\endgroup
\end{table}

\paragraph{Discrepancy review and aggregation.}
The aggregation and review rules are specified before formal
annotation. Unanimous ratings are accepted directly, and ratings
with a range of one are aggregated according to the predefined
rule. Only rating items with a difference of more than one point
between the highest and lowest scores undergo review and
adjudication. Thus, 87.19\% of the evaluated rating items are
processed without review or adjudication, whereas 12.81\%
undergo this process. Final MOS values are computed according
to the predefined aggregation rule after resolving these
larger disagreements. The reported agreement statistics are
computed from the initial independent ratings and do not
incorporate any changes made during review or adjudication.

\subsection{Representative Scoring Example}

Table~\ref{tab:example_overall_satisfaction_scoring} shows 4 representative Text-to-Image generation pattern with world-knowledge grounding using the same instruction:

\begin{tcolorbox}[
    colbacktitle=gray!20,
    coltitle=black,
    colback=gray!5,
    colframe=gray!40,
    title={\textbf{Representative \texttt{Overall Satisfaction} Patterns}.},
    fonttitle=\small,
    left=1pt,
    right=1pt,
    breakable
]
    \centering
    \renewcommand{\arraystretch}{1.18}
    \setlength{\tabcolsep}{6pt}
    \begin{longtable}{m{4cm}m{3cm}m{5cm}}
        \label{tab:example_overall_satisfaction_scoring}
        \textbf{Example} & \textbf{Scoring} & \textbf{Reason} \\
        \toprule
        \multicolumn{3}{p{13cm}}{\textbf{Case A:} Ideal example: Knowledge is correct} \\
        \hfill Nano Banana 2 \hfill \null \par
            \vspace{2pt}
            \hfill \includegraphics[width=4cm]{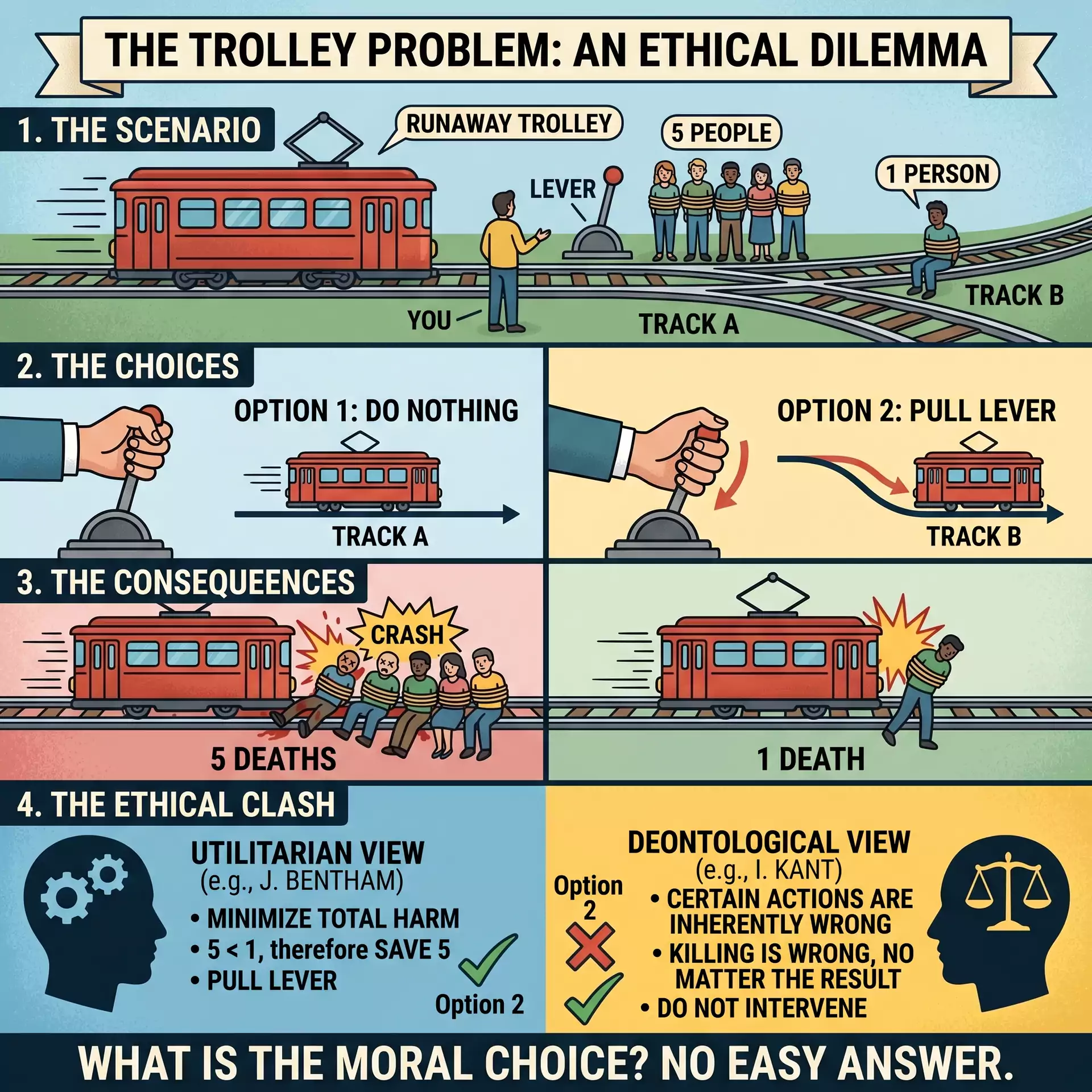} \hfill
        & \begin{itemize}[leftmargin=*]
            \item \texttt{KC.} \tagbox{5\_EXCELLENT}
            \item \texttt{KR.} \tagbox{5\_EXCELLENT}
            \item \texttt{IF.} \tagbox{5\_EXCELLENT}
            \item \texttt{IQ.} \tagbox{5\_EXCELLENT}
            \item \texttt{VA.} \tagbox{5\_EXCELLENT}
            \item \underline{\textbf{\texttt{OS.}}} \tagbox{5\_EXCELLENT}
        \end{itemize}
        &
        \textbf{Reason:} The image perfectly renders the philosophical concept with rich content and impeccable composition, achieving a flawless balance of instruction following and top-tier visual quality. Overall satisfaction: 5.
        \\

        \midrule

        \multicolumn{3}{p{13cm}}{\textbf{Case B:} Knowledge is correct, but knowledge expression is insufficient. Needs to improve visual rendering capability.} \\
        
        \hfill Recraft V4 \hfill \null \par
            \vspace{2pt}
            \hfill \includegraphics[width=4cm]{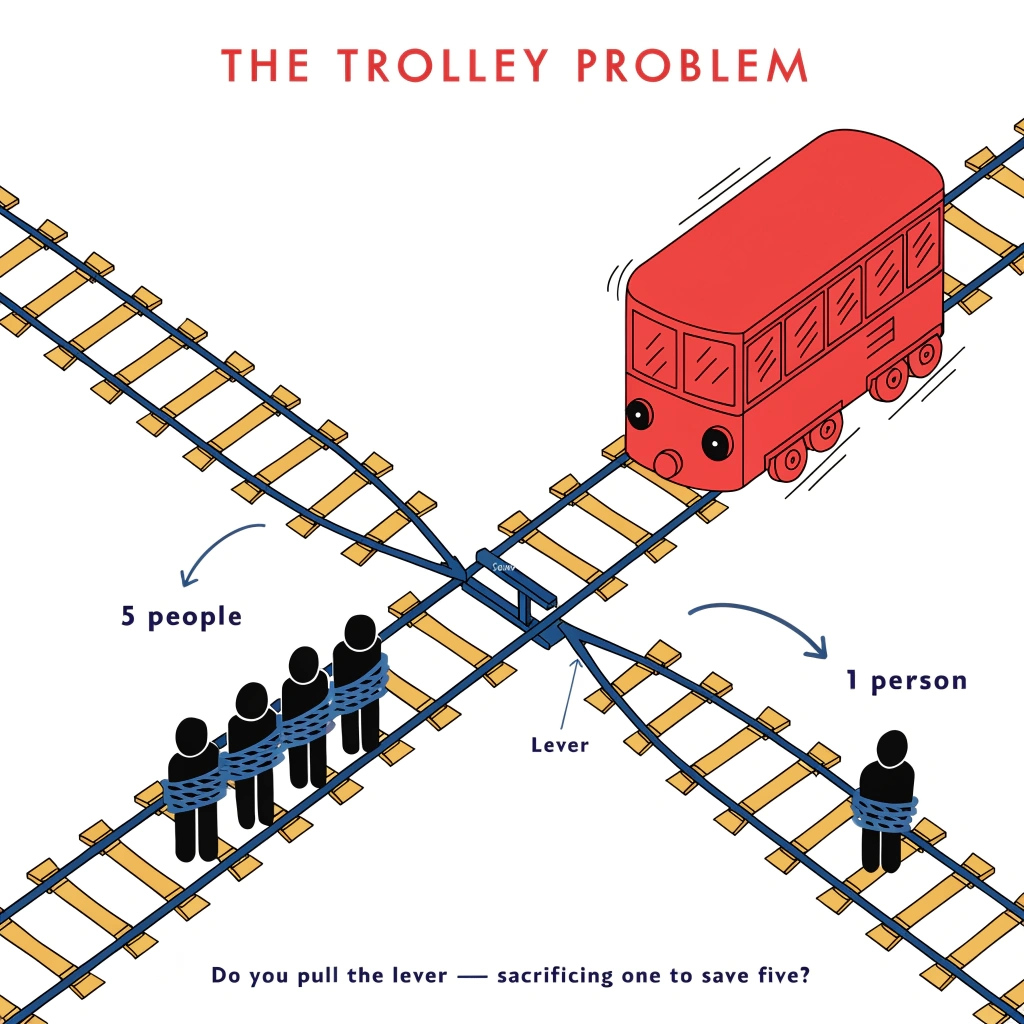} \hfill
        & \begin{itemize}[leftmargin=*]
            \item \texttt{KC.} \tagbox{5\_EXCELLENT}
            \item \texttt{KR.} \tagbox{4\_GOOD}
            \item \texttt{IF.} \tagbox{5\_EXCELLENT}
            \item \texttt{IQ.} \tagbox{4\_GOOD}
            \item \texttt{VA.} \tagbox{3\_FAIR}
            \item \underline{\textbf{\texttt{OS.}}} \tagbox{4\_GOOD}
        \end{itemize}
        &
        \textbf{Reason:} While the knowledge is accurate and instructions are fully followed, limited content richness and average visual aesthetics result in a relatively conventional presentation. Overall satisfaction: 4.
        \\
        
        \midrule
        
        \multicolumn{3}{p{13cm}}{\textbf{Case C:} Knowledge contains critical errors, while aesthetics remain comparatively strong. Needs to improve knowledge grounding and reasoning capability.} \\
        
        \hfill Midjourney v7 \hfill \null \par
                \vspace{2pt}
                \hfill \includegraphics[width=4cm]{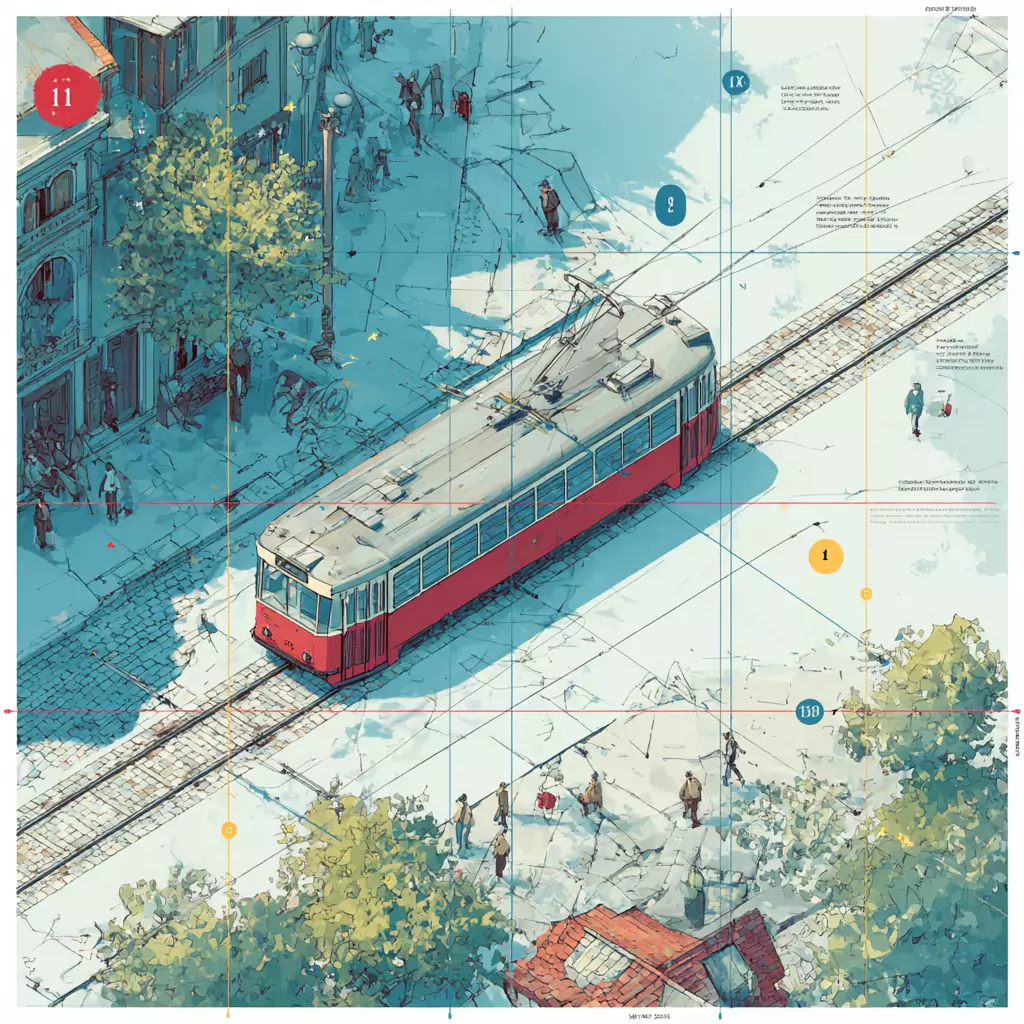} \hfill
        & \begin{itemize}[leftmargin=*]
            \item \texttt{KC.} \tagbox{1\_BAD}
            \item \texttt{KR.} \tagbox{2\_POOR}
            \item \texttt{IF.} \tagbox{2\_POOR}
            \item \texttt{IQ.} \tagbox{3\_FAIR}
            \item \texttt{VA.} \tagbox{4\_GOOD}
            \item \underline{\textbf{\texttt{OS.}}} \tagbox{2\_POOR}
        \end{itemize}
        &
        \textbf{Reason:} The image severely fails in theme-related instruction following yet stands out through its distinctive picture-book aesthetic and high-quality rendering, showcasing a sophisticated artistic sensibility. Overall satisfaction: 2.
        \\
        
        \midrule

        \multicolumn{3}{p{13cm}}{\textbf{Case D:} Knowledge is largely correct, but basic visual instruction following is poor. Needs to visual instruction following capability.} \\
        \hfill FLUX.2 Pro \hfill \null \par
            \vspace{2pt}
            \hfill \includegraphics[width=4cm]{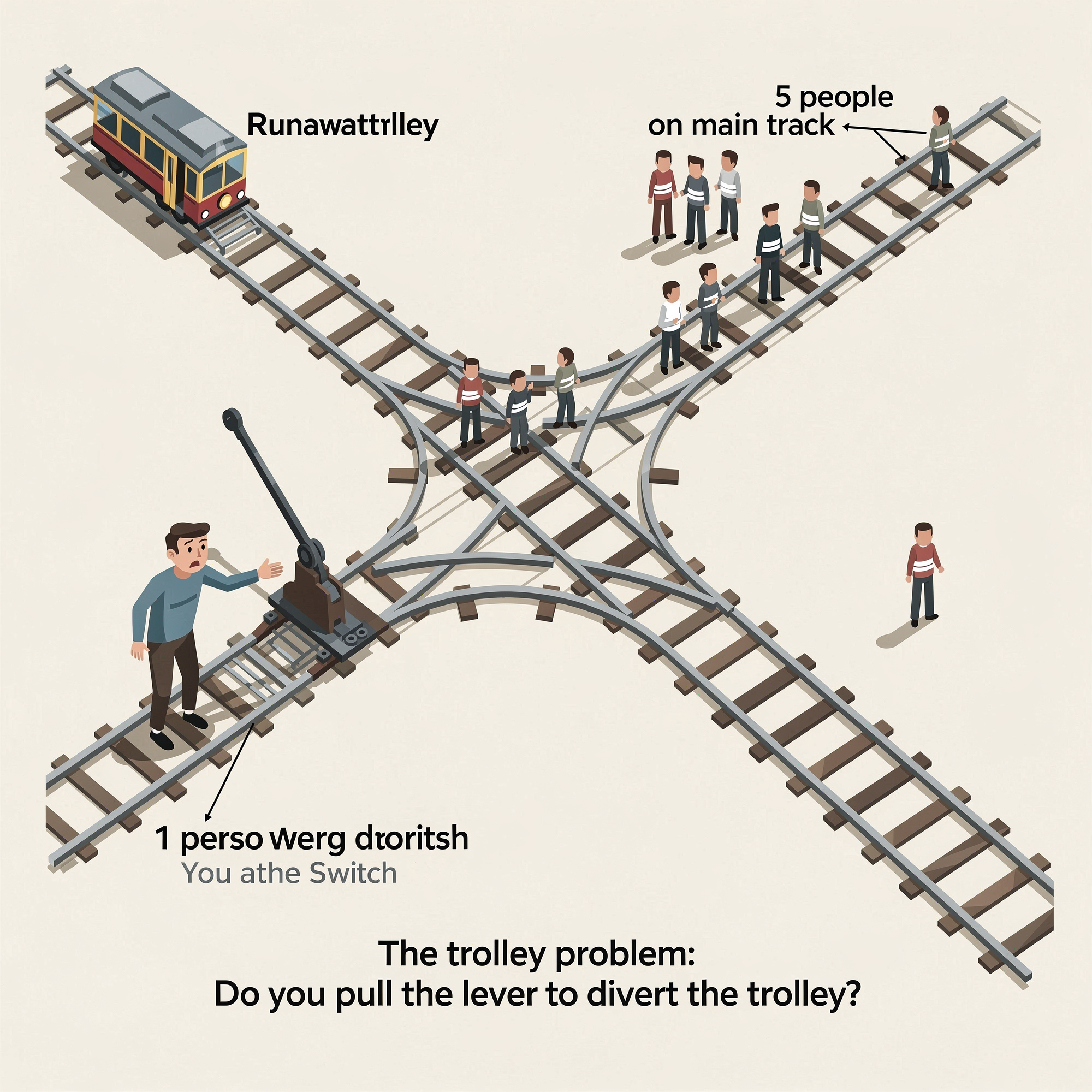} \hfill
        & \begin{itemize}[leftmargin=*]
            \item \texttt{KC.} \tagbox{3\_FAIR}
            \item \texttt{KR.} \tagbox{3\_FAIR}
            \item \texttt{IF.} \tagbox{3\_FAIR}
            \item \texttt{IQ.} \tagbox{3\_FAIR}
            \item \texttt{VA.} \tagbox{2\_POOR}
            \item \underline{\textbf{\texttt{OS.}}} \tagbox{3\_FAIR}
        \end{itemize}
        &
        \textbf{Reason:} The core knowledge is rendered adequately, but the cluttered background elements and minor glyph imperfections undermine the overall visual quality. Overall satisfaction: 3.
        \\
        \midrule
    \end{longtable}
\end{tcolorbox}

\section{Attribution Framework}
\label{app:attribution_framework}

The main experiments report model performance and key findings across the Skill-Tree Benchmark, the Generalized Benchmark, and the General-Domain T2I Benchmark. This appendix summarizes how these findings can be attributed through structured evaluation signals and case-level metadata. The goal is not to simply list failure cases, but to explain why models fail and where such failures are concentrated.

\subsection{Two-Level Attribution Framework}
\label{app:two_level_attribution}

We perform attribution at two levels. \textbf{Signal-level attribution} identifies the source of failure from intermediate evaluation signals, such as Required VQA failures, Optional VQA coverage, non-knowledge visual instruction following, annotation-related failures, and human MOS dimensions. \textbf{Slice-level attribution} then groups these signals by taxonomy and task metadata, such as L1--L4, education level, knowledge operation type, difficulty, and knowledge quantity.

\subsection{Attribution on the Skill-Tree Benchmark}
\label{app:skill_tree_attribution}

For the Skill-Tree Benchmark, attribution focuses on atomic knowledge-point mastery. We first identify low-performing L4 atomic knowledge points using case-level correctness. We then inspect failed Required VQAs and group them by component type, such as ENTITY, ENTITY\_TRAIT, ATTRIBUTE\_BINDING, ENTITY\_RELATION, ANNOTATION, ANNOTATION\_BINDING, and ANNOTATION\_RELATION. Finally, we place these L4 points back under their corresponding L3 branches to distinguish local atomic knowledge gaps from broader branch-level weaknesses.

This process helps determine whether a model fails because it cannot recall the target knowledge concept, because it renders the relevant entity with incorrect attributes or relations, or because it fails to express the required knowledge through annotations, formulas, arrows, or text-image bindings.

\begin{figure}[t]
\centering

\begin{minipage}[t]{0.49\linewidth}
\centering
\includegraphics[
    width=\linewidth,
    trim=15 0 0 0,
    clip
]{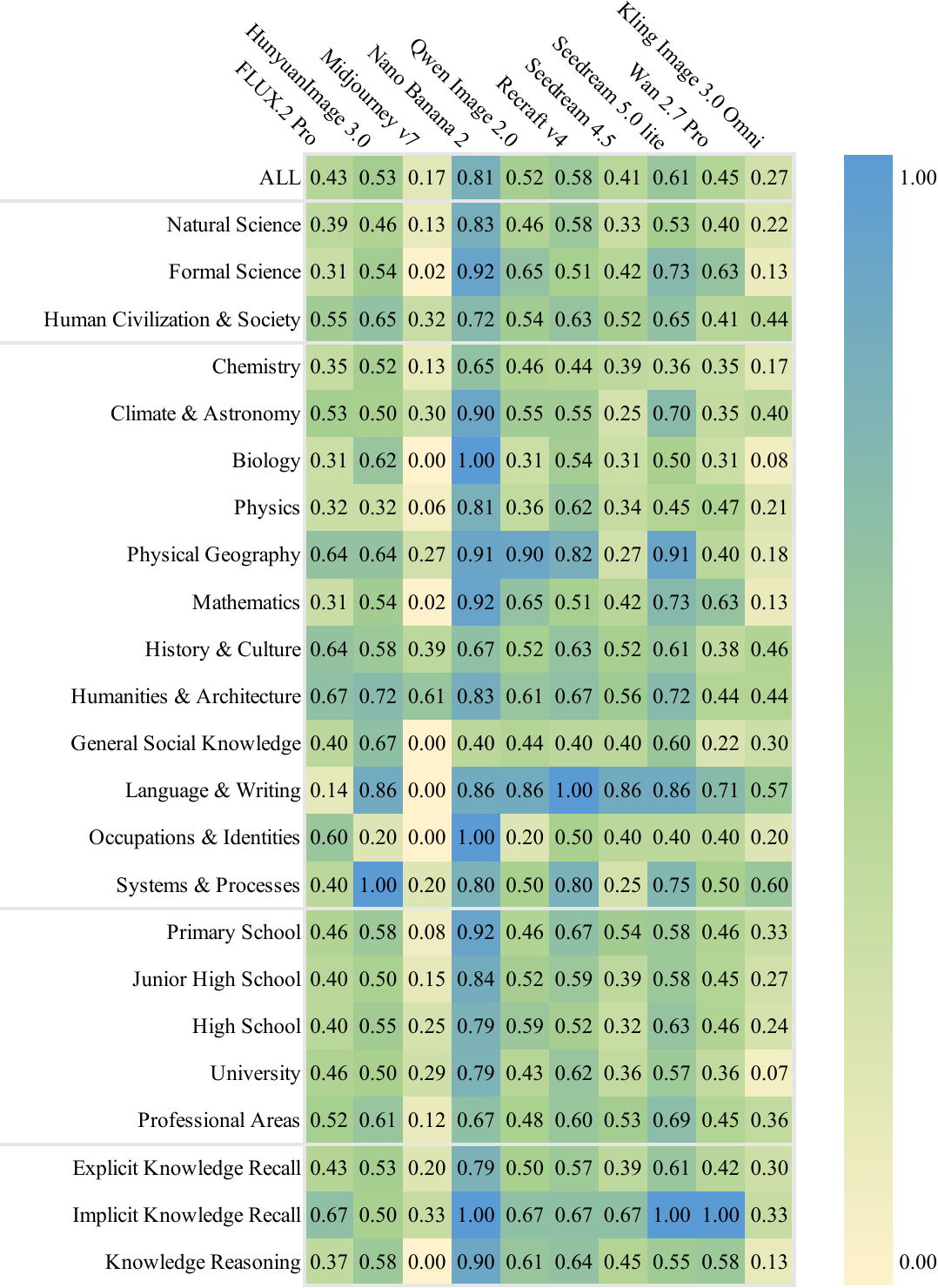}

\vspace{0.3em}
\textbf{(a) Full Skill-Tree case-level correctness.}
\end{minipage}
\hfill
\begin{minipage}[t]{0.49\linewidth}
\centering
\includegraphics[
    width=\linewidth,
    trim=25 0 0 0,
    clip
]{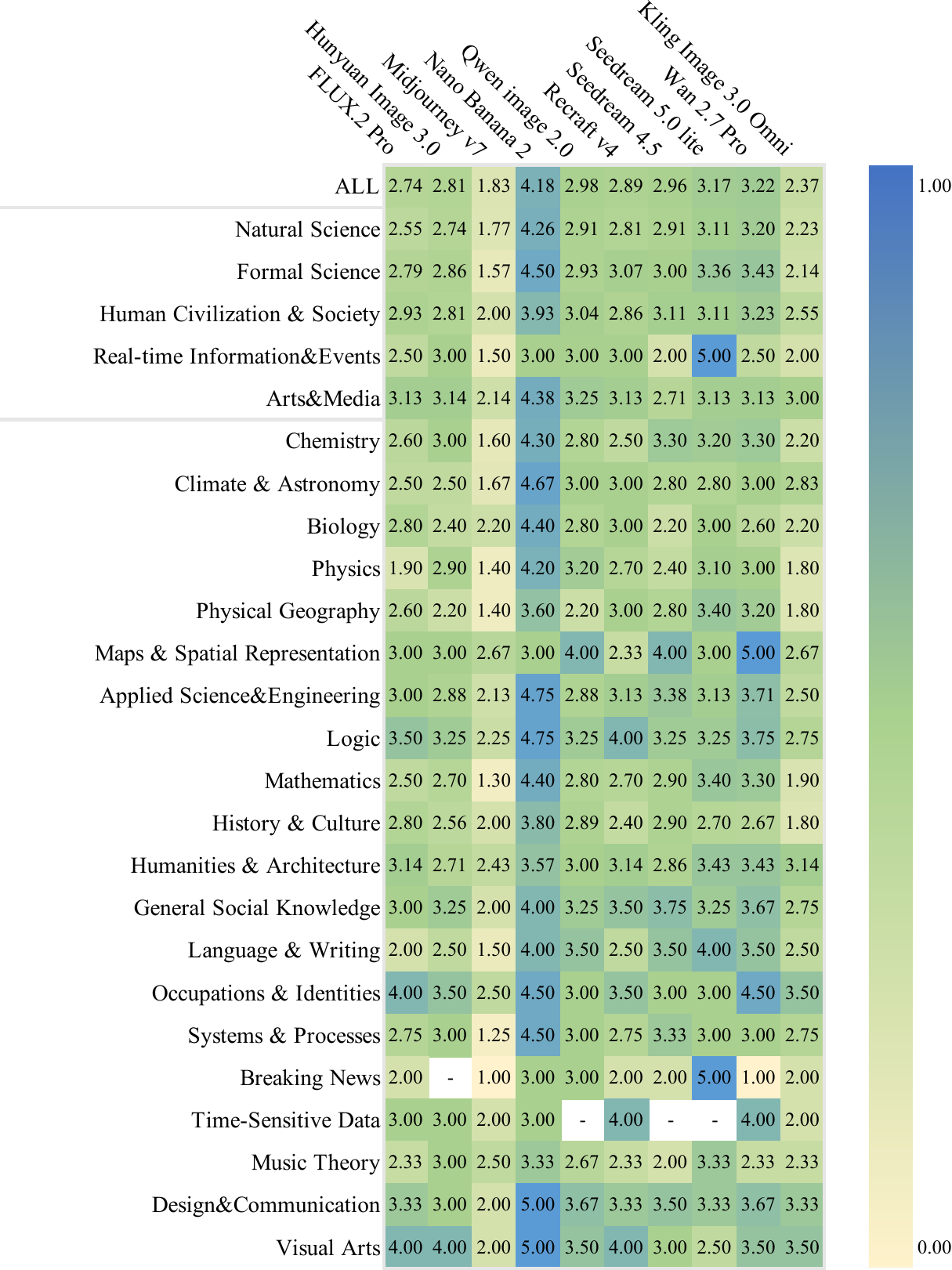}

\vspace{0.3em}
\textbf{(b) Generalized Benchmark MOS.}
\end{minipage}

\vspace{0.5em}
\caption{
Full heatmaps for cross-view diagnostic analysis.
(a) Skill-Tree case-level correctness across models, taxonomy slices, education levels, and knowledge operation types.
(b) Generalized Benchmark overall satisfaction MOS across models and taxonomy slices.
}
\label{fig:full_skill_tree_and_generalized_heatmaps}
\end{figure}

Figure~\ref{fig:full_skill_tree_and_generalized_heatmaps}(a) provides the full Skill-Tree case-level heatmap across models, L1/L2 taxonomy slices, education levels, and knowledge operation types.
This expanded view complements the compressed main-text visualization by showing how model performance varies across fine-grained knowledge domains and task metadata.
The heatmap is intended for diagnostic inspection rather than for deriving a single global ranking: different models may exhibit different strengths across taxonomy branches, education levels, and knowledge operation types.

\subsubsection{VQA-Level Component Failure Analysis}
\label{app:vqa_component_failure_analysis}

Case-level correctness identifies whether a Skill-Tree case succeeds or fails, but it does not reveal which part of the visual knowledge expression caused the failure.
To localize failure sources, we further analyze VQA-level pass rates by component type.
Each Required VQA is grouped according to its component label, including entity-level checks, attribute and binding checks, relation checks, and annotation-related checks.
This allows us to distinguish failures caused by missing or incorrect entities from failures caused by incorrect relations, attribute binding, textual labels, formulas, arrows, or annotation--image bindings.

Figure~\ref{fig:skill_tree_diagnostic_heatmaps}(b) reports the VQA-level component heatmap, which localizes Skill-Tree failures beyond case-level correctness.
We observe three main patterns.

First, entity-centered components generally obtain higher pass rates than relation- and annotation-centered components.
This suggests that models often recognize the relevant visual subject, but this alone is insufficient for knowledge correctness.
For example, a model may draw the correct object or scene while failing to express the required physical relation, structural constraint, or symbolic convention.

Second, annotation-related components form a major bottleneck.
Components such as \texttt{ANNOTATION}, \texttt{ANNOTATION\_BINDING}, and \texttt{ANNOTATION\_RELATION} require text, formulas, arrows, labels, charts, or callouts to be both visually legible and semantically bound to the correct image regions.
Failures in these components are especially harmful for world-knowledge-grounded generation, because many scientific, mathematical, and explanatory prompts rely on annotations as part of the knowledge expression rather than as decorative elements.

Third, model differences are more pronounced at the component level than at the entity level.
Some models can pass many entity-related checks but still fail relation or annotation checks, indicating that their bottleneck lies not in generating relevant objects, but in organizing structured knowledge evidence.
This supports the diagnostic value of our VQA decomposition: it separates entity generation failures, attribute-binding failures, relational reasoning failures, and annotation-rendering or annotation-binding failures, instead of collapsing them into a single case-level score.

\subsection{Attribution on the Generalized Benchmark}
\label{app:generalized_attribution}

For the Generalized Benchmark, attribution focuses on comprehensive failure decomposition in open-ended generation scenarios. For each case, we inspect judge-derived signals, including knowledge correctness, knowledge richness, and non-knowledge visual instruction following, together with human-rated MOS dimensions, including image quality, aesthetics, and overall satisfaction.

We then map the failure back to task metadata, including L3 branch, knowledge operation type, education level, difficulty, knowledge quantity, and involved L4 atomic knowledge points when available. This allows us to analyze whether low overall satisfaction is mainly caused by incorrect knowledge, incomplete knowledge expression, poor instruction following, weak image quality, or aesthetic issues, and whether such failures are concentrated in particular task slices.

\subsection{Reporting Formats}
\label{app:attribution_reporting_formats}

Table~\ref{tab:attribution_reporting_formats} summarizes the common reporting formats used for attribution analysis. These formats are not intended to introduce additional metrics; instead, they organize existing evaluation signals into interpretable views for inspecting where and why models fail.

\begin{table}[t]
\centering
\small
\caption{Common reporting formats for attribution analysis.}
\label{tab:attribution_reporting_formats}
\renewcommand{\arraystretch}{1.25}
\setlength{\tabcolsep}{5pt}

\begin{tabularx}{0.96\linewidth}{
    >{\raggedright\arraybackslash}p{0.18\linewidth}
    >{\raggedright\arraybackslash}p{0.30\linewidth}
    >{\raggedright\arraybackslash}X
}
\toprule
\textbf{Benchmark View} & \textbf{Reporting Format} & \textbf{Purpose} \\
\midrule

\multirow{4}{*}{Skill-Tree}
& L1/L2/L3/L4 hierarchical table
& Shows accuracy across domains, branches, and L4 atomic knowledge points. \\
\cline{2-3}

& VQA component heatmap
& Identifies dominant failure types, such as entity, relation, annotation, or binding errors. \\
\cline{2-3}

& L4-under-L3 heatmap
& Reveals which L4 atomic knowledge points fail within the same L3 branch. \\
\cline{2-3}

& Annotation-related failure table
& Measures whether text, formulas, arrows, labels, or callouts form a bottleneck for knowledge expression. \\

\midrule

\multirow{4}{*}{Generalized}
& Dimension-wise score table
& Reports knowledge correctness, knowledge richness, non-knowledge visual instruction following, and human MOS dimensions. \\
\cline{2-3}

& Slice-based heatmap
& Analyzes failures by L3 branch, difficulty, knowledge operation type, education level, or knowledge quantity. \\
\cline{2-3}

& Dimension-vs-overall plot
& Examines how sub-dimensions relate to overall satisfaction. \\
\cline{2-3}

& Representative case card
& Presents a multi-dimensional attribution path for individual cases. \\

\bottomrule
\end{tabularx}
\end{table}

\section{Empirical Analysis of Automatic Evaluation Failures}
\label{app:empirical_analysis_of_automatic_evaluation_failure}

\subsection{Qualitative Analysis of Failure Patterns}

\paragraph{Visual reasoning failures.} The MLLM judge exhibits insufficient visual reasoning capability, manifesting in the following sub-types: 
\begin{itemize}[leftmargin=*]
    \item Incorrect reasoning about spatial orientation and coordinates involving human figures;
    \item Misrecognition of textual and graphical content;
    \item Errors in structural understanding.
\end{itemize}

\paragraph{Alignment failures.} The judgment criteria of the MLLM judge deviate from human evaluation standards in two opposing directions:
\begin{itemize}[leftmargin=*]
    \item Over-strict judgment, most notably a systematic tendency to penalize handwritten or rendered text content;
    \item Over-lenient judgment, where the presence of duplicate or redundant elements is overlooked—the judge passes a criterion as long as at least one conforming instance is found, regardless of additional incorrect instances.
\end{itemize}

Representative failure cases are shown in Table~\ref{tab:mllm_judge_failure_patterns}. For each case, we report the issue category, the VQA question, the judge model's reasoning, and the identified failure type.

\begin{tcolorbox}[
    colbacktitle=gray!20,
    coltitle=black,
    colback=gray!5,
    colframe=gray!40,
    title={\textbf{Failure Patterns} of automatic evaluation system\\ It has been shown that the MLLM's visual perception and reasoning performance significantly impact the evaluation of Text-to-Image (T2I) generation.},
    fonttitle=\small,
    left=1pt,
    right=1pt,
    breakable
]
    \label{tab:mllm_judge_failure_patterns}
    \centering
    \renewcommand{\arraystretch}{1.18}
    \setlength{\tabcolsep}{4pt}
    \begin{longtable}{m{4cm}m{9cm}}
        \textbf{Samples} & \textbf{VQA, Reasoning, and Issue } \\
        \toprule
        \includegraphics[width=4cm]{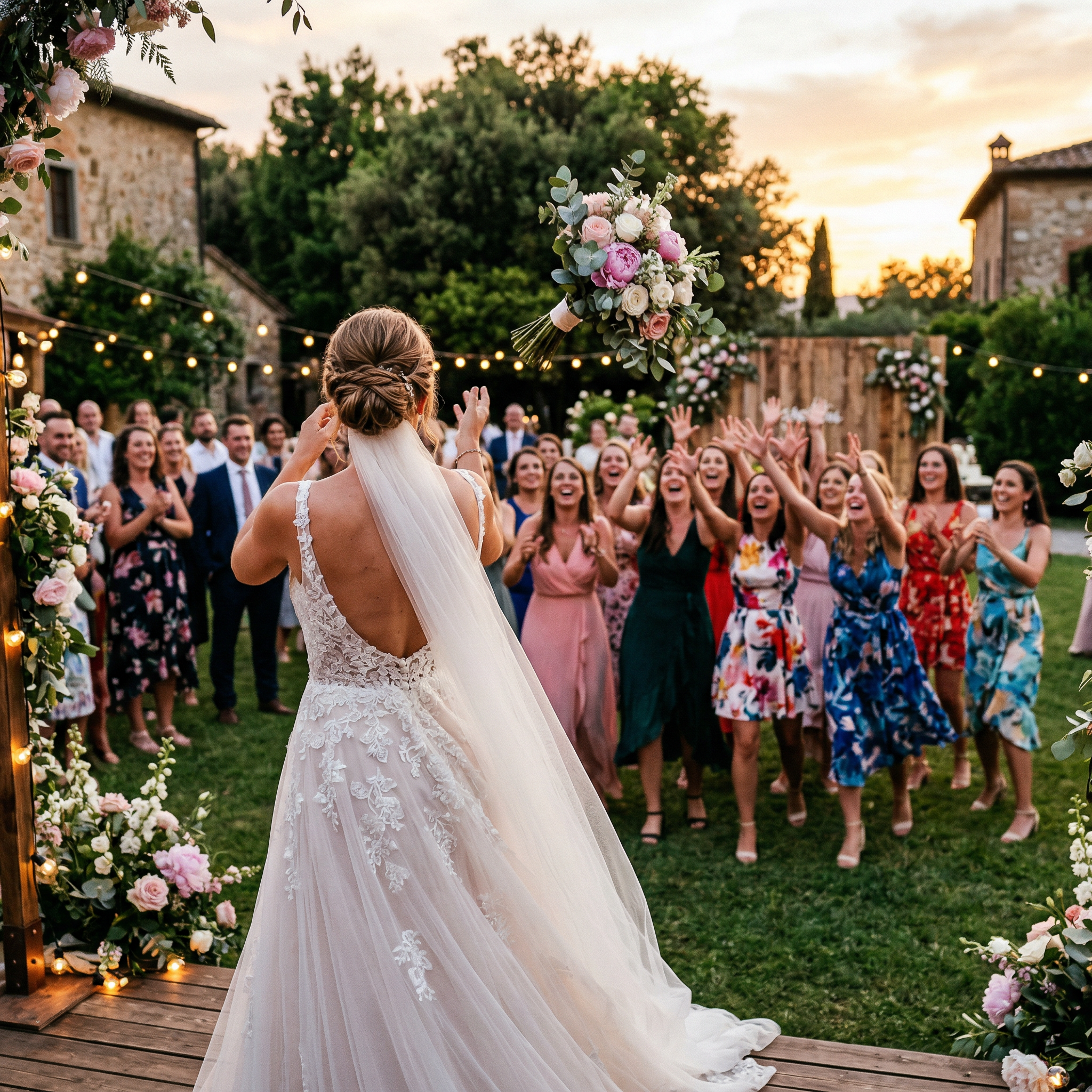}
        & 
        \textbf{ID:} \texttt{c8ae24b5-f58c-549a-8882-2638730408d0}
        \newline
        \textbf{VQA:} \textit{The bride's back is facing the group of women.}
        \newline
        \textbf{Reasoning:} In the image, \uline{the bride wearing a white wedding dress has her back to the audience}, while multiple women in the background are facing toward the bride, \uline{clearly demonstrating the spatial relationship of the bride standing with her back to the crowd.} The pass condition is satisfied.
        \newline
        \textbf{Issue:} Misidentification of real-world viewers as in-scene audience; confusion of spatial relationships.
        \\
        \midrule
        \includegraphics[width=4cm]{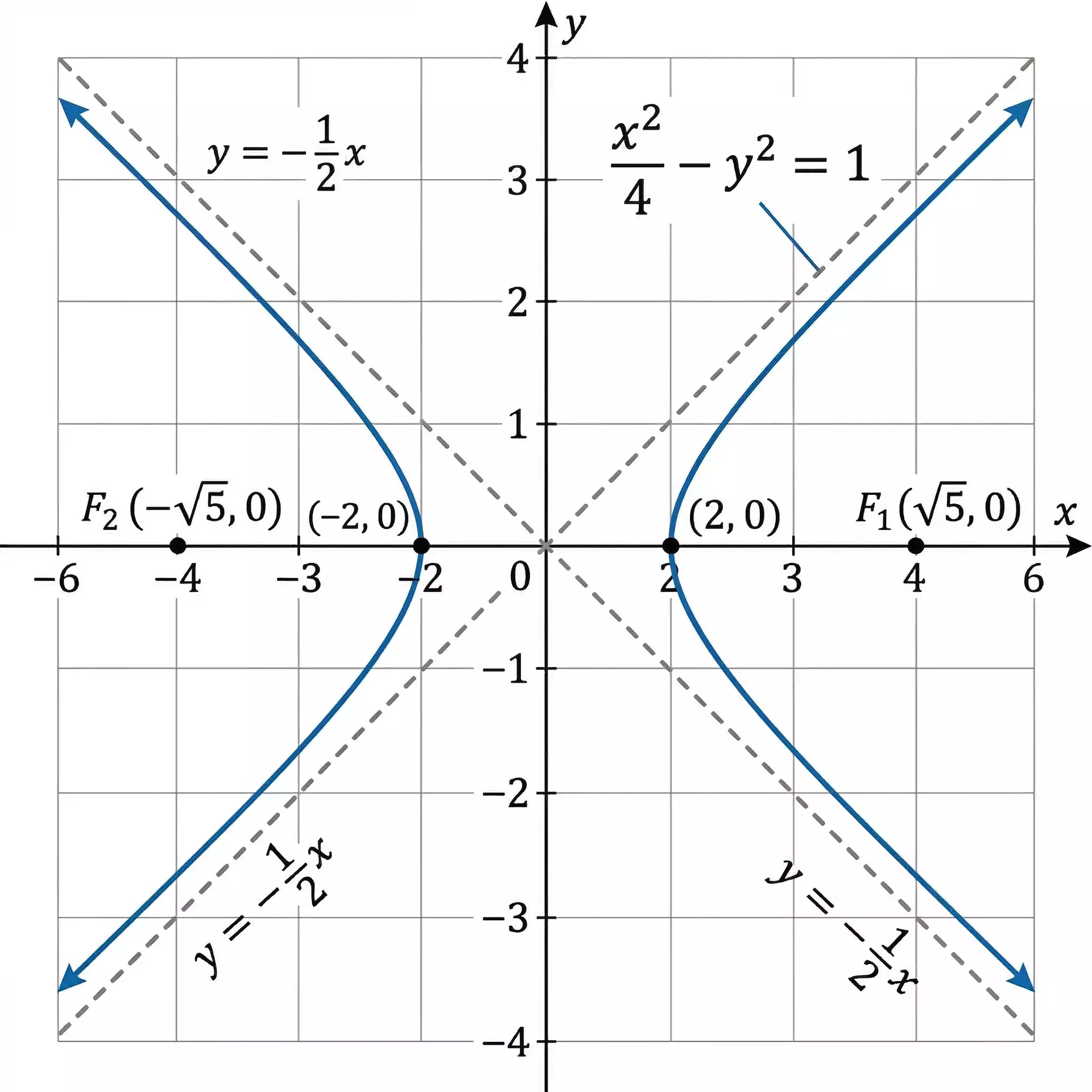}
        & 
        \textbf{ID:} \texttt{0b4b9e8f-e4e7-5265-a3d7-92576e68347d}
        \newline
        \textbf{VQA:} \textit{The annotations for the foci in the image accurately provide the coordinate values $\left(\sqrt{5},\, 0\right) \quad$ and $\quad \left(-\sqrt{5},\, 0\right)$, and the corresponding marker points are located between the values 2 and 3 (and between -2 and -3) on the X-axis.}
        \newline
        \textbf{Reasoning:} As an optional component, the focus annotations provide accurate coordinate values $\left(\sqrt{5},\, 0\right) \quad$ and $\quad \left(-\sqrt{5},\, 0\right)$. In the image, the solid markers representing \uline{these two foci indeed lie on the x-axis within the grid intervals between 2 and 3 (closer to 2) and between -2 and -3 (closer to -2)}, respectively, corresponding to the precise positions of approximately $\pm2.236$. The points are fully consistent with the given values, satisfying the acceptance criteria.
        \newline
        \textbf{Issue:} Error in spatial coordinate identification/inference of the annotation.
        \\
        \midrule
        
        \includegraphics[width=4cm]{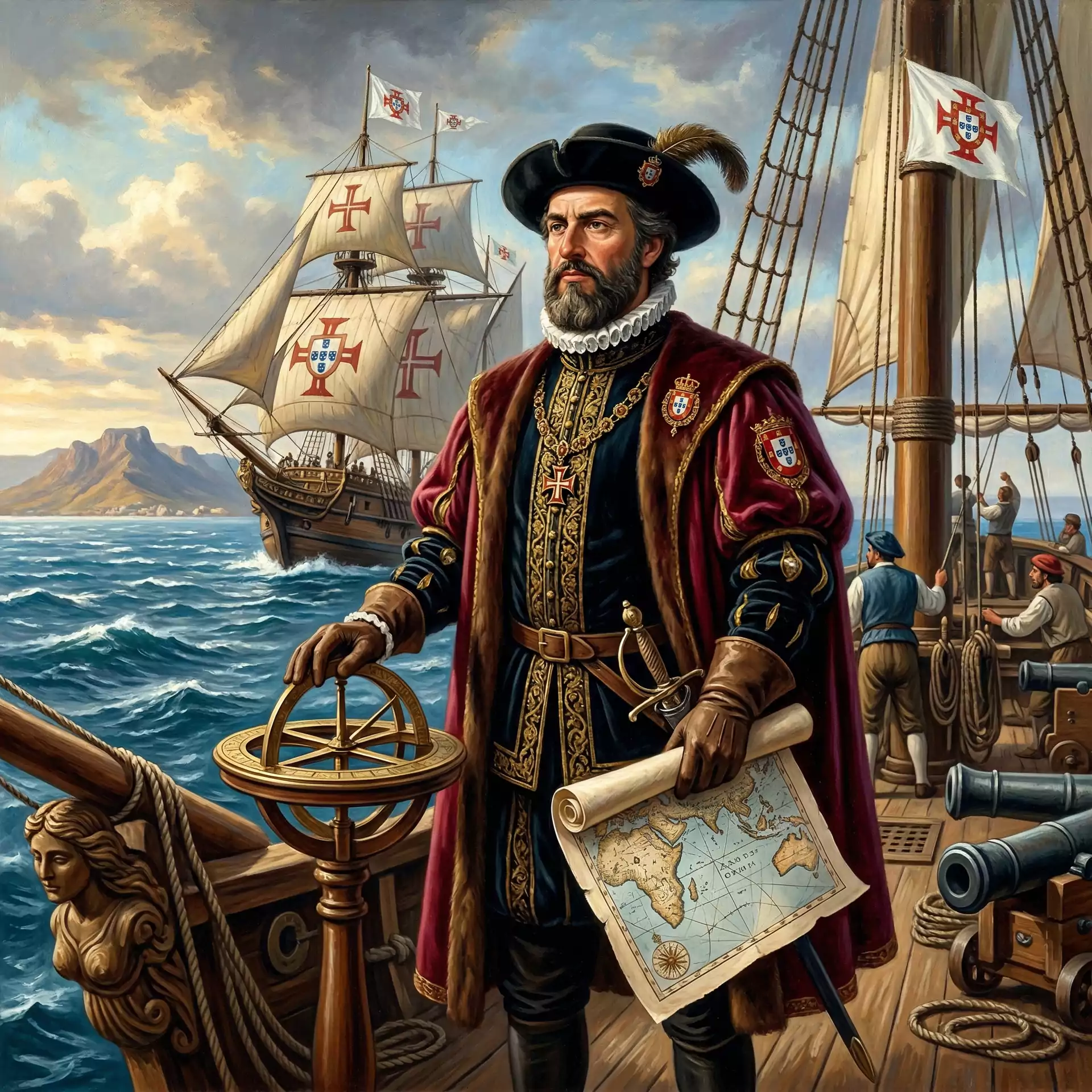}
        &
        \textbf{ID:} \texttt{f16b8fe9-035d-5780-89bf-0cc538e22d9f}
        \newline
        \textbf{VQA:} \textit{There are additional text annotations in the image.}
        \newline
        \textbf{Reasoning:} As an optional element, the model generated textual annotations on the nautical chart held by the character. However, upon closer inspection, the text—composed of Latin-like characters \uline{(e.g., “A U O D P M”, “O R Y M I”)}, is nonsensical when taken together and forms unintelligible pseudo-gibberish. Therefore, it is judged as a failure.
        \newline
        \textbf{Issue:} The automated evaluation is overly strict. It incorrectly treats decorative or visually embedded content attached to an object as meaningful annotation text, and the criteria for assessing written content are excessively harsh.
        \\
        
        \\
        \bottomrule
    \end{longtable}
\end{tcolorbox}

\subsection{Quantitative Breakdown of Failure Modes}

\begin{figure}[t]
\centering

\begin{minipage}[t]{0.56\linewidth}
\vspace{0pt}
\centering
\includegraphics[
    width=\linewidth,
    trim=20 0 0 0,
    clip
]{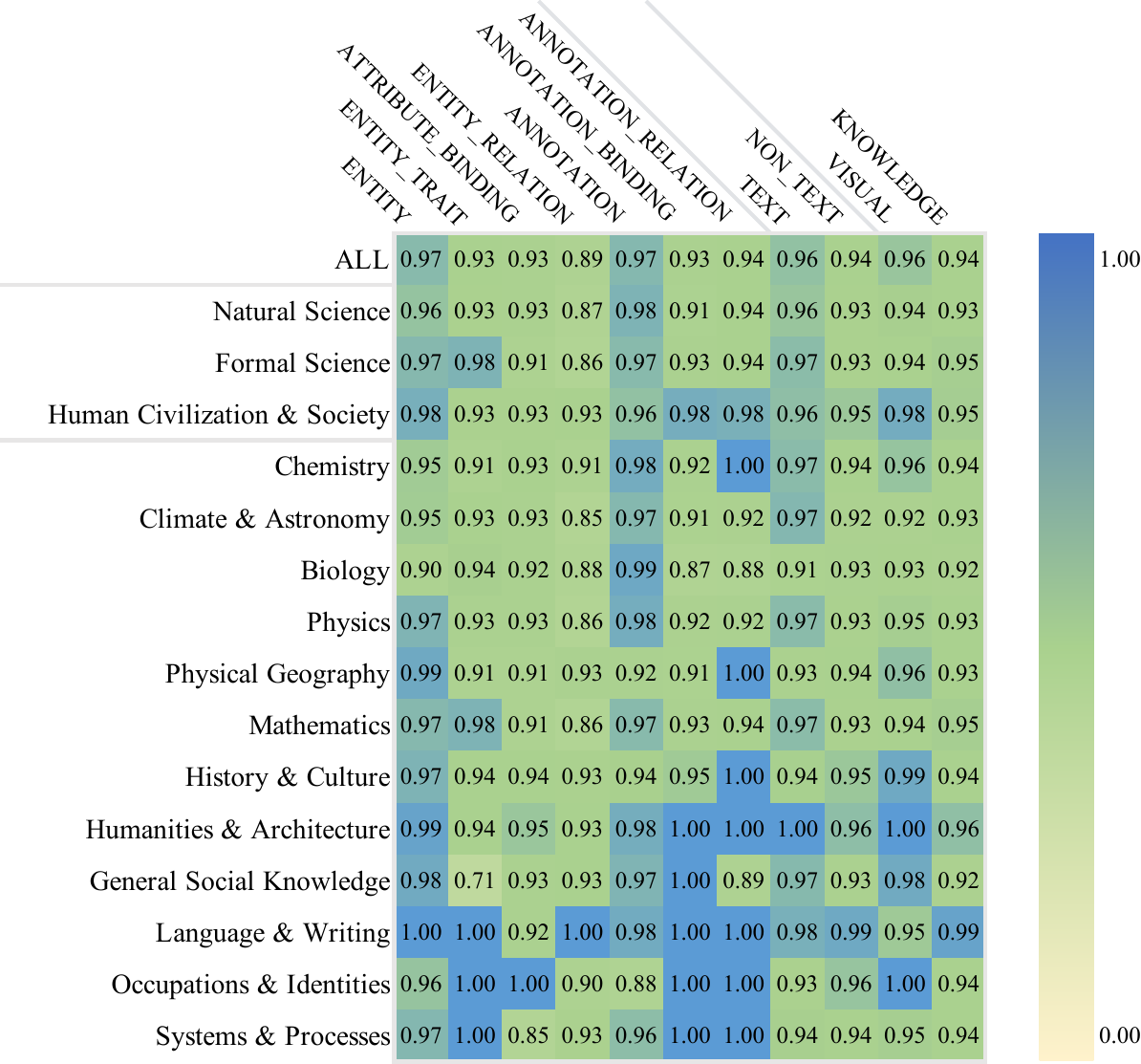}

\vspace{0.3em}
\textbf{(a) VQA-level human-judge agreement.}
\end{minipage}
\hfill
\begin{minipage}[t]{0.40\linewidth}
\vspace{5em}
\centering
\textbf{(b) Top-10 lower-agreement slices.}

\vspace{1em}
\scriptsize
\setlength{\tabcolsep}{3pt}
\renewcommand{\arraystretch}{1.10}
\begin{tabular}{
>{\raggedright\arraybackslash}p{0.30\linewidth}
l
>{\centering\arraybackslash}p{0.13\linewidth}
}
\toprule
\textbf{Category Slice} & \textbf{Component} & \textbf{Agree.} \\
\midrule
General Social Knowledge & \texttt{ENTITY\_TRAIT} & 0.71 \\
Climate \& Astronomy & \texttt{ENTITY\_RELATION} & 0.85 \\
Systems \& Processes & \texttt{ATTRIBUTE\_BINDING} & 0.85 \\
Mathematics & \texttt{ENTITY\_RELATION} & 0.86 \\
Physics & \texttt{ENTITY\_RELATION} & 0.86 \\
Biology & \texttt{ANNOTATION\_BINDING} & 0.87 \\
Biology & \texttt{ENTITY\_RELATION} & 0.88 \\
Biology & \texttt{ANNOTATION\_RELATION} & 0.88 \\
Occupations \& Identities & \texttt{ANNOTATION} & 0.88 \\
General Social Knowledge & \texttt{ANNOTATION\_RELATION} & 0.89 \\
\bottomrule
\end{tabular}
\end{minipage}

\vspace{0.8 em}
\caption{
Judge reliability analysis. 
(a) VQA-level human--judge agreement across taxonomy slices and VQA categories. 
(b) Top-10 lower-agreement slices.
}
\label{fig:judge_reliability_analysis}
\end{figure}

To complement the qualitative case analysis above, we provide a quantitative breakdown of the MLLM judge's
remaining failure modes. Figure~\ref{fig:judge_reliability_analysis}(a) reports VQA-level human--judge agreement
across taxonomy slices and VQA categories, while Figure~\ref{fig:judge_reliability_analysis}(b) lists the top-10
lower-agreement cells. Overall, agreement remains high across most slices, supporting the use of the structured
judge for large-scale batch evaluation.

The lower-agreement cells are not uniformly distributed. The lowest agreement appears in
\texttt{General Social Knowledge} with \texttt{ENTITY\_TRAIT} agreement of 0.71. Other low-agreement cells are
mainly associated with relation-, binding-, and annotation-related VQA categories, including
\texttt{ENTITY\_RELATION}, \texttt{ATTRIBUTE\_BINDING}, \texttt{ANNOTATION\_BINDING}, and
\texttt{ANNOTATION\_RELATION}. These results indicate that the remaining judge instability is concentrated in
fine-grained attribute recognition and structured visual reasoning, rather than uniformly affecting all VQA
categories.

\paragraph{Relation-sensitive judgments.}
The overly lenient judgments described above motivate closer examination of
relation-sensitive checks. In the evaluated \texttt{ENTITY\_RELATION}
slice, 774 VQAs account for 12.66\% of all VQAs. Agreement with human
annotations reaches 89.28\% accuracy and Cohen's $\kappa=0.7483$.
These statistics characterize the component as a whole; they do not isolate
the frequency of errors involving exact counts, duplicates, or inconsistent
instances. Such failure-specific effects require a targeted analysis.

\FloatBarrier
\subsection{Impact of Judge Errors on Model Rankings}
\label{app:judge_error_ranking_impact}

The preceding qualitative and component-level analyses identify where the
judge makes errors, but do not establish how those errors affect comparative
model rankings. We therefore conduct a human-correction audit of a total of 50 prompt--image pairs sampled from the Skill-Tree outputs of five T2I models. We correct the identified judge errors and
recompute the model rankings within the audited subset.

\begin{table}[htbp]
\centering
\small
\caption{Model rankings before and after human correction of identified
judge errors in the five-model audit. These ranks are computed within the
audited subset and should not be equated with the ten-model rankings in
Table~\ref{tab:cross_judge_complete_rankings}.}
\label{tab:judge_error_corrected_rankings}
\begin{tabular}{lrrr}
\toprule
T2I model & Before & After & Absolute shift \\
\midrule
Nano Banana 2 & 1 & 1 & 0 \\
Seedream 5.0 Lite & 2 & 3 & 1 \\
Qwen Image 2.0 & 3 & 2 & 1 \\
FLUX.2 Pro & 4 & 4 & 0 \\
{\klingimageomnithree} & 5 & 5 & 0 \\
\bottomrule
\end{tabular}
\end{table}

As shown in Table~\ref{tab:judge_error_corrected_rankings}, Nano Banana 2
remains first, the fourth and fifth positions remain unchanged, and only
the second and third positions are exchanged. The rankings before and after
correction yield Kendall's $\tau=0.8000$ and Spearman's $\rho=0.9000$.
No model moves by more than one position. Within this audited subset, the
corrections preserve the leading model and produce only a local change in
ordering. The audit does not establish that judge errors are negligible
across the full benchmark or within every task category.

This analysis concerns the combined effect of the judge errors identified
in the audit. It does not isolate the effect of overly lenient PASS decisions
from other failure types. In particular, the component-level agreement
statistics in H.2 should not be interpreted as a direct measurement of
ranking changes caused specifically by duplicate-sensitive errors.

\paragraph{Future refinement.}
The limited ranking changes observed in the audit do not remove the need to
address the failure patterns identified above. In particular, overly lenient
judgments motivate explicit checks for object counts, duplicates,
consistency across instances, and subject--attribute bindings when required
by the task. We plan to evaluate whether these refinements reduce false-PASS
errors while avoiding additional false-FAIL errors, and to explore human
review for ambiguous cases. These refinements have not been incorporated
into the protocol used for the reported results.

\section{Additional Experimental Results}

\subsection{Complementary Results on the General-Domain T2I Benchmark}
\label{app:general_domain_t2i}

The General-Domain T2I Benchmark is used as a supplementary control to distinguish world-knowledge grounding from general image-generation capability.
Unlike the Skill-Tree and Generalized benchmarks, this benchmark focuses on non-knowledge-intensive prompts and evaluates models along general instruction following, visual quality, and aesthetic quality.
Its purpose is not to rank models by general visual performance, but to contextualize whether low world-knowledge scores can be explained by broad deficiencies in basic T2I generation.

\begin{figure}[t]
    \centering
    \includegraphics[width=\linewidth]{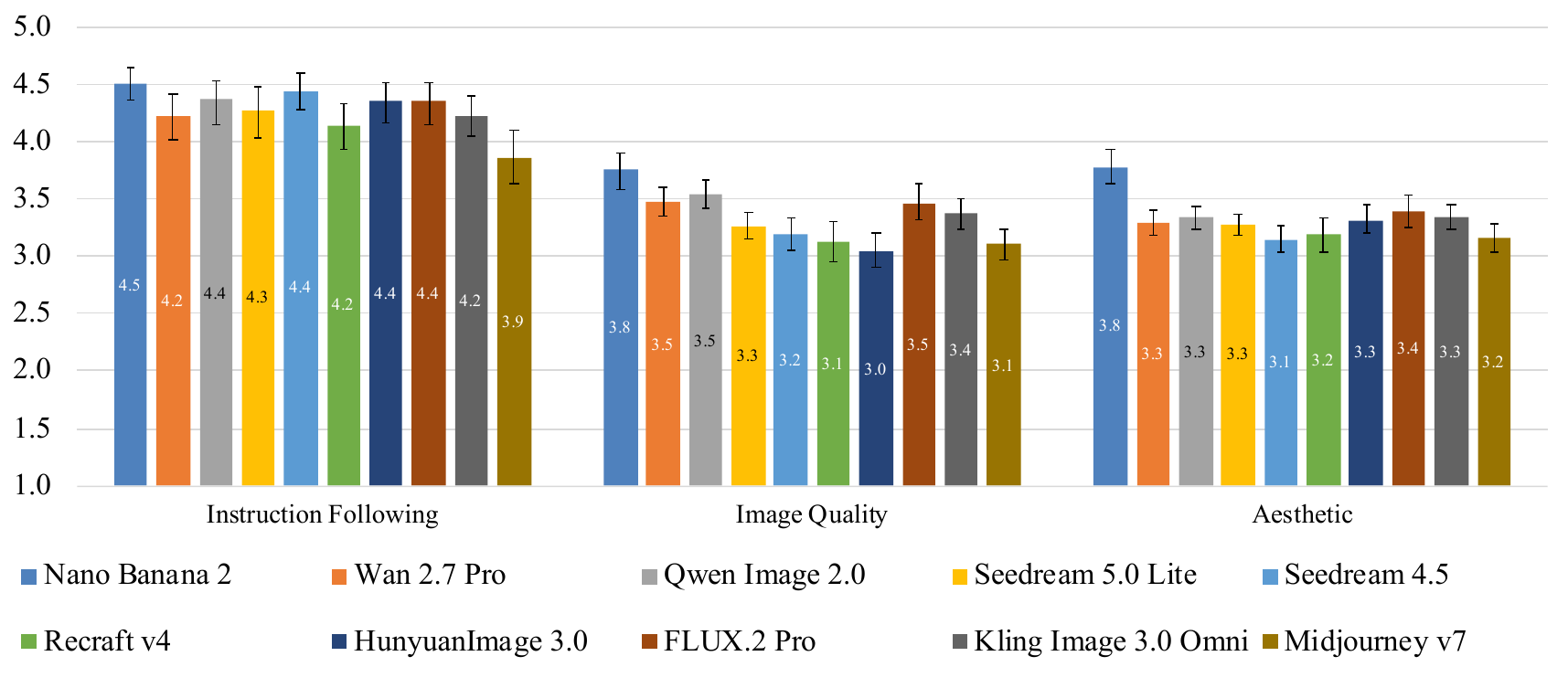}
    \caption{
    Complementary evaluation on the General-Domain T2I Benchmark.
    We compare models on instruction following, visual quality, and aesthetic quality to contextualize world-knowledge grounding results.
    }
    \label{fig:general_domain_t2i}
\end{figure}

Figure~\ref{fig:general_domain_t2i} reports the General-Domain T2I results.
Several models achieve competitive general-domain scores, especially on instruction following, visual quality, or aesthetics, while still showing larger differences on Skill-Tree correctness and Generalized knowledge dimensions.
This indicates that general T2I capability and world-knowledge grounding are related but not interchangeable.

For example, Qwen-Image 2.0 and FLUX.2 Pro perform well on general instruction following and visual quality, but still lag behind the strongest model in world-knowledge grounding.
Midjourney V7 shows aesthetic strength, yet its knowledge correctness and richness remain weaker on the world-knowledge benchmarks.
{\klingimageomnithree} remains within a comparable range on general-domain performance, suggesting that its weaker world-knowledge results should be interpreted more as a knowledge-specific limitation than as a broad deficiency in basic generation capability.

Overall, this supplementary benchmark supports the need for {\benchmark}: evaluating general instruction following, visual quality, or aesthetics alone cannot reliably reveal whether a model correctly grounds world knowledge in generated images.

\subsection{Default-Judge Reliability and Component Ablations}
\label{app:judge_reliability_ablation}

We examine the contribution of structured auxiliary fields using Gemini 3.1 Pro Preview on all 480 annotated images in the Skill-Tree subset, including both the development and validation splits. We separately remove
taxonomy context (\texttt{category}), knowledge hints
(\texttt{knowledge\_hint}), and visual grounding cues
(\texttt{visual\_grounding}), and also evaluate a setting without all three
fields. Models, samples, VQAs, scoring rules, and aggregation are held fixed.
Table~\ref{tab:judge_reliability_ablation} reports the complete ablation results.

\begin{table}[!htbp]
\centering
\small
\caption{Judge reliability and component ablation on the Skill-Tree Benchmark.
We report human-judge agreement at both the case level and the VQA level.
Values are percentages; values in parentheses indicate changes in percentage points compared with the full protocol.}
\label{tab:judge_reliability_ablation}

\begingroup
\renewcommand{\arraystretch}{1.18}
\setlength{\tabcolsep}{4pt}

\begin{tabular}{
>{\raggedright\arraybackslash}p{2.7cm}
ccc
ccc
ccc
}
\toprule
\multirow{2}{*}{\textbf{Setting}}
& \multicolumn{3}{c}{\textbf{Components}}
& \multicolumn{3}{c}{\makecell{\textbf{Case-level}\\\textbf{Agreement}}}
& \multicolumn{3}{c}{\makecell{\textbf{VQA-level}\\\textbf{Agreement}}} \\
\cline{2-10}
& \textbf{Cat.} & \textbf{KH.} & \textbf{VG.}
& \textbf{Acc.} & \textbf{Rec.} & \textbf{Prec.}
& \textbf{Acc.} & \textbf{Rec.} & \textbf{Prec.} \\
\midrule

\rowcolor{green!5}
full protocol
& \cmark & \cmark & \cmark
& \textbf{88.12} & \textbf{85.19} & \textbf{95.11}
& \textbf{94.51} & \textbf{80.22} & \textbf{92.07} \\
\midrule

w/o category
& \xmark & \cmark & \cmark
& \makecell{86.46\\{\scriptsize\negdelta{(-1.66)}}}
& \makecell{85.86\\{\scriptsize\posdelta{(+0.67)}}}
& \makecell{91.73\\{\scriptsize\negdelta{(-3.38)}}}
& \makecell{93.03\\{\scriptsize\negdelta{(-1.48)}}}
& \makecell{77.44\\{\scriptsize\negdelta{(-2.78)}}}
& \makecell{87.29\\{\scriptsize\negdelta{(-4.78)}}} \\
\midrule

w/o knowledge hint
& \cmark & \xmark & \cmark
& \makecell{86.67\\{\scriptsize\negdelta{(-1.45)}}}
& \makecell{86.53\\{\scriptsize\posdelta{(+1.34)}}}
& \makecell{91.46\\{\scriptsize\negdelta{(-3.65)}}}
& \makecell{92.53\\{\scriptsize\negdelta{(-1.98)}}}
& \makecell{77.44\\{\scriptsize\negdelta{(-2.78)}}}
& \makecell{84.93\\{\scriptsize\negdelta{(-7.14)}}} \\
\midrule

w/o visual grounding
& \cmark & \cmark & \xmark
& \makecell{86.07\\{\scriptsize\negdelta{(-2.05)}}}
& \makecell{81.82\\{\scriptsize\negdelta{(-3.37)}}}
& \makecell{94.92\\{\scriptsize\negdelta{(-0.19)}}}
& \makecell{93.56\\{\scriptsize\negdelta{(-0.95)}}}
& \makecell{76.17\\{\scriptsize\negdelta{(-4.05)}}}
& \makecell{91.07\\{\scriptsize\negdelta{(-1.00)}}} \\
\midrule

w/o all components
& \xmark & \xmark & \xmark
& \makecell{86.04\\{\scriptsize\negdelta{(-2.08)}}}
& \makecell{83.84\\{\scriptsize\negdelta{(-1.35)}}}
& \makecell{92.91\\{\scriptsize\negdelta{(-2.20)}}}
& \makecell{92.49\\{\scriptsize\negdelta{(-2.02)}}}
& \makecell{76.89\\{\scriptsize\negdelta{(-3.33)}}}
& \makecell{85.21\\{\scriptsize\negdelta{(-6.86)}}} \\
\bottomrule

\end{tabular}
\endgroup
\end{table}

Removing any auxiliary field lowers case-level and VQA-level accuracy.
Removing all three fields decreases accuracy by 2.08 percentage points at
the case level and 2.02 points at the VQA level. The largest VQA-level
precision reduction occurs when knowledge hints are removed, from 92.07\%
to 84.93\%, a decrease of 7.14 points. These results show that explicit
auxiliary information improves agreement with human annotations under the
tested configuration. They do not establish that the judge is independent
of its own world knowledge, visual perception, or reasoning capabilities.

\FloatBarrier
\subsection{Human--Judge Agreement on the Generalized Benchmark}
\label{app:generalized_judge_agreement}

The Skill-Tree validation focuses on bounded atomic knowledge tasks. We
additionally assess the default Gemini 3.1 Pro Preview judge on 50 sampled
Generalized cases to examine agreement in open-ended scenarios. Human review
covers both Required VQAs, which assess core knowledge correctness, and
Optional VQAs, which assess predefined supplementary knowledge expressions.
We compare judge predictions with the human annotations at the VQA level,
rather than comparing only the final case-level scores.

\begin{table}[htbp]
\centering
\small
\caption{VQA-level accuracy of the default judge against human annotations
on 50 sampled Generalized cases. Required and Optional VQAs correspond to
knowledge correctness and predefined knowledge richness, respectively.}
\label{tab:generalized_judge_agreement}
\begin{tabular}{lr}
\toprule
VQA subset & Accuracy (\%) \\
\midrule
Required & 98.41 \\
Optional & 98.53 \\
All VQAs & 98.44 \\
\bottomrule
\end{tabular}
\end{table}

Table~\ref{tab:generalized_judge_agreement} shows high agreement with human
annotations for both Required and Optional VQAs on the sampled cases.
This analysis complements the Skill-Tree validation by directly examining
open-ended knowledge expressions. The two studies use different evaluation
samples, so their accuracy values should not be interpreted as a controlled
comparison of evaluation difficulty. The present analysis concerns the
default judge and does not establish cross-judge robustness on the
Generalized Benchmark.

\FloatBarrier
\subsection{Human--Judge Agreement Across Judge Settings}
\label{app:cross_judge_agreement}

We compare four evaluation settings: the original Gemini 3.1 Pro Preview
judge, a same-setting rerun, Gemini 3.5 Flash, and GPT 5.5. We report
agreement with human annotations at both the VQA and case levels. VQA-level
metrics assess individual checklist judgments, while case-level metrics
assess the aggregated Required-VQA pass/fail decisions.
Table~\ref{tab:cross_judge_full_agreement} extends the main-text comparison
with complete case-level results.

\begin{table}[htbp]
\centering
\small
\caption{Complete human--judge agreement across four judge settings.
Acc., Prec., and Rec. denote accuracy, precision, and recall, reported as
percentages. $\kappa$ denotes Cohen's $\kappa$. Original and rerun refer to
Gemini 3.1 Pro Preview; Flash refers to Gemini 3.5 Flash.}
\label{tab:cross_judge_full_agreement}
\begin{tabular}{lrrrr}
\toprule
Metric & Original & Rerun & Flash & GPT 5.5 \\
\midrule
VQA Acc. & 94.51 & 91.74 & 92.21 & 91.33 \\
VQA Prec. & 92.07 & 83.63 & 86.74 & 86.03 \\
VQA Rec. & 80.22 & 74.44 & 73.41 & 69.11 \\
VQA $\kappa$ & 0.8236 & 0.7366 & 0.7475 & 0.7140 \\
\midrule
Case Acc. & 88.12 & 83.54 & 83.33 & 84.79 \\
Case Prec. & 95.11 & 91.29 & 92.89 & 93.41 \\
Case Rec. & 85.19 & 81.14 & 79.12 & 81.14 \\
Case $\kappa$ & 0.7562 & 0.6628 & 0.6623 & 0.6903 \\
\bottomrule
\end{tabular}
\end{table}

Across all tested settings, VQA-level accuracy remains above 91\%.
Although the rerun and alternative judges yield lower agreement
metrics than the original evaluation, they retain substantial
agreement with human annotations: VQA-level Cohen's $\kappa$
ranges from 0.7140 to 0.7475, while case-level $\kappa$ ranges
from 0.6623 to 0.6903. Their accuracy ranges from 91.33\% to
92.21\% at the VQA level and from 83.33\% to 84.79\% at the
case level.

The same-setting rerun yields VQA-level and case-level $\kappa$
of 0.7366 and 0.6628, respectively, compared with 0.8236 and
0.7562 in the original evaluation. These results show that
agreement with human annotations remains substantial at both
levels, while its exact magnitude varies across evaluations.
The present comparison does not establish the cause of this
variation. Appendix~\ref{app:cross_judge_ranking} further examines
whether the main model-ranking trends remain stable across
these judge settings.

\FloatBarrier
\subsection{Model Ranking Consistency Across Judge Settings}
\label{app:cross_judge_ranking}

We compare the rankings of ten T2I models under the four judge settings,
using the original Gemini 3.1 Pro Preview ranking as the reference.
Table~\ref{tab:cross_judge_complete_rankings} provides the complete rankings and shifts relative to the baseline,
and Table~\ref{tab:cross_judge_rank_statistics} summarizes their consistency.
The original ranking is the reference ordering for this comparison, not an
independently established ground-truth ranking.

\begin{table}[htbp]
\centering
\small
\caption{Ranking shifts across judge settings for ten T2I models.
The baseline is the original Gemini 3.1 Pro Preview evaluation; the Gemini
3.1 Pro Preview column reports the same-setting rerun.
Rank 1 is best. We define $\Delta=r_{\mathrm{baseline}}-r_{\mathrm{judge}}$,
so positive values indicate an improvement in rank and negative values
indicate a decline.}
\label{tab:cross_judge_complete_rankings}
\begingroup
\setlength{\tabcolsep}{5pt}
\renewcommand{\arraystretch}{1.12}
\begin{tabular}{lrrrrrrr}
\toprule
\multirow{2}{*}{\textbf{T2I model}}
& \multirow{2}{*}{\makecell{\textbf{Baseline}\\\textbf{rank}}}
& \multicolumn{2}{c}{\makecell{\textbf{Gemini 3.1 Pro}\\\textbf{Preview (rerun)}}}
& \multicolumn{2}{c}{\makecell{\textbf{Gemini 3.5}\\\textbf{Flash}}}
& \multicolumn{2}{c}{\textbf{GPT 5.5}} \\
\cmidrule(lr){3-4}\cmidrule(lr){5-6}\cmidrule(lr){7-8}
& & Rank & $\Delta$ & Rank & $\Delta$ & Rank & $\Delta$ \\
\midrule
Nano Banana 2 & 1 & 1 & $0$ & 1 & $0$ & 1 & $0$ \\
Seedream 5.0 Lite & 2 & 2 & $0$ & 2 & $0$ & 2 & $0$ \\
Recraft V4 & 3 & 4 & $-1$ & 3 & $0$ & 3 & $0$ \\
HunyuanImage 3.0 & 4 & 5 & $-1$ & 5 & $-1$ & 5 & $-1$ \\
Qwen Image 2.0 & 5 & 3 & $+2$ & 4 & $+1$ & 4 & $+1$ \\
Wan 2.7 Pro & 6 & 8 & $-2$ & 8 & $-2$ & 8 & $-2$ \\
FLUX.2 Pro & 7 & 6 & $+1$ & 6 & $+1$ & 7 & $0$ \\
Seedream 4.5 & 8 & 7 & $+1$ & 7 & $+1$ & 6 & $+2$ \\
{\klingimageomnithree} & 9 & 9 & $0$ & 10 & $-1$ & 9 & $0$ \\
Midjourney V7 & 10 & 10 & $0$ & 9 & $+1$ & 10 & $0$ \\
\bottomrule
\end{tabular}
\endgroup
\end{table}

\begin{table}[htbp]
\centering
\small
\caption{Ranking consistency relative to the original Gemini 3.1 Pro Preview
ordering. Top-3 Jaccard measures overlap of the top-three model sets and does
not measure their internal ordering. Maximum shift is the largest absolute
rank difference among the ten models.}
\label{tab:cross_judge_rank_statistics}
\begingroup
\setlength{\tabcolsep}{4pt}
\begin{tabular}{lrrrr}
\toprule
Judge setting & Kendall's $\tau_b$ & Spearman's $\rho$ & Top-3 Jaccard & Max. shift \\
\midrule
Gemini 3.1 Pro Preview (rerun) & 0.8222 & 0.9273 & 0.5000 & 2 \\
Gemini 3.5 Flash & 0.8222 & 0.9394 & 1.0000 & 2 \\
GPT 5.5 & 0.8222 & 0.9394 & 1.0000 & 2 \\
\bottomrule
\end{tabular}
\endgroup
\end{table}

All three comparisons yield Kendall's $\tau_b=0.8222$, while Spearman's
$\rho$ ranges from 0.9273 to 0.9394. The first two models remain unchanged
in every setting. Gemini 3.5 Flash and GPT 5.5 also preserve the original
top-three ordering. In the Gemini rerun, Qwen Image 2.0 replaces Recraft V4
in third place, giving a top-three set overlap of two models and a Jaccard
index of $2/4=0.5$. No model moves by more than two positions.

These results support the stability of the main ranking trends under the
tested strong MLLM judges, while allowing local changes in ordering. They
do not establish that absolute benchmark scores are invariant, that small
score differences are statistically significant, or that smaller judge
models would provide comparable results. The distinct question of how
identified judge errors affect rankings after human correction is examined
in Appendix~\ref{app:judge_error_ranking_impact}.

\section{Limitation}
\label{sec:limitation}
Although this paper systematically explores task definition, benchmark construction, and evaluation protocol design, several limitations remain.

First, the modeling of knowledge richness requires further refinement. Our current approach primarily uses optional knowledge expressions to assess whether generated images convey knowledge more completely. However, how to integrate predefined knowledge expressions with valid additional knowledge expressed in generated outputs, and how to aggregate them in a stable and generalizable manner, remain open questions.

Second, although the scoring protocol separates knowledge correctness from image quality, residual perceptual coupling may remain in human evaluation. Factual or semantic inconsistencies may influence perceived image quality even when annotators are instructed to assess these dimensions separately. Future versions will investigate a more explicit distinction among narrow image quality, visual/structural plausibility, and knowledge plausibility, and evaluate whether this refinement improves inter-annotator agreement and error attribution. This finer-grained separation has not been implemented as a separate set of reported evaluation dimensions in the current benchmark.

Third, evaluation reliability depends on the capabilities of the MLLM judges. Although the protocol can be implemented with different judge backends, its reliability still depends on each judge’s world knowledge, visual perception, and reasoning capabilities. Knowledge hints and visual grounding cues provide explicit references but do not eliminate these dependencies or potential model-specific biases. Our cross-judge experiments support the stability of the main ranking trends among the tested strong MLLMs, rather than across arbitrary judge models. Preliminary exploration with smaller models has not established comparable reliability, and their suitability as substitutes for the current judges remains an open question. Developing reliable evaluation with less capable and more accessible judge models is therefore an important direction for future work.

Finally, this paper primarily focuses on static image generation and has not yet been systematically extended to more complex scenarios such as image editing, video generation, or multi-turn interactive generation. These tasks introduce additional challenges, including temporal consistency, cross-turn knowledge preservation, multimodal constraint alignment, and local editing consistency. Extending the benchmark and evaluation protocol to address these challenges remains an important direction for future work.

\section{Resources and Reproducibility}
\label{app:resources_and_reproducibility}

\subsection{Experimental Setup Details}
\label{subsec:experimental_setup_details}

This section provides additional details about the evaluation sets, data splits, and model selection used in our experiments. We separate these details from the main text to keep the experimental section concise while preserving reproducibility.

\subsubsection{Evaluation Sets}
\label{subsubsec:evaluation_sets}

We use three evaluation sets with complementary roles: the Skill-Tree Benchmark, the Generalized Benchmark, and a General-Domain T2I Benchmark. Table~\ref{tab:appendix_evaluation_sets} summarizes their full size, experimental subset size, coverage principle.
\begin{table}[t]
\centering
\small
\caption{Evaluation sets used in the experiments.}
\label{tab:appendix_evaluation_sets}
\renewcommand{\arraystretch}{1.12}
\setlength{\tabcolsep}{5pt}
\begin{tabularx}{\linewidth}{
>{\raggedright\arraybackslash}p{0.30\linewidth}
>{\centering\arraybackslash}p{0.18\linewidth}
>{\raggedright\arraybackslash}p{0.22\linewidth}
>{\raggedright\arraybackslash}X
}
\toprule
Evaluation Set & Full / Eval. & Release & Coverage \\
\midrule
Skill-Tree Benchmark
& 1,622 / 480
& Full benchmark
& All L4 atomic knowledge points \\

Generalized Benchmark
& 550 / 100
& Full benchmark
& All L2 subjects and L3 branches \\

General-Domain T2I Benchmark
& 600 / 100
& 100-case subset only
& General entities and scenes \\
\bottomrule
\end{tabularx}
\end{table}

\paragraph{Skill-Tree Benchmark.}
The full Skill-Tree Benchmark contains 1,622 cases covering 240 L4 atomic knowledge points. For experiments related to judge development, reliability validation, ablation, and metric comparison, we uniformly sample 480 cases from the full benchmark, ensuring coverage of all L4 atomic knowledge points. This subset is further split into 240 visible development cases and 240 unseen validation cases.

\paragraph{Generalized Benchmark.}
The full Generalized Benchmark contains 550 open-ended cases. For model evaluation, we uniformly sample 100 cases from the full set, ensuring coverage of all L2 subjects and L3 branches. This subset is used to evaluate comprehensive world-knowledge grounding under complex open-ended scenarios. The MLLM judge evaluates structured dimensions, including knowledge correctness, knowledge richness, and non-knowledge visual instruction following, while human annotators rate image quality, aesthetics, and overall satisfaction using MOS.

\paragraph{General-Domain T2I Benchmark.}
The General-Domain T2I Benchmark contains 100 general generation tasks. It covers common entity and scene categories, including portraits, animals, objects, plants, and historical and cultural scenes. It is used as a supplementary evaluation of general T2I capability. Its instruction-following dimension reuses a mature internal automatic evaluation system, while image quality and aesthetics are evaluated using human MOS.

\subsubsection{Data Splits}
\label{subsubsec:data_splits}
\paragraph{Skill-Tree split.}
The 480-case Skill-Tree experimental subset is sampled from the full 1,622-case Skill-Tree Benchmark with coverage over all L4 atomic knowledge points. The visible development set contains 240 cases and is used to inspect judge behavior, refine the MLLM-as-a-judge prompt, calibrate VQA construction guidelines, and iterate on the evaluation scheme. The unseen validation set contains another 240 cases and is used for model evaluation and for the ranking comparison across judges. Human--judge agreement and cross-judge agreement use the full 480-case subset.

\subsubsection{Agreement Subsets}
\label{subsubsec:agreement_subsets}

Human--judge agreement, human--human agreement, and cross-judge agreement use the subsets in Table~\ref{tab:appendix_agreement_subsets}. These subsets are taken from the experimental sets in Table~\ref{tab:appendix_evaluation_sets}.

\begin{table}[t]
\centering
\small
\caption{Subsets used for human--judge, human--human, and cross-judge agreement.}
\label{tab:appendix_agreement_subsets}
\renewcommand{\arraystretch}{1.12}
\setlength{\tabcolsep}{5pt}
\begin{tabularx}{\linewidth}{
>{\raggedright\arraybackslash}p{0.34\linewidth}
>{\raggedright\arraybackslash}X
>{\raggedright\arraybackslash}p{0.24\linewidth}
}
\toprule
Check & Source & Subset \\
\midrule
Human--judge agreement
& Skill-Tree 480-case subset
& All 480 cases \\

Human--judge agreement
& Generalized 100-case subset
& 50 sampled cases \\

Human--human agreement
& Skill-Tree 480-case subset
& 50 sampled cases \\

Cross-judge agreement
& Skill-Tree 480-case subset
& All 480 cases \\
Ranking consistency
& Skill-Tree unseen validation split
& All 240 cases \\
\bottomrule
\end{tabularx}
\end{table}

\subsubsection{T2I Model Selection}
\label{subsubsec:model_selection}

We evaluate models from three overlapping groups: general-purpose T2I models, aesthetics- or visual-quality-oriented models, and text-/knowledge-oriented models. These groups are designed to test whether world-knowledge grounding aligns with general image quality, aesthetics, text rendering, or knowledge-oriented generation capability. Since these capabilities are not mutually exclusive, some models appear in multiple groups. Table~\ref{tab:appendix_model_selection} summarizes the model groups and provides references to the corresponding technical reports or official model announcements.

\begin{table}[t]
\centering
\small
\caption{Model groups used in the experiments. Groups are overlapping by design.}
\label{tab:appendix_model_selection}
\renewcommand{\arraystretch}{1.18}
\setlength{\tabcolsep}{5pt}
\begin{tabularx}{\linewidth}{
>{\raggedright\arraybackslash}p{0.28\linewidth}
>{\raggedright\arraybackslash}X
}
\toprule
Capability Axis & Representative Models \\
\midrule
General-purpose T2I
& Nano Banana 2~\citep{raisinghani2026nanobanana2};
FLUX.2 Pro~\citep{bfl2025flux2};
Recraft V4~\citep{dorogush2026recraftv4};
Seedream 4.5~\citep{seedream2025seedream4};
Hunyuan Image 3.0~\citep{cao2025hunyuanimage};
{\klingimageomnithree}~\citep{klingai2026kling3} \\

Aesthetics / Visual quality
& Midjourney V7~\citep{midjourney2025v7};
{\klingimageomnithree}~\citep{klingai2026kling3};
Seedream 4.5~\citep{seedream2025seedream4} \\

Text / Knowledge
& Nano Banana 2~\citep{raisinghani2026nanobanana2};
Seedream 5.0 Lite~\citep{bytedance2026seedream5lite};
Wan 2.7 Pro~\citep{alibaba2026wan27image};
Qwen-Image 2.0~\citep{qwen2025image,qwen2026image2} \\
\bottomrule
\end{tabularx}
\end{table}

\subsubsection{T2I Generation Settings}
\label{subsubsec:t2i_generation_settings}

\paragraph{Prompt language.}
We use Chinese prompts for models developed in China and English
prompts for models developed outside China. Each model uses a fixed
prompt language throughout the evaluation. The bilingual prompts
preserve the same underlying task and semantic intent.

\paragraph{Output resolution.}
We select 2K output whenever supported by the selected model
variant; otherwise, we use the
highest available output resolution. Table~\ref{tab:t2i_generation_settings}
summarizes the prompt language and output image dimensions for
each model. All images use a square aspect ratio.

\begin{table}[t]
\centering
\small
\caption{Providers, prompt languages, and output resolutions of
the evaluated T2I models, ordered alphabetically by provider.
We prioritize 2K generation and otherwise use the highest
available resolution.}
\label{tab:t2i_generation_settings}
\renewcommand{\arraystretch}{1.15}
\setlength{\tabcolsep}{4pt}
\begin{tabularx}{\linewidth}{
    >{\raggedright\arraybackslash}X
    >{\raggedright\arraybackslash}p{0.23\linewidth}
    >{\centering\arraybackslash}p{0.13\linewidth}
    >{\centering\arraybackslash}p{0.22\linewidth}
}
\toprule
Model & Provider & Language & Output resolution \\
\midrule
Qwen-Image 2.0
& Alibaba & Chinese & $2048 \times 2048$ \\
Wan 2.7 Pro
& Alibaba & Chinese & $2048 \times 2048$ \\
FLUX.2 Pro
& Black Forest Labs & English & $2000 \times 2000$ \\
Seedream 4.5
& ByteDance & Chinese & $2048 \times 2048$ \\
Seedream 5.0 Lite
& ByteDance & Chinese & $2048 \times 2048$ \\
Nano Banana 2
& Google & English & $2048 \times 2048$ \\
{\klingimageomnithree}
& KlingAI & Chinese & $2048 \times 2048$ \\
Midjourney V7
& Midjourney & English & $1024 \times 1024$ \\
Recraft V4
& Recraft & English & $1024 \times 1024$ \\
Hunyuan Image 3.0
& Tencent & Chinese & $1024 \times 1024$ \\
\bottomrule
\end{tabularx}
\end{table}

\paragraph{Model-specific settings.}
Nano Banana 2 is used with web search and image search enabled.
Seedream 4.5 and Seedream 5.0 Lite are accessed through the
Jimeng web interface. Qwen-Image 2.0 is accessed through
\texttt{chat.qwen.ai} with Qwen 3.6 Plus selected as the
conversational interface. Midjourney uses version~7
(\texttt{--v 7}). For Recraft, we use the V4 Standard variant.

\paragraph{Collection period.}
The image files used for evaluation were archived in April 2026.

\subsubsection{Evaluation Consistency}
\label{subsubsec:evaluation_consistency}

To ensure fair and reproducible comparison, all models are evaluated on the same case sets, semantic prompt intents, benchmark metadata, scoring rubrics, and MLLM-as-a-Judge protocol within each evaluation set. 
Because each case includes bilingual prompts, the selected language variant is kept fixed according to the model-specific language setting used in the experiments. 
Unless otherwise specified, automatic scores are produced by the same MLLM judge under the same judge prompts, auxiliary metadata, and inference configuration. 
\subsubsection{Statistical Significance}
\label{subsubsec:statistical_significance}

All quantitative evaluation results reported in this paper are accompanied by 95\% confidence intervals (CIs), computed via bootstrap resampling (1,000 resampling iterations) over the test set. The CI bounds reflect variability due to random sampling of test examples, and are reported as asymmetric intervals to avoid out-of-range values. Error bars in all figures and tables correspond to these bootstrap 95\% CIs.

\subsection{Released Resources}
\label{app:released_resources}

We release the benchmark schema, case metadata, bilingual prompts, prompt templates, VQA checklists, evaluation scripts, and reproducibility instructions.

\paragraph{Code.} Please refer to \url{https://github.com/QuantumWhisper42/KnowVis} for complete codebase.
\paragraph{Dataset.} Please refer to \url{https://huggingface.co/datasets/QuantumWhisper42/KnowVis} for complete datasets.

\subsection{Model API Configuration}

Dataset construction, checklist decomposition, labeling, and the default judge use Google Gemini models through the Vertex AI API (\url{https://cloud.google.com/products/gemini-enterprise-agent-platform}). Table~\ref{tab:model_api_configurations} lists those settings. Table~\ref{tab:rebuttal_judge_configurations} lists Gemini 3.5 Flash and GPT 5.5, the two additional judges used for the cross-judge checks. Gemini 3.1 Pro Preview is unchanged from the Judge column of Table~\ref{tab:model_api_configurations}.

\begin{table}[ht!]
    \centering
    \renewcommand{\arraystretch}{1.18}
    \setlength{\tabcolsep}{2pt}
    \small
    \begin{threeparttable}
        \caption{Google Gemini API configurations for dataset construction, checklist decomposition, labeling, and the default judge.}
        \label{tab:model_api_configurations}
        \begin{tabular}{lccccc}
            \toprule
        
            \multirow{2}{*}{\textbf{Configurations}}
            & \multicolumn{2}{c}{\textbf{Dataset Construction}}
            & \multicolumn{2}{c}{\textbf{Automatic Evaluation}}
            & \textbf{Misc}
            \\
            
            \cline{2-6}

            & Generation
            & Filtering
            & VQA
            & Judge
            & Labeling
            \\
            \midrule
            
            \textbf{API Entrypoint\tnote{$\star$}}
            & \texttt{3.1-pro}
            & \texttt{3-flash}
            & \texttt{3.1-pro-customtools}
            & \texttt{3.1-pro}
            & \texttt{3-flash}
            \\
            \midrule
            
            \textbf{Thinking Level}
            & \multicolumn{4}{c}{\texttt{HIGH}}
            & \texttt{UNSPECIFIED}
            \\
            \midrule
            
            \textbf{Media Resolution}
            & \multicolumn{4}{c}{\texttt{HIGH}}
            & N/A\tnote{$\star\star$}
            \\
            \midrule
            
            \textbf{Built-in Tools}
            & \multicolumn{4}{c}{\texttt{Grounding with Google Search \& URL Context}}
            & -\tnote{$\dagger$}
            \\
            \midrule

            \textbf{Custom Tools}
            & -
            & -
            & -
            & -
            & Wikimedia Commons \tnote{$\ddagger$}
            \\
            
            \bottomrule
        \end{tabular}
        \begin{tablenotes}
            \scriptsize
            \item[$\star$] \begin{itemize}[leftmargin=*]
                \item \texttt{3.1-pro} $\to$ \texttt{gemini-3.1-pro-preview},
                \item \texttt{3.1-pro-customtools} $\to$ \texttt{gemini-3.1-pro-preview-customtools},
                \item \texttt{3-flash} $\to$ \texttt{gemini-3-flash-preview}.
            \end{itemize}
            \item[$\star\star$] Not Applicable, as no image modal were input.
            \item[$\dagger$] No need to search the internet and get the context just for some simple labeling tasks.
            \item[$\ddagger$] Custom made tool to search and retrieve images from wikimedia commons. Used to curate high quality \texttt{visual\_grounding} images from trusted open access data source automatically.
        \end{tablenotes}
    \end{threeparttable}
\end{table}

\begin{table}[ht!]
    \centering
    \small
    \caption{Additional judges used for the cross-judge checks. Gemini 3.5 Flash reuses the default Judge settings in Table~\ref{tab:model_api_configurations}.}
    \label{tab:rebuttal_judge_configurations}
    \renewcommand{\arraystretch}{1.15}
    \setlength{\tabcolsep}{6pt}
    \begin{tabularx}{\linewidth}{
        >{\raggedright\arraybackslash}p{0.26\linewidth}
        >{\raggedright\arraybackslash}X
        >{\raggedright\arraybackslash}X
    }
        \toprule
        Configuration & Gemini 3.5 Flash & GPT 5.5 \\
        \midrule
        Model
        & \texttt{gemini-3.5-flash}
        & \texttt{gpt-5.5} \\
        Thinking / reasoning
        & \texttt{HIGH}
        & reasoning effort \texttt{high} \\
        Media resolution
        & \texttt{HIGH}
        & not set \\
        Built-in tools
        & Google Search and URL Context
        & none \\
        Structured output
        & same as the default judge
        & native \\
        \bottomrule
    \end{tabularx}
\end{table}

\subsection{Local Computing Resources}

We used a MacBook Pro equipped with an Apple M3 Pro chip and 36 GB of unified memory to compute CLIPScore and VQAScore. Due to memory constraints, CLIP-FlanT5-XL was used to compute VQAScore.

\section{Broader Impacts}
\label{app:broader_impacts}

{\benchmark} is an evaluation benchmark and does not introduce or deploy a new generative model. Its potential positive impacts include improving the factual reliability of text-to-image generation, supporting educational and scientific visualization, and providing the community with a structured diagnostic resource for evaluating world-knowledge grounding. By decomposing knowledge constraints into VQA-based evaluation units, the benchmark can help identify whether model failures arise from knowledge recall, reasoning, annotation errors, visual instruction following, or general image quality issues.

Potential negative impacts are mainly indirect. The benchmark and its feedback could be misused to optimize models for generating more convincing but misleading knowledge-intensive images, such as incorrect educational diagrams, deceptive scientific illustrations, or fabricated professional-domain content. In addition, over-reliance on automated MLLM-as-a-judge evaluation may propagate judge errors or biases, especially in high-stakes, time-sensitive, or professional domains.

To mitigate these risks, we recommend releasing the benchmark with clear documentation, intended-use statements, and known limitations. For time-sensitive or professional-domain cases, future versions should include explicit temporal references, source verification, and update mechanisms where appropriate. Human review should remain necessary for high-stakes applications, and MLLM-as-a-judge results should be interpreted as diagnostic signals rather than definitive judgments.


\newpage
\section*{NeurIPS Paper Checklist}

\begin{enumerate}

\item {\bf Claims}
    \item[] Question: Do the main claims made in the abstract and introduction accurately reflect the paper's contributions and scope?
    \item[] Answer: \answerYes{} 
    \item[] The Abstract and Introduction~\ref{sec:introduction} clearly state the paper's three contributions and scope. Experimental results are consistent with the claimed goals, demonstrating that {\benchmark} reveals structural weaknesses of existing T2I models in world-knowledge grounding.
    \item[] Guidelines:
    \begin{itemize}
        \item The answer \answerNA{} means that the abstract and introduction do not include the claims made in the paper.
        \item The abstract and/or introduction should clearly state the claims made, including the contributions made in the paper and important assumptions and limitations. A \answerNo{} or \answerNA{} answer to this question will not be perceived well by the reviewers. 
        \item The claims made should match theoretical and experimental results, and reflect how much the results can be expected to generalize to other settings. 
        \item It is fine to include aspirational goals as motivation as long as it is clear that these goals are not attained by the paper. 
    \end{itemize}

\item {\bf Limitations}
    \item[] Question: Does the paper discuss the limitations of the work performed by the authors?
    \item[] Answer:\answerYes{} 
    \item[] Justification: The paper includes a dedicated Limitation~\ref{sec:limitation} section covering three aspects: knowledge richness metric refinement, coupled evaluation dimensions in open-ended scenarios, and the restriction to static image generation. The key assumption of MLLM judge reliability is addressed through validation against human annotations.
    \item[] Guidelines:
    \begin{itemize}
        \item The answer \answerNA{} means that the paper has no limitation while the answer \answerNo{} means that the paper has limitations, but those are not discussed in the paper. 
        \item The authors are encouraged to create a separate ``Limitations'' section in their paper.
        \item The paper should point out any strong assumptions and how robust the results are to violations of these assumptions (e.g., independence assumptions, noiseless settings, model well-specification, asymptotic approximations only holding locally). The authors should reflect on how these assumptions might be violated in practice and what the implications would be.
        \item The authors should reflect on the scope of the claims made, e.g., if the approach was only tested on a few datasets or with a few runs. In general, empirical results often depend on implicit assumptions, which should be articulated.
        \item The authors should reflect on the factors that influence the performance of the approach. For example, a facial recognition algorithm may perform poorly when image resolution is low or images are taken in low lighting. Or a speech-to-text system might not be used reliably to provide closed captions for online lectures because it fails to handle technical jargon.
        \item The authors should discuss the computational efficiency of the proposed algorithms and how they scale with dataset size.
        \item If applicable, the authors should discuss possible limitations of their approach to address problems of privacy and fairness.
        \item While the authors might fear that complete honesty about limitations might be used by reviewers as grounds for rejection, a worse outcome might be that reviewers discover limitations that aren't acknowledged in the paper. The authors should use their best judgment and recognize that individual actions in favor of transparency play an important role in developing norms that preserve the integrity of the community. Reviewers will be specifically instructed to not penalize honesty concerning limitations.
    \end{itemize}

\item {\bf Theory assumptions and proofs}
    \item[] Question: For each theoretical result, does the paper provide the full set of assumptions and a complete (and correct) proof?
    \item[] Answer: \answerNA{} 
    \item[] Justification: This paper presents a benchmark and evaluation protocol without theoretical results, proofs, or formal theorems.
    \item[] Guidelines:
    \begin{itemize}
        \item The answer \answerNA{} means that the paper does not include theoretical results. 
        \item All the theorems, formulas, and proofs in the paper should be numbered and cross-referenced.
        \item All assumptions should be clearly stated or referenced in the statement of any theorems.
        \item The proofs can either appear in the main paper or the supplemental material, but if they appear in the supplemental material, the authors are encouraged to provide a short proof sketch to provide intuition. 
        \item Inversely, any informal proof provided in the core of the paper should be complemented by formal proofs provided in appendix or supplemental material.
        \item Theorems and Lemmas that the proof relies upon should be properly referenced. 
    \end{itemize}

    \item {\bf Experimental result reproducibility}
    \item[] Question: Does the paper fully disclose all the information needed to reproduce the main experimental results of the paper to the extent that it affects the main claims and/or conclusions of the paper (regardless of whether the code and data are provided or not)?
    \item[] Answer: \answerYes{} 
    \item[] The automated data construction and filtering pipeline, benchmark data, and code are all publicly released, along with API configurations, system prompts, and other implementation details. Full reproducibility information is consolidated in Appendix~\ref{app:resources_and_reproducibility}.
    \item[] Guidelines:
    \begin{itemize}
        \item The answer \answerNA{} means that the paper does not include experiments.
        \item If the paper includes experiments, a \answerNo{} answer to this question will not be perceived well by the reviewers: Making the paper reproducible is important, regardless of whether the code and data are provided or not.
        \item If the contribution is a dataset and\slash or model, the authors should describe the steps taken to make their results reproducible or verifiable. 
        \item Depending on the contribution, reproducibility can be accomplished in various ways. For example, if the contribution is a novel architecture, describing the architecture fully might suffice, or if the contribution is a specific model and empirical evaluation, it may be necessary to either make it possible for others to replicate the model with the same dataset, or provide access to the model. In general. releasing code and data is often one good way to accomplish this, but reproducibility can also be provided via detailed instructions for how to replicate the results, access to a hosted model (e.g., in the case of a large language model), releasing of a model checkpoint, or other means that are appropriate to the research performed.
        \item While NeurIPS does not require releasing code, the conference does require all submissions to provide some reasonable avenue for reproducibility, which may depend on the nature of the contribution. For example
        \begin{enumerate}
            \item If the contribution is primarily a new algorithm, the paper should make it clear how to reproduce that algorithm.
            \item If the contribution is primarily a new model architecture, the paper should describe the architecture clearly and fully.
            \item If the contribution is a new model (e.g., a large language model), then there should either be a way to access this model for reproducing the results or a way to reproduce the model (e.g., with an open-source dataset or instructions for how to construct the dataset).
            \item We recognize that reproducibility may be tricky in some cases, in which case authors are welcome to describe the particular way they provide for reproducibility. In the case of closed-source models, it may be that access to the model is limited in some way (e.g., to registered users), but it should be possible for other researchers to have some path to reproducing or verifying the results.
        \end{enumerate}
    \end{itemize}

\item {\bf Open access to data and code}
    \item[] Question: Does the paper provide open access to the data and code, with sufficient instructions to faithfully reproduce the main experimental results, as described in supplemental material?
    \item[] Answer: \answerYes{} 
    \item[] Justification: The dataset and code are publicly released with sufficient instructions for reproducing all main experimental results, as detailed in Appendix~\ref{app:resources_and_reproducibility}.
    \item[] Guidelines:
    \begin{itemize}
        \item The answer \answerNA{} means that paper does not include experiments requiring code.
        \item Please see the NeurIPS code and data submission guidelines (\url{https://neurips.cc/public/guides/CodeSubmissionPolicy}) for more details.
        \item While we encourage the release of code and data, we understand that this might not be possible, so \answerNo{} is an acceptable answer. Papers cannot be rejected simply for not including code, unless this is central to the contribution (e.g., for a new open-source benchmark).
        \item The instructions should contain the exact command and environment needed to run to reproduce the results. See the NeurIPS code and data submission guidelines (\url{https://neurips.cc/public/guides/CodeSubmissionPolicy}) for more details.
        \item The authors should provide instructions on data access and preparation, including how to access the raw data, preprocessed data, intermediate data, and generated data, etc.
        \item The authors should provide scripts to reproduce all experimental results for the new proposed method and baselines. If only a subset of experiments are reproducible, they should state which ones are omitted from the script and why.
        \item At submission time, to preserve anonymity, the authors should release anonymized versions (if applicable).
        \item Providing as much information as possible in supplemental material (appended to the paper) is recommended, but including URLs to data and code is permitted.
    \end{itemize}

\item {\bf Experimental setting/details}
    \item[] Question: Does the paper specify all the training and test details (e.g., data splits, hyperparameters, how they were chosen, type of optimizer) necessary to understand the results?
    \item[] Answer: \answerYes{} 
    \item[] Justification: All experimental details including evaluated models, API configurations, and evaluation protocol settings are specified in the paper and Appendix~\ref{app:resources_and_reproducibility}.
    \item[] Guidelines:
    \begin{itemize}
        \item The answer \answerNA{} means that the paper does not include experiments.
        \item The experimental setting should be presented in the core of the paper to a level of detail that is necessary to appreciate the results and make sense of them.
        \item The full details can be provided either with the code, in appendix, or as supplemental material.
    \end{itemize}

\item {\bf Experiment statistical significance}
    \item[] Question: Does the paper report error bars suitably and correctly defined or other appropriate information about the statistical significance of the experiments?
    \item[] Answer: \answerYes{} 
    \item[] Justification: Confidence intervals for all main evaluation results are reported using bootstrap resampling (1,000 iterations, 95\% CI). No parametric distributional assumptions (e.g., normality) are made, as MOS-style scores derived from human judgments on text-to-image generation tasks do not follow a well-defined parametric distribution. For details, refer to Appendix~\ref{app:resources_and_reproducibility} Subsection~\ref{subsubsec:statistical_significance}.
    \item[] Guidelines:
    \begin{itemize}
        \item The answer \answerNA{} means that the paper does not include experiments.
        \item The authors should answer \answerYes{} if the results are accompanied by error bars, confidence intervals, or statistical significance tests, at least for the experiments that support the main claims of the paper.
        \item The factors of variability that the error bars are capturing should be clearly stated (for example, train/test split, initialization, random drawing of some parameter, or overall run with given experimental conditions).
        \item The method for calculating the error bars should be explained (closed form formula, call to a library function, bootstrap, etc.)
        \item The assumptions made should be given (e.g., Normally distributed errors).
        \item It should be clear whether the error bar is the standard deviation or the standard error of the mean.
        \item It is OK to report 1-sigma error bars, but one should state it. The authors should preferably report a 2-sigma error bar than state that they have a 96\% CI, if the hypothesis of Normality of errors is not verified.
        \item For asymmetric distributions, the authors should be careful not to show in tables or figures symmetric error bars that would yield results that are out of range (e.g., negative error rates).
        \item If error bars are reported in tables or plots, the authors should explain in the text how they were calculated and reference the corresponding figures or tables in the text.
    \end{itemize}

\item {\bf Experiments compute resources}
    \item[] Question: For each experiment, does the paper provide sufficient information on the computer resources (type of compute workers, memory, time of execution) needed to reproduce the experiments?
    \item[] Answer: \answerYes{} 
    \item[] Justification: The experiments primarily rely on API calls to commercial T2I and MLLM models, with no large-scale local compute required. Local computation is limited to running CLIPScore and VQAScore, which require only standard CPU/GPU resources. Detailed API usage and local computation resources is described in Appendix~\ref{app:resources_and_reproducibility}.
    \item[] Guidelines:
    \begin{itemize}
        \item The answer \answerNA{} means that the paper does not include experiments.
        \item The paper should indicate the type of compute workers CPU or GPU, internal cluster, or cloud provider, including relevant memory and storage.
        \item The paper should provide the amount of compute required for each of the individual experimental runs as well as estimate the total compute. 
        \item The paper should disclose whether the full research project required more compute than the experiments reported in the paper (e.g., preliminary or failed experiments that didn't make it into the paper). 
    \end{itemize}
    
\item {\bf Code of ethics}
    \item[] Question: Does the research conducted in the paper conform, in every respect, with the NeurIPS Code of Ethics \url{https://neurips.cc/public/EthicsGuidelines}?
    \item[] Answer: \answerYes{} 
    \item[] Justification: The research conforms to the NeurIPS Code of Ethics. The benchmark data does not involve personal privacy. Human annotators were compensated at or above the local minimum hourly wage. The benchmark covers diverse knowledge domains to mitigate cultural bias, and potential misuse and judge-related risks are discussed in Appendix~\ref{app:broader_impacts}.
    \item[] Guidelines:
    \begin{itemize}
        \item The answer \answerNA{} means that the authors have not reviewed the NeurIPS Code of Ethics.
        \item If the authors answer \answerNo, they should explain the special circumstances that require a deviation from the Code of Ethics.
        \item The authors should make sure to preserve anonymity (e.g., if there is a special consideration due to laws or regulations in their jurisdiction).
    \end{itemize}

\item {\bf Broader impacts}
    \item[] Question: Does the paper discuss both potential positive societal impacts and negative societal impacts of the work performed?
    \item[] Answer: \answerYes{} 
    \item[] Justification: Broader impacts including positive impacts, potential negative impacts, and mitigation strategies are discussed in Appendix~\ref{app:broader_impacts}.
    \item[] Guidelines:
    \begin{itemize}
        \item The answer \answerNA{} means that there is no societal impact of the work performed.
        \item If the authors answer \answerNA{} or \answerNo, they should explain why their work has no societal impact or why the paper does not address societal impact.
        \item Examples of negative societal impacts include potential malicious or unintended uses (e.g., disinformation, generating fake profiles, surveillance), fairness considerations (e.g., deployment of technologies that could make decisions that unfairly impact specific groups), privacy considerations, and security considerations.
        \item The conference expects that many papers will be foundational research and not tied to particular applications, let alone deployments. However, if there is a direct path to any negative applications, the authors should point it out. For example, it is legitimate to point out that an improvement in the quality of generative models could be used to generate Deepfakes for disinformation. On the other hand, it is not needed to point out that a generic algorithm for optimizing neural networks could enable people to train models that generate Deepfakes faster.
        \item The authors should consider possible harms that could arise when the technology is being used as intended and functioning correctly, harms that could arise when the technology is being used as intended but gives incorrect results, and harms following from (intentional or unintentional) misuse of the technology.
        \item If there are negative societal impacts, the authors could also discuss possible mitigation strategies (e.g., gated release of models, providing defenses in addition to attacks, mechanisms for monitoring misuse, mechanisms to monitor how a system learns from feedback over time, improving the efficiency and accessibility of ML).
    \end{itemize}
    
\item {\bf Safeguards}
    \item[] Question: Does the paper describe safeguards that have been put in place for responsible release of data or models that have a high risk for misuse (e.g., pre-trained language models, image generators, or scraped datasets)?
    \item[] Answer: \answerYes{} 
    \item[] Justification: Portions of the dataset contain prompts generated by LLMs and some fields sourced from Wikimedia Commons under open licenses (CC and public domain), both have undergone human review to ensure quality and safety, discussed in Appendix~\ref{app:skill_tree_benchmark_synthesis}. We do not release a new image-generation model, and the benchmark does not include high-risk personal data.
    \item[] Guidelines:
    \begin{itemize}
        \item The answer \answerNA{} means that the paper poses no such risks.
        \item Released models that have a high risk for misuse or dual-use should be released with necessary safeguards to allow for controlled use of the model, for example by requiring that users adhere to usage guidelines or restrictions to access the model or implementing safety filters. 
        \item Datasets that have been scraped from the Internet could pose safety risks. The authors should describe how they avoided releasing unsafe images.
        \item We recognize that providing effective safeguards is challenging, and many papers do not require this, but we encourage authors to take this into account and make a best faith effort.
    \end{itemize}

\item {\bf Licenses for existing assets}
    \item[] Question: Are the creators or original owners of assets (e.g., code, data, models), used in the paper, properly credited and are the license and terms of use explicitly mentioned and properly respected?
    \item[] Answer: \answerYes{} 
    \item[] Justification: All evaluated T2I models are properly cited. Wikimedia Commons assets are used under CC and public domain licenses, with detailed per-asset attribution (title, author, source, and license) included in the released dataset on Hugging Face, as described in Appendix~\ref{app:resources_and_reproducibility}.
    \item[] Guidelines:
    \begin{itemize}
        \item The answer \answerNA{} means that the paper does not use existing assets.
        \item The authors should cite the original paper that produced the code package or dataset.
        \item The authors should state which version of the asset is used and, if possible, include a URL.
        \item The name of the license (e.g., CC-BY 4.0) should be included for each asset.
        \item For scraped data from a particular source (e.g., website), the copyright and terms of service of that source should be provided.
        \item If assets are released, the license, copyright information, and terms of use in the package should be provided. For popular datasets, \url{paperswithcode.com/datasets} has curated licenses for some datasets. Their licensing guide can help determine the license of a dataset.
        \item For existing datasets that are re-packaged, both the original license and the license of the derived asset (if it has changed) should be provided.
        \item If this information is not available online, the authors are encouraged to reach out to the asset's creators.
    \end{itemize}

\item {\bf New assets}
    \item[] Question: Are new assets introduced in the paper well documented and is the documentation provided alongside the assets?
    \item[] Answer: \answerYes{} 
    \item[] Justification: The released dataset and code are accompanied by documentation on Hugging Face and GitHub, covering dataset structure, field descriptions, license, and usage instructions, as described in Appendix~\ref{app:resources_and_reproducibility}.
    \item[] Guidelines:
    \begin{itemize}
        \item The answer \answerNA{} means that the paper does not release new assets.
        \item Researchers should communicate the details of the dataset\slash code\slash model as part of their submissions via structured templates. This includes details about training, license, limitations, etc. 
        \item The paper should discuss whether and how consent was obtained from people whose asset is used.
        \item At submission time, remember to anonymize your assets (if applicable). You can either create an anonymized URL or include an anonymized zip file.
    \end{itemize}

\item {\bf Crowdsourcing and research with human subjects}
    \item[] Question: For crowdsourcing experiments and research with human subjects, does the paper include the full text of instructions given to participants and screenshots, if applicable, as well as details about compensation (if any)? 
    \item[] Answer: \answerNA{} 
    \item[] Justification: The paper does not involve crowdsourcing or research with human subjects. Human evaluation was conducted by internal annotators and is not the subject of study.
    \item[] Guidelines:
    \begin{itemize}
        \item The answer \answerNA{} means that the paper does not involve crowdsourcing nor research with human subjects.
        \item Including this information in the supplemental material is fine, but if the main contribution of the paper involves human subjects, then as much detail as possible should be included in the main paper. 
        \item According to the NeurIPS Code of Ethics, workers involved in data collection, curation, or other labor should be paid at least the minimum wage in the country of the data collector. 
    \end{itemize}

\item {\bf Institutional review board (IRB) approvals or equivalent for research with human subjects}
    \item[] Question: Does the paper describe potential risks incurred by study participants, whether such risks were disclosed to the subjects, and whether Institutional Review Board (IRB) approvals (or an equivalent approval/review based on the requirements of your country or institution) were obtained?
    \item[] Answer: \answerNA{} 
    \item[] Justification: The paper does not involve crowdsourcing or research with human subjects, and therefore does not require IRB approval.
    \item[] Guidelines:
    \begin{itemize}
        \item The answer \answerNA{} means that the paper does not involve crowdsourcing nor research with human subjects.
        \item Depending on the country in which research is conducted, IRB approval (or equivalent) may be required for any human subjects research. If you obtained IRB approval, you should clearly state this in the paper. 
        \item We recognize that the procedures for this may vary significantly between institutions and locations, and we expect authors to adhere to the NeurIPS Code of Ethics and the guidelines for their institution. 
        \item For initial submissions, do not include any information that would break anonymity (if applicable), such as the institution conducting the review.
    \end{itemize}

\item {\bf Declaration of LLM usage}
    \item[] Question: Does the paper describe the usage of LLMs if it is an important, original, or non-standard component of the core methods in this research? Note that if the LLM is used only for writing, editing, or formatting purposes and does \emph{not} impact the core methodology, scientific rigor, or originality of the research, declaration is not required.
    \item[] Answer: \answerYes{} 
    \item[] Justification: LLMs are core components of this work in two respects: as evaluated subjects (T2I-related models assessed by the benchmark) and as the backbone of the MLLM-based VQA judge protocol for automatic evaluation. Both usages are described in detail in the experimental setup and evaluation protocol sections.
    \item[] Guidelines:
    \begin{itemize}
        \item The answer \answerNA{} means that the core method development in this research does not involve LLMs as any important, original, or non-standard components.
        \item Please refer to our LLM policy in the NeurIPS handbook for what should or should not be described.
    \end{itemize}

\end{enumerate}

\end{document}